%% file: main.tex
\documentclass[10pt,twocolumn,letterpaper]{article}

\input{math_commands.tex}

\usepackage{times}
\usepackage{epsfig}
\usepackage{graphicx}
\usepackage{amsmath}
\usepackage{amssymb}
\usepackage{xcolor}
\usepackage[T1]{fontenc}
\newif\ifcomments

\usepackage{mathtools}
\newcommand{\norm}[1]{\left\lVert#1\right\rVert}

\usepackage{cvpr}
\usepackage[normalem]{ulem}
\usepackage{bm}
\usepackage{dsfont}
\usepackage{diagbox}
\usepackage{array}
\usepackage{multirow}
\usepackage{booktabs}
\usepackage{nccmath}
\usepackage{caption}
\usepackage{subcaption}
\usepackage{float} %% For plotting the figure right after the text
\usepackage{ulem}

\newcommand{\class}[1]{\ensuremath{\mathsf{#1}\xspace}}

\newcommand{\subsec}[1]{\noindent{\textbf{#1~~}}}
\newcommand{\methodname}[2]{\noindent{\textbf{#1}~#2~~}}

\usepackage[pagebackref=true,breaklinks=true,letterpaper=true,colorlinks,bookmarks=false]{hyperref}
\cvprfinalcopy % *** Uncomment this line for the final submission

\def\cvprPaperID{****} % *** Enter the CVPR Paper ID here

\ifcvprfinal\fi
\begin{document}

%%%%%%%%% TITLE
\newcommand{\papertitle}{SAM: The Sensitivity of Attribution Methods to Hyperparameters}
\title{\papertitle}
%\title{\LaTeX\ Author Guidelines for CVPR Proceedings}

\author{Naman Bansal$^*$\\
Auburn University\\
{\tt\small bnaman50@gmail.com}
\and
Chirag Agarwal$^*$\\
University of Illinois at Chicago\\
{\tt\small chiragagarwall12@gmail.com}
\and
Anh Nguyen\thanks{Equal contribution. CA performed this work during his internship at Auburn University.}\\
Auburn University\\
{\tt\small anh.ng8@gmail.com}
}
%\and
%Chirag Agarwal\\
%University of Illinois at Chicago\\
%{\tt\small chiragagarwall12@gmail.com}
%}
%\and
%Anh Nguyen\\
%Auburn University\\

\maketitle
%\thispagestyle{empty}

%%%%%%%%% ABSTRACT
\begin{abstract}
	Attribution methods can provide powerful insights into the reasons for a classifier's decision.
	We argue that a key desideratum of an explanation method is its robustness to input hyperparameters which are often randomly set or empirically tuned.
	High sensitivity to arbitrary hyperparameter choices does not only impede reproducibility but also questions the correctness of an explanation and impairs the trust of end-users.
	In this paper, we provide a thorough empirical study on the sensitivity of existing attribution methods.
	We found an alarming trend that many methods are highly sensitive to changes in their common hyperparameters e.g. even changing a random seed can yield a different explanation!
	Interestingly, such sensitivity is not reflected in the average explanation accuracy scores over the dataset as commonly reported in the literature.
	In addition, explanations generated for robust classifiers (i.e. which are trained to be invariant to pixel-wise perturbations) are surprisingly more robust than those generated for regular classifiers.
	
%	However, the current methods 
%   Interpretability algorithms have surfaced as an important medium to explain model decisions.
%   Current methods have a set of hyperparameters that are empirically tuned and are thus prone to sensitivity and reproducibility issues.
%   In this paper, we provide a thorough study about the sensitivity of explanation methods to their hyperparameters.
%   Through extensive experiments we show that existing approaches are very sensitive and unreliable under some hyperparameter settings.
%   We show that relying on the evaluation performance of output heatmaps can be misleading.
%   Our findings were consistent across two different robust and non-robust classifiers.
\end{abstract}

%%%%%%%%% BODY TEXT
\section{Introduction}
	\label{sec:intro}
	
	\begin{figure}[t]
		\centering
		{
			\begin{flushleft}
				\hspace{-0.1cm}\rotatebox{90}{\hspace{-9.4cm}SG~\cite{smilkov2017smoothgrad}\hspace{1.4cm}MP~\cite{fong2017interpretable}\hspace{1.2cm}SP~\cite{zeiler2014visualizing}\hspace{1.3cm}LIME~\cite{ribeiro2016should}}
			\end{flushleft}	
		}
		\begin{flushleft}
			\vspace{-0.5cm}
			\text{\hspace{0.4cm}Input image\hspace{0.5cm}Attribution maps (\ie explanations)}
		\end{flushleft}
		\vspace{-0.25cm}
		\includegraphics[width=0.45\textwidth]{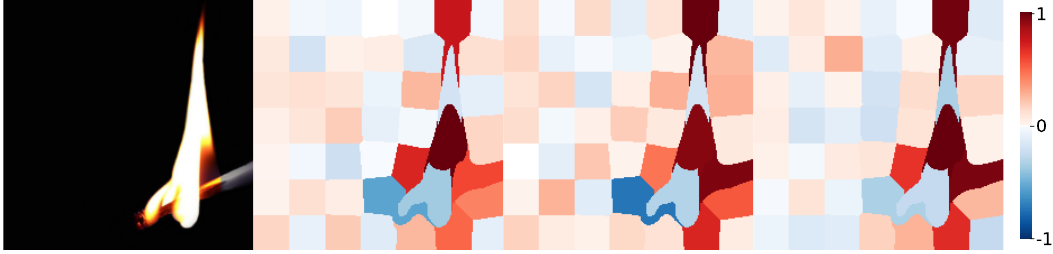}\\
		\begin{flushleft}
			\vspace{-0.4cm}
			\text{\hspace{0.2cm}Random seed:\hspace{0.9cm}$0$\hspace{1.7cm}$1$\hspace{1.75cm}$2$}
			%		\vspace{-0.5cm}
			%		\text{\hspace{2.6cm}seed$ =0$\hspace{0.7cm}seed$ =1$\hspace{0.7cm}seed$ =2$}		
		\end{flushleft}
		\vspace{-0.25cm}
		\includegraphics[width=0.45\textwidth]{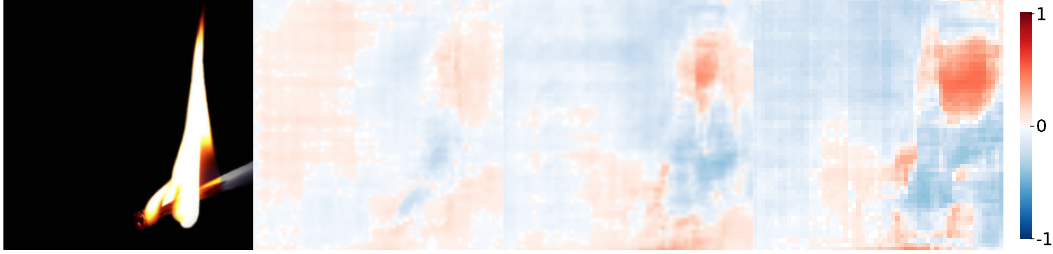}\\
		\begin{flushleft}
			\vspace{-0.4cm}
			\text{\hspace{0.5cm}Patch size:\hspace{0.7cm}$5\times5$\hspace{0.85cm}$29\times29$\hspace{0.9cm}$53\times53$}
			%		\vspace{-0.5cm}
			%		\text{\hspace{2.7cm}$p=5$\hspace{0.9cm}$p=29$\hspace{0.9cm}$p=53$}		
		\end{flushleft}
		\vspace{-0.25cm}		
		\includegraphics[width=0.45\textwidth]{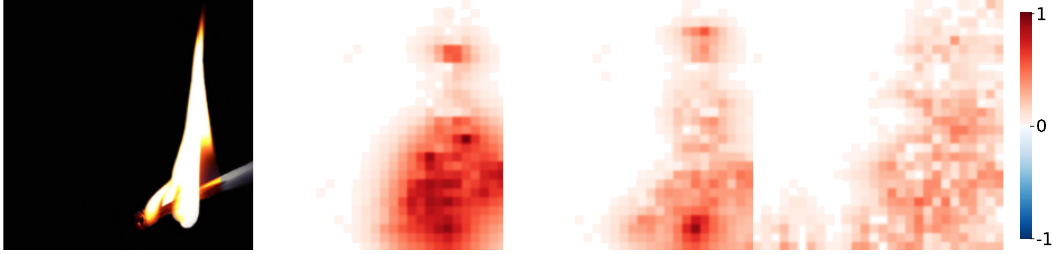}\\
		\begin{flushleft}
			\vspace{-0.4cm}
			\text{\hspace{0.4cm}Blur radius:\hspace{1.0cm}$5$\hspace{1.6cm}$10$\hspace{1.7cm}$30$}		
			%		\vspace{-0.5cm}
			%		\text{\hspace{2.6cm}$\sigma_r =5$\hspace{0.8cm}$\sigma_r =10$\hspace{0.7cm}$\sigma_r =30$}		
		\end{flushleft}
		\vspace{-0.25cm}
%		\vspace{-0.25cm}
		\includegraphics[width=0.45\textwidth]{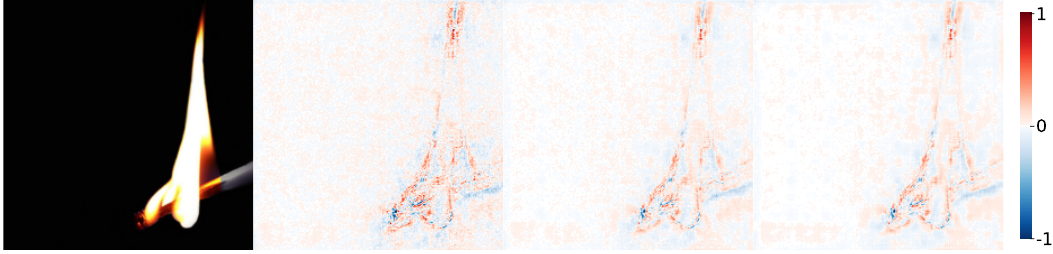}\\
		\begin{flushleft}
			\vspace{-0.4cm}
			\text{\hspace{0.3cm}Sample size:\hspace{0.95cm}$50$\hspace{1.5cm}$200$\hspace{1.4cm}$800$}
		\end{flushleft}
		\caption{
%			The output heatmaps of four well-known, representative attribution methods all change, to different extents, when we vary their most common hyperparameters: increasing the number of samples of SmoothGrad (SG), increasing the patch size of Sliding-Patch (SP), changing the random seed of LIME, and changing the Gaussian blur radius of Meaningful Perturbation (MP).
			Attribution maps by four methods to explain the same prediction (\class{match~stick}: $0.535$) made by a ResNet-50 classifier to an ImageNet image.
			In each row, the explanations are generated by running the default settings of a method while varying only one common hyperparameter.
			All 12 explanations are unique and can be interpreted differently.
			\textbf{LIME:} an explanation changes when one re-runs the algorithm with a different random seed.
			\textbf{SP:} the positive evidence for the fire (top-right red blob) grows together with the patch size.
			\textbf{MP:} attribution maps become more scattered as the Gaussian blur radius increases. 
			\textbf{SG:} heatmaps becomes smoother as the number of samples increases.
		}
		\label{fig:teaser}
	\vspace*{-0.4cm}		
	\end{figure}
	
	Why did a self-driving car decide to run into a truck \cite{lambert2016understanding}? 
	Why is a patient being predicted to have breast cancer \cite{wu2019deep} or to be a future criminal \cite{machine2019propublica}?
%	 or to be unqualified for a job \cite{lahoti2019ifair}?
	The explanations for such predictions made by machine learning (ML) models can impact our lives in many ways, under scientific \cite{szegedy2013intriguing,nguyen2015deep}, social \cite{doshi2017towards} or legal \cite{goodman2017european,doshi2017accountability} aspects.
%	, explaining such decisions of a predictive model is increasingly important and can impact our lives in many ways.

	A popular medium for visually explaining an image classifier's decisions is an \emph{attribution map} \ie a heatmap that highlights the input pixels that are the evidence for and against the classification outputs \cite{montavon2018methods}.
	Dozens of attribution methods (Fig.~\ref{fig:teaser}) have been proposed \cite{samek2019towards} and applied to a variety of domains including natural images \cite{montavon2018methods}, medical brain scans \cite{zintgraf2017visualizing}, text \cite{arras2017}, videos \cite{SriICASSP17}, and speech \cite{becker2018interpreting}.
	Notably, attribution maps have been useful \eg in localizing malignant tumors in a breast x-ray scan \cite{rajpurkar2017chexnet} or in revealing biases in object recognition models \cite{lapuschkin2016analyzing,lapuschkin2017understanding}.
	Yet are these explanations reliable enough to convince medical doctors or judges to accept a life-critical prediction by a machine \cite{lipton2017doctor}?
	
	First, ML techniques often have a set of hyperparameters to be tuned empirically and most attribution methods are not an exception.
	Second, a major cause of the current replication crisis in ML \cite{hutson2018crisis} is that many methods, \eg in reinforcement learning, are notoriously sensitive to hyperparameters---a factor which is also often overlooked in the interpretability field.
	Aside from being faithful, an explanation needs to be reproducible and invariant to arbitrary hyperparameter choices.
	In this paper, we studied an important question: \emph{How sensitive are attribution maps to their hyperparameters?} on 7 well-known attribution methods and found that:\footnote{Code is available at \url{https://github.com/anguyen8/sam}}
%	\anh{Say that the observed inconsistency of heatmaps found in this study also suggests that there can be multiple \emph{correct} explanations; however, existing methods often only show one explanation?}
	
%	\anh{Change $N_{SG}$ to $N$?}
	
	\begin{enumerate}
		\item Gradient heatmaps, for \emph{robust} image classifiers \ie models trained to ignore adversarial pixel-wise noise \cite{engstrom2019learning}, exhibit visible structures (Fig.~\ref{fig:grad_sec_1_1_qual}) in stark contrast to the noisy, uninterpretable gradient images for regular classifiers reported in prior work \cite{smilkov2017smoothgrad} (Sec.~\ref{sec:robust_classifier}).
		
		\item The gradient images from a robust and a regular classifier are different but would appear $\sim$1.5$\times$ more similar, potentially causing misinterpretation, under several prior methods that attempted to de-noise the original explanations \cite{springenberg2014striving,smilkov2017smoothgrad} (Sec.~\ref{sec:question_smoothing}).
		
		\item For many attribution methods \cite{ribeiro2016should,fong2017interpretable,zeiler2014visualizing}, their output heatmaps can change dramatically (Fig.~\ref{fig:teaser}) when a common hyperparameter changes (Sec.~\ref{sec:sensitivity}).
		This sensitivity of an individual explanation also translates into the sensitivity of its accuracy scores  (Sec.~\ref{sec:eval_sensitivity}).
%        \naman{Accuracy score in not a correct term here. Correctness measure is better}
		
		\item Some hyperparameters cause up to $10\times$ more variation in the accuracy of explanations than others (Sec.~\ref{sec:eval_sens_hyper}).
		
		\item Explanations for robust classifiers are not only more invariant to pixel-wise image changes (Sec.~\ref{sec:robust_classifier}) but also to hyperparameter changes (Sec.~\ref{sec:sensitivity})

	\end{enumerate}

\section{Methods and Related Work}
\label{sec:related_work}

Let $f:\sR^{d\times d \times 3} \to [0,1]$ be a classifier that maps a color image $\vx$ of spatial size $d \times d$ onto a probability of a target class.
An attribution method is a function $A$ that takes three inputs---an image $\vx$, the model $f$, and a set of hyperparameters $\gH$---and outputs a matrix $\va = A (f, \vx, \gH) \in [-1,1]^{d\times d}$.
% \ie $A (f, \vx, \gH) = \va$.
Here, the explanation $\va$ associates each input pixel $x_i$ to a scalar $a_i \in [-1,1]$, which indicates how much $x_i$ contributes for or against the classification score $f(\vx)$.

\paragraph{Methods}
Attribution methods can be categorized into two main types: (1) exact and (2) approximate approaches.
\noindent\textbf{Exact approaches} may derive an attribution map by upsampling a feature map of a convolutional network \cite{zhou2016learning}, or from the analytical gradients of the classification w.r.t. the input \ie $\nabla_\vx{f}$ \cite{simonyan2013deep,adebayo2018sanity}, or by combining both the gradients and the feature maps \cite{selvaraju2016grad}.
These approaches enjoy fast derivation of explanations and have no hyperparameters in principles.
However, they require access to the internal network parameters---which may not be available in practice.
Also, taking gradients as attributions faces several issues: (1) gradient images are often noisy \cite{smilkov2017smoothgrad} limiting their utility; (2) gradient saturation \cite{sundararajan2017axiomatic} \ie when the function $f$ flattens within the vicinity of a pixel $x_i$, its gradient becomes near-zero and may misrepresent the actual importance of $x_i$; (3) sudden changes in the gradient $\partial{f} /\ \partial{x_i}$ (\eg from ReLUs \cite{nair2010rectified}) may yield misleading interpretation of the attribution of pixel $x_i$ \cite{shrikumar2016not}.

Therefore, many \textbf{approximate methods} have been proposed to modify the vanilla gradients to address the aforementioned issues \cite{smilkov2017smoothgrad,sundararajan2017axiomatic,bach2015pixel}.
Among gradient-based methods, we chose to study the following four representatives.

\methodname{Gradient}{\cite{simonyan2013deep,adebayo2018sanity}} The gradient image $\nabla_\vx{f}$ quantifies how a small change of each input pixel modifies the classification and therefore commonly serves as an attribution map. 

\methodname{SmoothGrad (SG)}{\cite{smilkov2017smoothgrad}} proposed to smooth out a gradient image by averaging out the gradients over a batch of $N_{SG}$ noisy versions $\vx_n$ of the input image $\vx_0$. That is, an SG heatmap is $\frac{1}{N_{SG}} \sum_{1}^{N_{SG}} \nabla_\vx{f}(\vx_0 + \epsilon)$ where $\epsilon \sim \gN(0, \sigma)$.

\methodname{Gradient $\odot$ Input (GI)}{\cite{shrikumar2016not}}
As gradients are often noisy and thus not interpretable \cite{smilkov2017smoothgrad}, element-wise multiplying the gradient image with the input \ie $\nabla_\vx{f} \odot \vx$ can yield less-noisy heatmaps in practice.
Here, the input image acts as a model-independent smoothing filter.
GI is an approximation of a family of related LRP methods \cite{bach2015pixel} as shown in \cite{ancona2018towards} and is also a representative for other explicit gradient-based extensions \cite{zintgraf2017visualizing, zhang2018top, montavon2017explaining, shrikumar2016not}.

\methodname{Integrated Gradients (IG)}{\cite{sundararajan2017axiomatic}} In order to ameliorate the gradient saturation problem \cite{sundararajan2017axiomatic}, IG intuitively replaces the gradient in GI \cite{shrikumar2016not} with an average of the gradients evaluated for $N_{IG}$ images linearly sampled along a straight line between the original image $\vx$ and a zero image.
IG is intuitively a smooth version of GI and depends on the sample size $N_{IG}$ while GI has no hyperparameters.

Furthermore, there exist other approximate methods 
%\naman{put it like this - ... approximate methods also called perturbation-based methods, that attempt ...}
that attempt to compute the attribution of an input region by replacing it with zeros \cite{zeiler2014visualizing,ribeiro2016should}, random noise \cite{dabkowski2017real}, or blurred versions of the original content \cite{fong2017interpretable}.
These methods inherently depend on many empirically-chosen hyperparameters.  
%\naman{...empirically chosen hyperparameters.}.
Among the family of perturbation-based methods, we chose to study the following three famous representatives.

\methodname{Sliding Patch (SP)}{\cite{zeiler2014visualizing}} slides a square, occlusion patch of size $p \times p$ across the input image and records the prediction changes into an attribution map.
This approach is applicable to any black-box classifier $f$ and widely used \cite{zintgraf2017visualizing,MATLAB,Petsiuk2018rise,ancona2018towards}.

\methodname{LIME}{\cite{ribeiro2016should}} Instead of a square patch, LIME generates $N_{LIME}$ masked images $\{\bm{\bar \vx}^i \}$ by masking out a random set of $S$ non-overlapping superpixels in the input image.
Intuitively, the attribution for a superpixel $k$ is proportional to the average score $f(\bm{\bar x}^i)$ over a batch of $N_{LIME}$ perturbed images where the superpixel $k$ is not masked out.
% in the image $\bm{\bar \vx}^i$ \naman{I am not sure about the correctness of this last sentence}.

\methodname{Meaningful-Perturbation (MP)}{\cite{fong2017interpretable}} finds a minimal Gaussian blur mask of radius $b_R$ such that when applied over the input image would produce a blurred version that has a near-zero classification score.
%\naman{this is too convoluted >-Fong et al. proposes an iterative process to optimize a Gaussian blur mask applied over the input image. The key is to find the minimal mask so that the blurred image has a near zero prediction score.}
MP is the basis for many extensions
\cite{wagner2019interpretable, qi2019visualizing, carletti2018understanding, wang2018learning,uzunova2019interpretable,agarwal2019removing}.
In this paper, we evaluate MP sensitivity to three common hyperparameters: the blur radius $b_R$, the number of steps $N_{iter}$, and the random seed (which determines the random initialization).

See Sec.~\ref{sec:method_description_implementation_details} for a detailed description of all methods.

\paragraph{Explanation sensitivity} 
First, recent work has argued that some attribution methods have a problem of being highly sensitive to small pixel-wise changes in the input image \cite{kindermans2019reliability,alvarez2018robustness,ghorbani2019interpretation}.
%\naman{and pose the input insensitivity an important requirement for the explanation methods \cite{kindermans2019reliability}.}
%However, we note that this desideratum for robustness to input changes assumes that a classifier $f$ also extracts robust features, which may not be the case in practice \cite{szegedy2013intriguing}.
Our results suggest that such sensitivity to image changes also depends on the classifier $f$.
That is, gradient-based explanations of a robust classifier stay more consistent when the input image is perturbed with pixel-wise noise (Sec.~\ref{sec:robust_classifier}).
Second, some attribution methods were found to behave akin to an edge detector \ie producing similar explanations despite that $f$'s parameters are randomized to various degrees \cite{adebayo2018sanity}.
In sum, previous work has studied the sensitivity of explanations to input image changes \cite{kindermans2019reliability,alvarez2018robustness,ghorbani2019interpretation} and classifier changes \cite{adebayo2018sanity}.
In this paper, we present the first systematic study on the sensitivity of explanations to changes in the \emph{hyperparameters} $\gH$, which are often randomly or heuristically tuned \cite{smilkov2017smoothgrad,wei2018explain}.

\begin{figure*}[h]
	\centering
	% 	{	
	% 		\small
	% 		\vspace*{-0.15cm}
	% 		\begin{flushleft}
	% 			\hspace{0.05cm}(a) Input image
	% 			\hspace{0.4cm}(b) Grad
	% 			\hspace{0.6cm}(c) $N_{SG}$=$50$
	% 			\hspace{0.4cm}(d) $N_{SG}$=$100$
	% 			\hspace{0.3cm}(e) $N_{SG}$=$200$
	%			\hspace{0.3cm}(f) $N_{SG}$=$500$
	%			\hspace{0.1cm}(g) $N_{SG}$=$800$
	%			\hspace{0.1cm}(h) Robust Grad	
	% 		\end{flushleft}		
	% 	}
	{	
		\small
		\vspace*{-0.15cm}
		\begin{flushleft}
			\hspace{0.01cm}(a) Input image
			\hspace{0.2cm}(b) Grad
			\hspace{0.2cm}(c) $N_{SG}$=$50$
			\hspace{0.1cm}(d) $N_{SG}$=$100$
			\hspace{0.1cm}(e) $N_{SG}$=$200$
			\hspace{0.1cm}(f) $N_{SG}$=$500$
			\hspace{0.1cm}(g) $N_{SG}$=$800$
			\hspace{0.25cm}(h) GB \cite{springenberg2014striving}				
			\hspace{0.1cm}(i) ResNet-R Grad	
		\end{flushleft}		
	}
	\vspace*{-0.3cm}
	\includegraphics[width=0.99\textwidth]{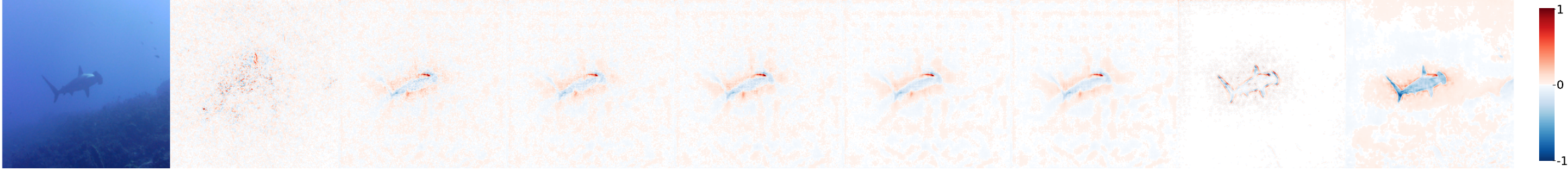}
	%	{	
	%		\small
	%		\vspace*{-0.2cm}
	%		\begin{flushleft}
	%			\hspace{1.0cm}SSIM:
	%			\hspace{0.8cm}0.518
	%			\hspace{1.3cm}0.667
	%			\hspace{1.2cm}0.706
	%			\hspace{1.3cm}0.712
	%			\hspace{1.3cm}0.737
	%			\hspace{1.3cm}0.738
	%			\hspace{1.4cm}1.000	
	%		\end{flushleft}		
	%	}
	{	
		\small
		\vspace*{-0.15cm}
		\begin{flushleft}
			\hspace{0.6cm}SSIM
			\hspace{0.8cm}0.5182
			\hspace{0.9cm}0.6674
			\hspace{0.8cm}0.7057
			\hspace{0.9cm}0.7117
			\hspace{0.9cm}0.7374
			\hspace{0.9cm}0.7380
			\hspace{0.9cm}0.7560
			\hspace{0.9cm}1.0000				
		\end{flushleft}		
	}
	\caption{
		The SmoothGrad \cite{smilkov2017smoothgrad} explanations (b--g) for a prediction by ResNet are becoming increasingly similar to the explanation for a different prediction by a ResNet-R as we increase $N_{SG}$---a hyperparameter that governs the smoothness of SG explanations.
		Similarly, under GuidedBackprop (GB) \cite{springenberg2014striving}, the explanation appears substantially closer to that of a different model (h vs. i) compared the original heatmaps (b vs. i).
		Below each heatmap is the SSIM similarity score between that heatmap and the ResNet-R heatmap (i).
		See more examples in Fig.~\ref{fig:appendix_guided_top10}.
		%
		%		Modifying gradients using SG or GuidedBackprop changes the interpretation of model. 
		%		Qualitative trend showing the increase in similarity between SmoothGrad (SG) of ResNet (c-g) and gradient of ResNet-R (i) on increasing the number of samples, $N_{SG}$.
		%		GuidedBackprop \cite{springenberg2014striving} attribution map of the \class{hammerhead ~shark} image using ResNet is shown in (h).
		%		Starting from the gradient of ResNet (b) we observe an increase in SSIM scores with increasing $N_{SG}$.
		%		Across the dataset, the \class{hammerhead~shark} image had the maximum SSIM score between the attribution map of SG, using 800 samples (g), of ResNet and gradient of the ResNet-R (i).
		%		Interestingly, the GuidedBackprop attribution map is most similar (SSIM=0.7560) to the gradient of ResNet-R.
		%		0.239, 0.334, 0.377
		%			\todo{change the selecting criteria to top-1} \todo{add SSIM bottom and number of samples above}
		%			\anh{If time permits (backlog), add $SG_{800}$ for GuidedBackProp}
	}
	\label{fig:SG_trend_qual}
	\vspace*{-0.4cm}	
\end{figure*}

\section{Experiment framework}
%	\label{sec:method}
%	\subsection{Datasets and Network}
	\label{sec:dataset_network}

	\paragraph{Explanation evaluation metrics} 
	Currently, there is not yet a common ground-truth dataset for evaluating the accuracy of attribution methods \cite{doshi2017towards}.
	However, researchers often approximate explanation correctness via two main techniques: (1) object localization \cite{zhou2016learning}; 
	and (2) Insertion \& Deletion \cite{Petsiuk2018rise}.
	The \textbf{localization error} measures how accurately an attribution map localizes the main object in the input image \cite{zhou2016learning}---a reasonable approximation for the ImageNet images \cite{russakovsky2015imagenet}, which are object-centric and paired with human-labeled segmentation masks.
	We did not use evaluation metrics like Pointing Game accuracy \cite{zhang2018top} and Saliency Metric \cite{dabkowski2017real} as they are derivatives of the localization task.
	The \textbf{Deletion} metric \cite{Petsiuk2018rise} measures the classification score changes as we gradually zero out the input pixels in the descending order of their attributions.
%    \naman{gradually zero-out the input pixels in the decreasing order of attribution values. one-by-one is not fixed (another hyperparameter)}
	The idea is if the attribution values correctly reflect the discriminative power of the input pixels, knocking out the highest-attribution pixels should quickly cause the probability to approach zero.
	In contrast, \textbf{Insertion} \cite{Petsiuk2018rise} tests whether inserting the highest-attribution pixels into a zero image would quickly increase the probability.
	We used all three above mentioned metrics\footnote{We used the Insertion and Deletion code by the authors \cite{Petsiuk2018rise}.} to quantify how much the variation of explanations translates into the sensitivity of their accuracy (Sec.~\ref{sec:eval_sensitivity}).
	
	\subsec{Classifiers}
	All of our experiments were conducted on two groups of classifiers:
	(a) GoogLeNet \cite{szegedy2015going} \& ResNet-50 \cite{he2016deep} (hereafter, ResNet)  pre-trained on the 1000-class 2012 ImageNet dataset \cite{russakovsky2015imagenet};
	and (b) the robust versions of them \ie GoogLeNet-R \& ResNet-R that were trained to also be invariant to small adversarial changes in the input image \cite{engstrom2019adversarial}.
	We obtained the two regular models from the PyTorch model zoo \cite{torchvis88:online}, the ResNet-R from \cite{engstrom2019adversarial}, and we trained GoogLeNet-R by ourselves using the code released by \cite{engstrom2019adversarial}.
	While the two robust classifiers are more invariant to pixel-wise noise they have lower ImageNet validation-set accuracy scores (50.94\% and 56.25\%) than those of the original GoogLeNet \& ResNet (68.86\% and 75.59\%).

	\subsec{Datasets}
	From the 50,000 ImageNet validation-set images, we randomly sampled a set of $1735$ images that all four models correctly classify.
	We used this set of images in all experiments throughout the paper.

%	Our experiments were conducted on GoogLeNet \cite{szegedy2015going} and ResNet-50 \cite{he2016deep} image classifiers that were pre-trained on the 1000-class 2012 ImageNet dataset \cite{russakovsky2015imagenet}. We also used the robust versions of these models, GoogLeNet-R \& ResNet-50-R \cite{engstrom2019adversarial}, that were adversarially trained using Projected Gradient Descent \cite{madry2017towards} adversarial attack. Going forward, we denote ResNet-50 and ResNet-50-R as ResNet and ResNet-R respectively. Also, we use the term regular models models for ResNet \& GoogLeNet modaels and robust models for their counterparts. For GoogLeNet and ResNet, we used the models officially released by PyTorch model zoo \cite{torchvis88:online} and for ResNet-R we used the model released by the authors \cite{robustness}. We trained GoogLeNet-R, the robust version of GoogLeNet, using the training code provided by the authors \cite{robustness}. See Sec.~\ref{sec:googLeNet_training} for training details.\\

	\subsec{Similarity metrics}
	To quantify the sensitivity of attribution maps, we followed Adebayo et al. \cite{adebayo2018sanity} and used three measures\footnote{We used the implementation by scikit-image \cite{scikit-image}.} that cover a wide range of similarity notions: Spearman rank correlation, Pearson correlation of the histogram of gradients (HOGs), and the structural similarity index (SSIM).
	To quantify the sensitivity of the accuracy scores of explanations, we used the standard deviation (std).

\section{Experiments and Results}
	\label{sec:experiment}
	
\subsection{Gradient maps of robust classifiers are smooth and insensitive to pixel-wise image noise}
\label{sec:robust_classifier}

Gradient saliency maps of image classifiers are (1) notoriously noisy \cite{simonyan2013deep,smilkov2017smoothgrad,bach2015pixel} limiting their utility and (2) sensitive to input changes \cite{alvarez2018robustness}.
Therefore, a number of techniques have been proposed to de-noise the gradient images \cite{shrikumar2016not,smilkov2017smoothgrad,springenberg2014striving,selvaraju2016grad}.
However, are these smoothing techniques necessary for gradients of robust classifiers?

\begin{figure}[h]
	\centering
	{	
		\small
		\begin{flushleft}
			\hspace{0.8cm}(a)
			\hspace{1.2cm}(b)
			\hspace{1.0cm}(c)
			\hspace{1.1cm}(d)
			\hspace{1.0cm}(e)
		\end{flushleft}		
	}
	\vspace{-0.7cm}	
	{	
		\small
		\begin{flushleft}
			\hspace{0.6cm}Input
			\hspace{0.35cm}GoogLeNet
			\hspace{0.01cm}GoogLeNet-R
			\hspace{0.01cm}ResNet
			\hspace{0.2cm}ResNet-R
		\end{flushleft}
	}
	{
		\begin{flushleft}
			\hspace{-0.1cm}\rotatebox{90}{\hspace{-2.2cm}Noisy\hspace{0.7cm}Clean}
		\end{flushleft}
	}
	\vspace{-0.9cm}
	\includegraphics[width=0.45\textwidth]{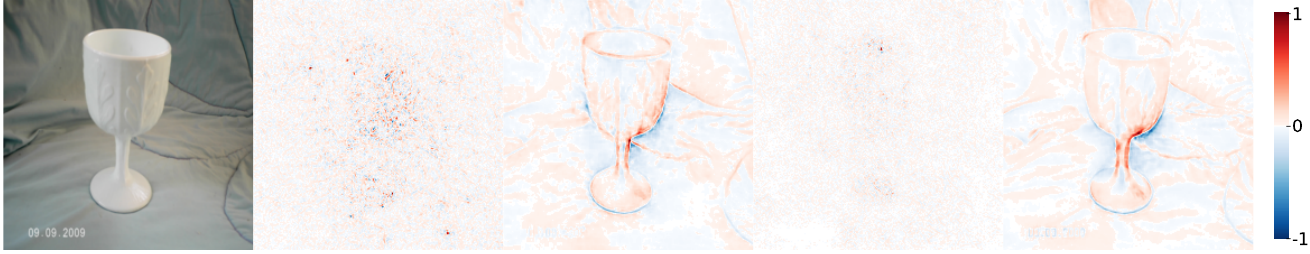}
	\includegraphics[width=0.45\textwidth]{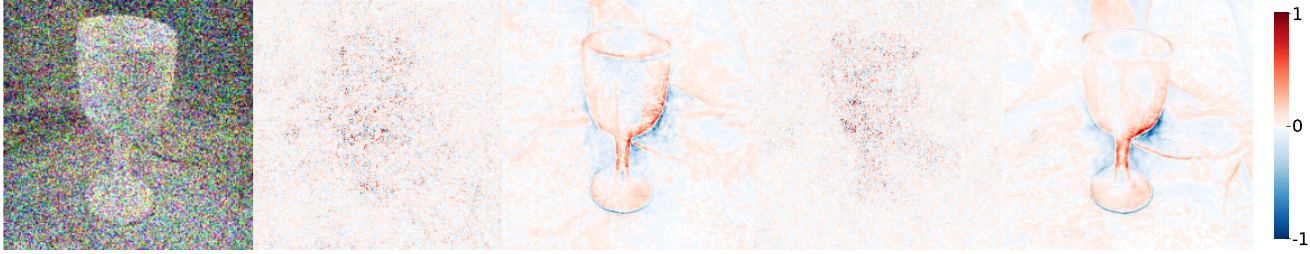}\\
	{	
		\small
		\vspace*{-0.15cm}
		\begin{flushleft}
			\hspace{0.6cm}SSIM:
			\hspace{0.7cm}0.2163
			\hspace{0.4cm}\textbf{0.7372}
			\hspace{0.5cm}0.4740
			\hspace{0.65cm}\textbf{0.8067}
		\end{flushleft}		
	}
	\caption{
		\textbf{Top:} The gradients of robust classifiers (c \& e) reflect the structure of the goblet in an example input image (a), which is in stark contrast to the commonly reported noisy gradients of regular classifiers (b \& d).
		\textbf{Bottom:} The gradients of robust classifiers remain similar before and after the addition of noise to the input image (c \& e---higher SSIM scores).
		An SSIM similarity score is for the two images in each column.
		%        \naman{We need to say this is Top-1 image, no?}
		%		Gradients for regular models, (b) \& (d), and their robust counterparts, (c) \& (e), for clean (row 1) and noisy (row 2) \class{goblet} image. 
		%		Adding noise, $\mathcal{N}(0,\,0.1)$, to the \class{goblet} image distorts the gradients (lower SSIM score) of regular models (b,d) but has little effect (higher SSIM score) on the gradients of robust models (c,e).
		%			The SSIM score between the gradients of clean and noisy images for GoogLeNet (b) and ResNet (d) are $0.2163$ and $0.4740$ whereas they are $0.7372$ ($\sim240\%$ increase) and $0.8067$ ($\sim70\%$ increase) for GoogLeNet-R (c) and ResNet-R (e) respectively.
	}
	\label{fig:grad_sec_1_1_qual}
	\vspace*{-0.4cm}	
\end{figure}

First, we observed, for the first time, that the vanilla gradients of robust classifiers consistently exhibit visible structures (see the outline of the goblet in Fig.~\ref{fig:grad_sec_1_1_qual}c \& e), which is surprising!
They are in stark contrast to the noisy gradients of regular classifiers (Fig.~\ref{fig:grad_sec_1_1_qual}b \& d).

Second, we found that the gradient explanations of robust classifiers are significantly more invariant to a large amount of random noise added to the input image.
Specifically, for each image $\vx$ in the dataset, we added noise $\sim\gN(0,\,0.1)$ to generate a noisy version $\vx_n$ (Fig.~\ref{fig:grad_sec_1_1_qual}; bottom) and measured the similarity between the saliency maps for the pair ($\vx$, $\vx_n$) using all three similarity metrics described in Sec.~\ref{sec:dataset_network}.
Across all images and all three quantitative metrics, the gradients of robust classifiers are substantially more invariant to noise than their regular counterparts (Figs.~\ref{fig:eval_SSIM}~\&~\ref{fig:grad_sec_1_1_quant}).
For example, the average similarity of the gradient pairs from robust models is $\sim$36$\times$ higher than that of the counterparts under the Spearman rank correlation (Fig.~\ref{fig:grad_sec_1_1_quant}; leftmost bars).
This result interestingly show that the gradients of \emph{robust} models are fairly insensitive to minor pixel-wise image changes---a concern in \cite{kindermans2019reliability,alvarez2018robustness,ghorbani2019interpretation}.

\begin{figure*}[t]
	\centering
	{	
		\small
		\vspace*{-0.15cm}
		\begin{flushleft}
			\hspace{0.3cm}Input image
			\hspace{0.5cm}$b_R$=$5$
			\hspace{0.8cm}$b_R$=$10$
			\hspace{0.7cm}$b_R$=$30$
			\hspace{-6.6cm}\rotatebox{90}{\hspace{-3.9cm}ResNet-R\hspace{1.1cm}ResNet}
			\hspace{8.5cm}Input image
			\hspace{0.2cm}$N_{iter}$=$10$
			\hspace{0.2cm}$N_{iter}$=$150$
			\hspace{0.2cm}$N_{iter}$=$300$
			\hspace{0.2cm}$N_{iter}$=$450$
			\hspace{-8.7cm}\rotatebox{90}{\hspace{-3.8cm}ResNet-R\hspace{1.0cm}ResNet}
		\end{flushleft}
	}
	\vspace{-0.3cm}
	\subcaptionbox{Sensitivity to changes in the blur radius $b_R$\label{fig:MP_blur_qual}}
	[0.4\linewidth]{
		\includegraphics[width=0.4\textwidth]{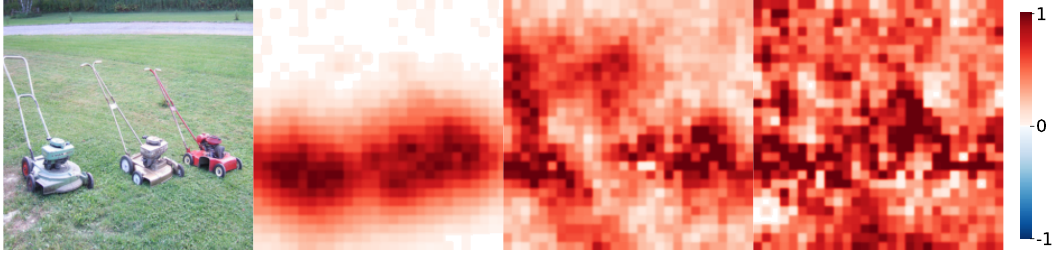}
		\text{SSIM: $0.2669$}
		\includegraphics[width=0.4\textwidth]{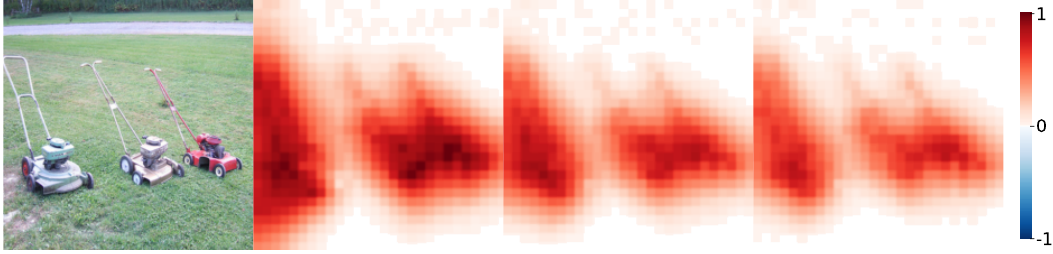}
		\text{SSIM: $0.8493$}
	}
	\hspace{1.0cm}
	\subcaptionbox{Sensitivity to changes in the number of iterations $N_{iter}$\label{fig:MP_iter_qual}}
	[0.5\linewidth]{
		\includegraphics[width=0.5\textwidth]{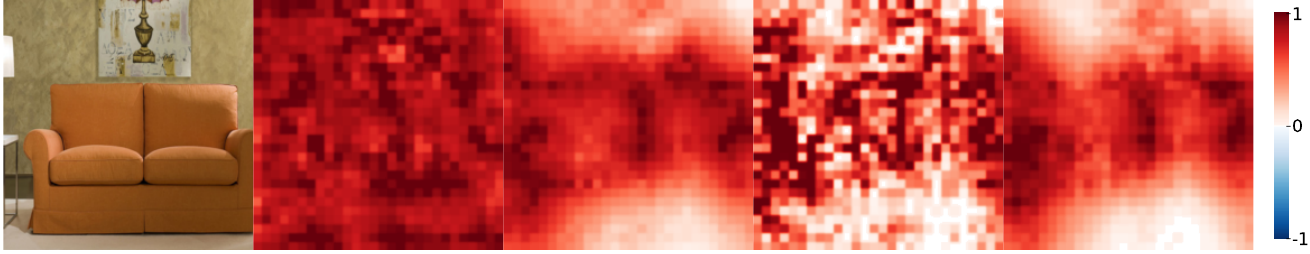}
		\text{SSIM: $0.3536$}		
		\includegraphics[width=0.5\textwidth]{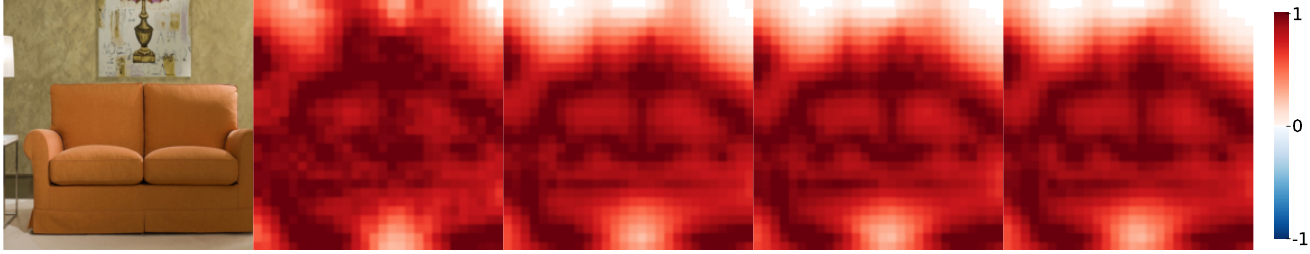}
		\text{SSIM: $0.9313$}		
	}
	\caption{
		%		In MP \cite{fong2017interpretable}, the authors heuristically fix the Gaussian blur radius, $b_R$, and the total number of iterations, $N_{iter}$. 
		MP attribution maps generated for a regular model (ResNet) are highly sensitive to changes (\ie low SSIM scores) in the Gaussian blur radius $b_R$ (a) and in the number of iterations $N_{iter}$ (b).
		In contrast, the same MP explanations for a robust model (ResNet-R) are substantially more stable (see Fig.~\ref{fig:MP_blur_iter_sensitivity} for quantitative results).
		%		Qualitative figures showing the sensitivity of MP attribution maps to Gaussian blur radius $b_R$ (a) and the total number of iterations $N_{iter}$ (b).
		Two reference images in this figure are the top-2 that cause the largest differences between the SSIM scores of ResNet vs. ResNet-R heatmaps.
		%		All the other MP hyperparameters were kept to their respective default values.
		%		
		%		\textbf{(a)}: We observe large variation in the attribution maps for ResNet. 
		%		ResNet heatmaps become more scattered as we increase the Gaussian blur radius.
		%		\textbf{(b)}: 
		%		ResNet-R (row 2) optimize faster and the intermediate masks remain consistent.
		MP being more unstable with ResNet compared to that with ResNet-R can be seen quantitatively in the loss plot (Fig.~\ref{fig:MP_loss}) and qualitatively in the evolution of the MP heatmaps (Fig.~\ref{fig:appendix_MP_iteration}).
		See Fig.~\ref{fig:appendix_MP_blur} for more examples of the blur sensitivity experiments.
		%		See  for the evolution of the respective masks of ResNet and ResNet-R and  for their respective optimization loss curves.
	}
	\label{fig:MP_blur_iter_qual}
	\vspace*{-0.4cm}	
\end{figure*}

%	We then compute the gradient and Input $\times$ Grad heatmaps of both clean and noisy image.
\subsection{De-noising explanations may cause misinterpretation}
	\label{sec:question_smoothing}
% The gradient images from a robust and a regular classifier are different but would appear $\sim$1.5$\times$ more similar, potentially causing mis-interpretation, under several prior methods that attempted to de-noise the original explanations \cite{springenberg2014striving,smilkov2017smoothgrad} (Sec.~\ref{sec:question_smoothing}).

%%%%%%%%%%%%%%%

%%%%%%%%%%%%%%%

We have shown that the vanilla gradients of robust classifiers can be fairly smooth (Sec.~\ref{sec:robust_classifier}).
That result naturally raises a follow-up question: Do the smoothing techniques \cite{smilkov2017smoothgrad,selvaraju2016grad,springenberg2014striving} improve or mislead our interpretation of explanations?
To shed light on that question, we quantify the similarity between (a) the de-noised explanations by SG \cite{smilkov2017smoothgrad} for a regular classifier and (b) the vanilla gradient saliency maps for a robust classifier.

\subsec{Experiment} 
For each image, we generated SG explanations for regular models by sweeping across a range of the sample size $N_{SG}\in\{0, 50, 100, 200, 500, 800\}$.
Here, $N_{SG}$ = 0 yields the vanilla gradient.
We measured the similarity between each SG heatmap of a regular model and the vanilla gradient of a robust counterpart model (\eg ResNet vs. ResNet-R).

\subsec{Results}
We observed that as the sample size $N_{SG}$ increases, the resultant explanations of ResNet become increasingly more similar to the explanation of ResNet-R---a completely different classifier!
That is, the SSIM similarity between two heatmaps increases up to $\sim$1.4$\times$ (Fig.~\ref{fig:SG_trend_qual}; b--g) on average.
This monotonic trend is also observed across three similarity metrics and two pairs of regular vs. robust models (Fig.~\ref{fig:SG_trend_quant}).

Additionally, we generated an explanation using another popular explanation method, GuidedBackprop (GB) \cite{springenberg2014striving}, which \emph{modifies} the gradient by only letting the \emph{positive} forward activations and backward gradients to flow through during backpropagation.
Across the dataset, the average similarity between a pair of (ResNet GB heatmap, ResNet-R gradient heatmap) is 0.377 while the original similarity between the vanilla gradients of two models is only 0.239.

	In sum, our result shows that two explanations from two completely different classifiers (ResNet vs. ResNet-R) may become substantially more similar under explanations techniques (here, SG and GB) that attempt to heuristically de-noise heatmaps, potentially misleading user interpretation.
%    These results reveal that the end-user may wrongly assume the robustness of their model based on these modified gradient explanations.
%    The Pearson correlation of HOG feautres for ResNet increases from $-0.0659$ (at $N_{SG}=0$) to $0.5819$ (at $N_{SG}=800$).
%    Similar trend was observed for other metrics and GoogLeNet model (Fig.~\ref{fig:SG_trend_quant}a).
%    Both SG and IG, intuitively, averages the gradients using multiple input samples. 
	We reached the same conclusion by comparing GI and its approximate version \ie IG \cite{sundararajan2017axiomatic} (see Sec.~\ref{sec:compare_IG_inpGrad}).

\subsection{Gradient-based attribution maps are sensitive to hyperparameters}
\label{sec:sensitivity}

In practice, attribution methods often have various hyperparameters that are either randomly set (\eg a random seed \cite{ribeiro2016should}) or empirically tuned (\eg the number of optimization steps \cite{fong2017interpretable}).
It is important to understand how such choices made by the end-user vary the explanations (Fig.~\ref{fig:teaser}), which impedes reproducibility and can impair users' trust \eg a medical doctor's trust in a model's explanation of its prediction \cite{lipton2017doctor,doshi2017towards}.
Here, we quantify the sensitivity of attribution maps generated by two representative methods (SG and MP) as a common hyperparameter changes.
In all experiments, we compare the average pair-wise similarity between a \emph{reference} heatmap---the explanation generated using the default settings provided by the authors--- and those generated by changing one hyperparameter.

%The resulting explanations can drastically change on varying these hyperparameter values (Fig.~\ref{fig:teaser}).
%Hence, in this section we quantify the sensitivity of different explanation methods on their respective hyperparameter settings.
%Our main aim in this section is to quantify the sensitivity of attribution methods on their respective hyperparameter settings irrespective of the model.
%In all our sensitivity experiments, we choose the attribution map generated using the default hyperparameter setting, provided by the authors, as our baseline. For a given attribution map and hyperparameter setting, this baseline acts as our reference for the computation of similarity metrics (Sec.~\ref{sec:dataset_network}).
\subsubsection{SmoothGrad is sensitive to sample sizes} 
\label{sec:SG_exp}

%\anh{Editing here}
SG was created to combat the issue that gradient images for image classifiers are often too noisy to be human-interpretable---an issue reported in many previous papers \cite{smilkov2017smoothgrad,springenberg2014striving,bach2015pixel,simonyan2013deep} and also shown in Sec.~\ref{sec:robust_classifier}.
While SG does qualitatively sharpen the explanations \cite{smilkov2017smoothgrad} (see Fig.~\ref{fig:SG_trend_qual}b vs. c), the method also introduces two hyperparameters (1) the sample size $N_{SG}$ and (2) the Gaussian std $\sigma$ that were empirically tuned \cite{smilkov2017smoothgrad}.
Here, we test the sensitivity of SG explanations when varying these two hyperparameters.

%Smilkov et al. \cite{smilkov2017smoothgrad} mentioned that they did not notice any qualitative difference in gradients obtained by smoothing above $50$ noisy samples. 
%Using our sensitivity experiments, we shed some quantitative light on the above postulation. 
%Additionally, we also evaluated the sensitivity of SG to the standard deviation (std) $\sigma$ of the Gaussian noise used for creating the noisy samples.

\subsec{Experiment}
%We design two experiments to evaluate the sensitivity of SG across $N_{SG}$ and std, $\sigma$, of Gaussian noise.
To test the sensitivity to sample sizes, we measure the average pair-wise similarity between a reference heatmap at $N_{SG}=50$ (Fig.~\ref{fig:SG_sample_qual}; ii)---\ie the default value in \cite{smilkov2017smoothgrad}---and each of the four heatmaps generated by sweeping across $N_{SG}\in\{100, 200, 500, 800\}$ (Fig.~\ref{fig:SG_sample_qual}; iii---vi) on the same input image.
$\sigma$ is constant at $0.15$.

%First, for each input image, we generated one reference explanation using $N_{SG}=50$ and $\sigma=0.15$ \ie the default setting in \cite{smilkov2017smoothgrad}.
%Then, we generated four more SG explanations by sweeping across $N_{SG}\in\{100, 200, 500, 800\}$ while keeping all other hyperparameters constant.
%Per input image, we compute the similarity between each of the four heatmaps and the reference

%Second, we took different $\sigma$ and fixed $N_{SG}=50$ (See Sec.~\ref{sec:appendix_SG} for more details).
%\naman{see here you have said number of samples, N(SG). It would be better if if use number of samples, N(SG) for the first time N(SG) appears in this paragraph/section}. 
%, i.e. $\sigma\in\{0.1, 0.2, 0.3\}$ to generate three heatmaps

%Considering the case $N_{SG}=50$ as baseline (the default setting in \cite{smilkov2017smoothgrad}), we calculate the average sensitivity scores for a given image using all four possible pairs for the $N_{SG}$ hyperparameter settings.

%We repeated the above experiments measuring the similarity of explanations at 
%In the second experiment, we took different $\sigma$ and fixed $N_{SG}=50$ (details in Sec.~\ref{sec:appendix_SG}).

%\naman{We need to somehow convey that N(SG) = 50 is not a arbitrary choice. It is the default/best setting of the author} 
\subsec{Results}
%SG attribution maps are sensitive to both $N_{SG}$ and $\sigma$ hyperparameters.
We found that the SG explanations for robust models exhibited near-maximum consistency (Fig.~\ref{fig:SG_sample_quant}; all scores are near 1.0).
In contrast, the robustness of SG when running on regular models is consistently lower under all three metrics (Fig.~\ref{fig:SG_sample_quant}; light vs. dark red or light vs. dark green).
SG heatmaps for robust classifiers appear sharper and less noisy compared to those of regular models (Fig.~\ref{fig:SG_sample_qual}; top~vs.~bottom).
Furthermore, while SG heatmaps may appear qualitatively stable (Fig.~\ref{fig:SG_sample_qual}; ii--vi), the actual pixel-wise variations are not. 
For example, the $L_1$ pixel-wise difference between the ResNet heatmaps at the two extreme settings (\ie $N_{SG} = 50$ vs. $800$) is over $5\times$ larger than the difference between the respective ResNet-R explanations (Fig.~\ref{fig:SG_sample_qual}; vii).

In sum, we showed that it is non-trivial how to tune a hyperparameter, here $N_{SG}$, to yield an accurate explanation because the heatmaps vary differently for different classifiers.
Similarly, we further found SG heatmaps to be highly sensitive to changes in the amount of noise \ie Gaussian std $\sigma$ (Sec.~\ref{sec:appendix_SG}) added to the input image.

\subsubsection{Meaningful-Perturbation is sensitive to the number of iterations, the Gaussian blur radius, and the random seed}
\label{sec:MP_exp}

MP \cite{fong2017interpretable} is a representative of a family of methods that attempt to learn an explanation via iterative optimization \cite{wagner2019interpretable, qi2019visualizing, carletti2018understanding, wang2018learning,uzunova2019interpretable,agarwal2019removing}.
%\naman{Should we not site the related papers one more time?}. 
However, in practice, optimization problems are often non-convex and thus the stopping criteria for iterative solvers are heuristically set. For instance, it can be controlled by a pre-defined number of iterations $N_{iter}$.
Also, MP learns to blur the input image to minimize the classification scores and thus depends on the Gaussian blur radius $b_R$.
Here, we test MP sensitivity to three common hyperparameters: $N_{iter}$, $b_R$, and the random seed which governs random initializations.

%An important step in the MP mask optimization is generating the perturbed image using a fixed Gaussian blur of radius $b_R$ (See Sec.~\ref{sec:method_description_implementation_details}). 
%One would expect the attribution maps to be consistent for different $b_R$'s. 
%We found that MP is sensitive to the $\sigma_r$.
%Also, the only stopping criteria for the MP optimization is a heuristically chosen value for the number of iterations and we found that the masks are unstable as the optimization progresses.\\
%In order to quantify the effect of these hyperparameters, we designed the following experiments.
\subsec{Experiment} 
%One can expect that an explanation algorithm to give consistent results across different random initializations. 
%We observed that MP heatmaps change as we simply change the random seed of the mask initialization. 
%An important step in the MP mask optimization is generating the perturbed image using a fixed Gaussian blur width \cite{fong2017interpretable}. Intuitively, the output attribution map from MP should not vary much on changing the Gaussian blur width. 
%
In order to test the sensitivity to the number of iterations, we measure the average similarity between a reference heatmap at $N_{iter}=300$ which is the default seting in \cite{fong2017interpretable} and each of the three heatmaps generated by sweeping across $N_{iter}\in\{10, 150, 450\}$ (Fig.~\ref{fig:MP_iter_qual}) on the same input image.
To measure the sensitivity to the blur radius settings, we repeated a similar comparison to the above for a reference heatmap at $b_R = 10$ and other heatmaps by sweeping across $b_R \in \{5, 30\}$ (Fig.~\ref{fig:MP_blur_qual}).
For other hyperparameters, we used all default settings as in \cite{fong2017interpretable}.

%First, we initialized the MP mask using five random seeds, where seed$~\in\{0, 1, 2, 3, 4\}$, and chose seed$~=0$ as baseline.

%First, we took three different $b_R$, where $b_R \in \{5, 10, 30\}$, for generating the perturbed image with $b_{R}$=$10$ being the baseline.
%Second, we initialized the mask in 3 different ways, i.e. all ones, random and circular initialization, to obtain $k_{2}=3$ heatmaps. 
%Second, we ran MP algorithm for four different maximum number of iterations, $N_{iter}\in\{10, 150, 300, 450\}$, to obtain four heatmaps and chose $300$ number of iterations as the baseline. We then calculate the average similarity metric scores between the two and four attribution map pairs for the two respective hyperparameter settings.

\subsec{Results} 
We found that MP explanations are sensitive to changes in the blur radius but interestingly in opposite ways for two different types of classifiers.
That is, as we increase $b_R$, the heatmaps for ResNet tend to be more noisy and sparse; however, those for ResNet-R become gradually more localized and smoother (Fig.~\ref{fig:MP_blur_qual}; top vs. bottom). 
See Fig.~\ref{fig:appendix_MP_blur} for more examples.
%\anh{Need to check whether this statement is true or not (do we observe the trends across many images?)}
%\todo{Add any hypothesis for why this happened.}
%\chirag{I am just writing informally right now! my hypothesis is that ResNet-R is comparatively more robust to blur noises than ResNet. On plotting the histogram of the prediction probabilities of the blurred images at different $b_{R}$, we observed that ResNet-R has higher prediction probabilities. This means that they are still able to retain some discriminative features, with respect to the target class, for images blurred with larger radius. I think this might be the reason why they can still give us some localized attibution maps as compared to ResNet.}

Across the number of iterations, MP explanations for regular classifiers vary dramatically.
In contrast, the heatmaps for robust models are 1.4$\times$ more consistent under SSIM similarity metrics (Figs.~\ref{fig:MP_iter_qual} \& \ref{fig:appendix_MP_iteration}). 
The MP optimization runs for robust models converged substantially faster within only $\sim$10 steps (compared to the default $N_{iter} = 300$ \cite{fong2017interpretable}) which can be seen in both the loss plot (Fig.~\ref{fig:MP_loss}) and the sequence of heatmaps (Fig.~\ref{fig:appendix_MP_iteration}).
This inconsistent behavior of MP suggests that when comparing MP explanations between these two classifiers, an end-user may draw an entirely different conclusion depending on when optimization stops (which is heuristically chosen).

\subsec{Sensitivity to the random seed}
Our previous experiments followed exactly the setup in \cite{fong2017interpretable} where the authors used a blur circular mask that suppresses the target probability by 99\% as the initial heatmap.
This initialization, however, strongly biases the optimization towards a certain type of explanation.
To avoid that, in practice, MP users randomly initialize the explanation before optimization \cite{chang2019explaining}.
By running experiments similar to the previous ones, we found that MP is also sensitive to the random seed, which controls the random initializations.
That is, on average across 3 similarity metrics, heatmaps for robust classifiers are $1.22\times$ more consistent than those for regular classifiers (see Sec.~\ref{sec:appendix_MP} for more details and Fig.~\ref{fig:MP_seed_sensitivity} for results).

%For MP mask optimization, Fong et al. \cite{fong2017interpretable} used a circular mask initialization that suppresses the score of the target class by $99\%$ when compared to that of using a completely blurred image. We argue that this circular mask acts as an inherent prior to the optimization since ImageNet mostly contains object centric images. Hence, we evaluate the sensitivity of MP attribution maps by initializing masks with different random seeds (corresponding to different mask initialization).
In sum, consistent with SG results (Sec.~\ref{sec:SG_exp}), robust classifiers yield more stable explanations than regular models for the three aforementioned hyperparameters of MP (Fig.~\ref{fig:MP_blur_iter_sensitivity}). 
%\naman{convoluted >-Consistent with SG results, robust models yield stable explanations than the regular models for the three aforementioned hyperparameters of MP}
That is, not only the gradients of robust classifiers are more interpretable but also more invariant to pixel-wise image changes, yielding more robust explanations (Fig.~\ref{fig:MP_iter_qual}). 

\begin{figure*}[]
	\centering
	\subcaptionbox{SSIM\label{fig:eval_SSIM}}
	[0.24\textwidth]{
		\includegraphics[width=0.24\textwidth]{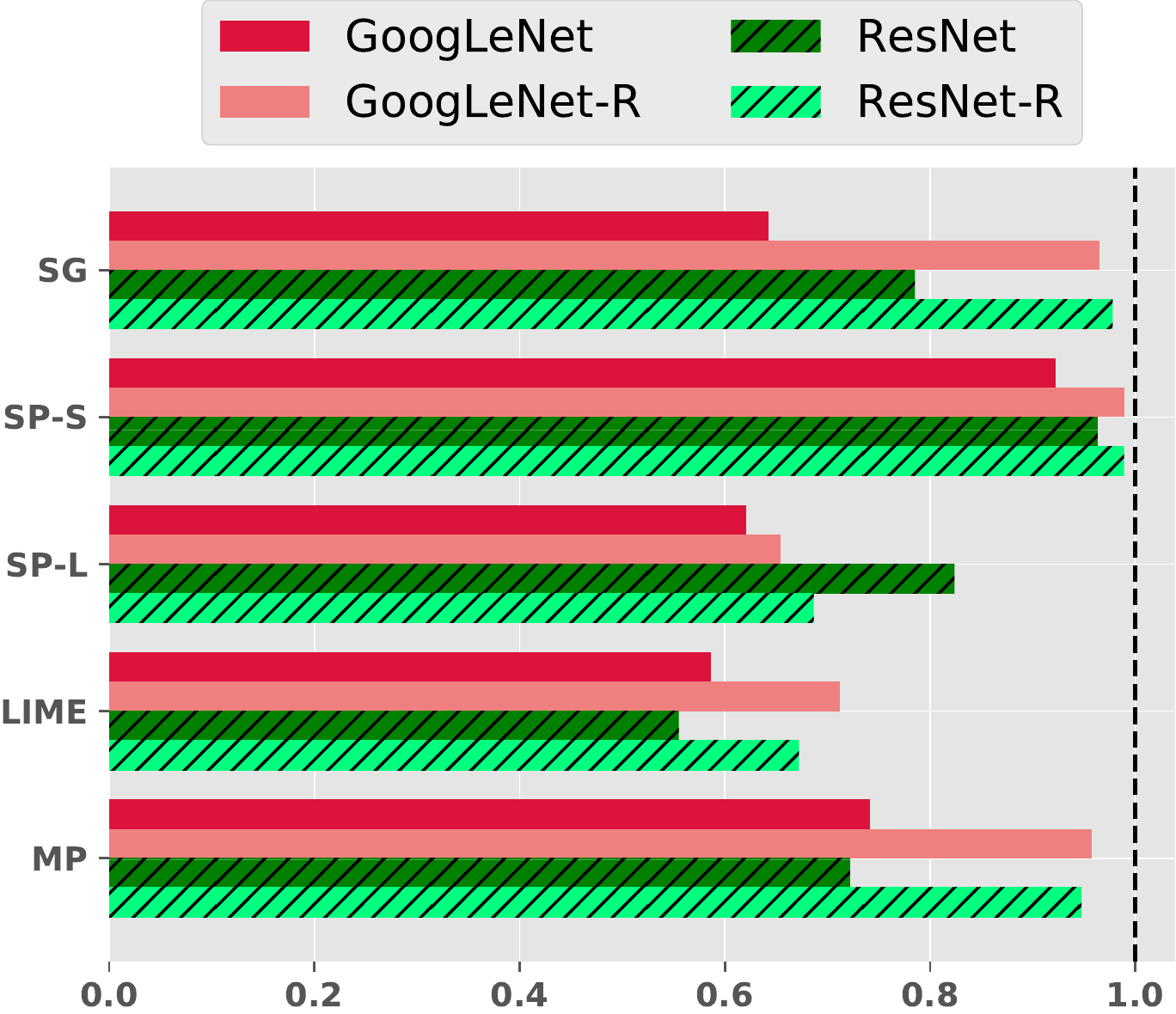}}
	\subcaptionbox{Localization error\label{fig:eval_WSL}}
	[0.25\textwidth]{
		\includegraphics[width=0.25\textwidth]{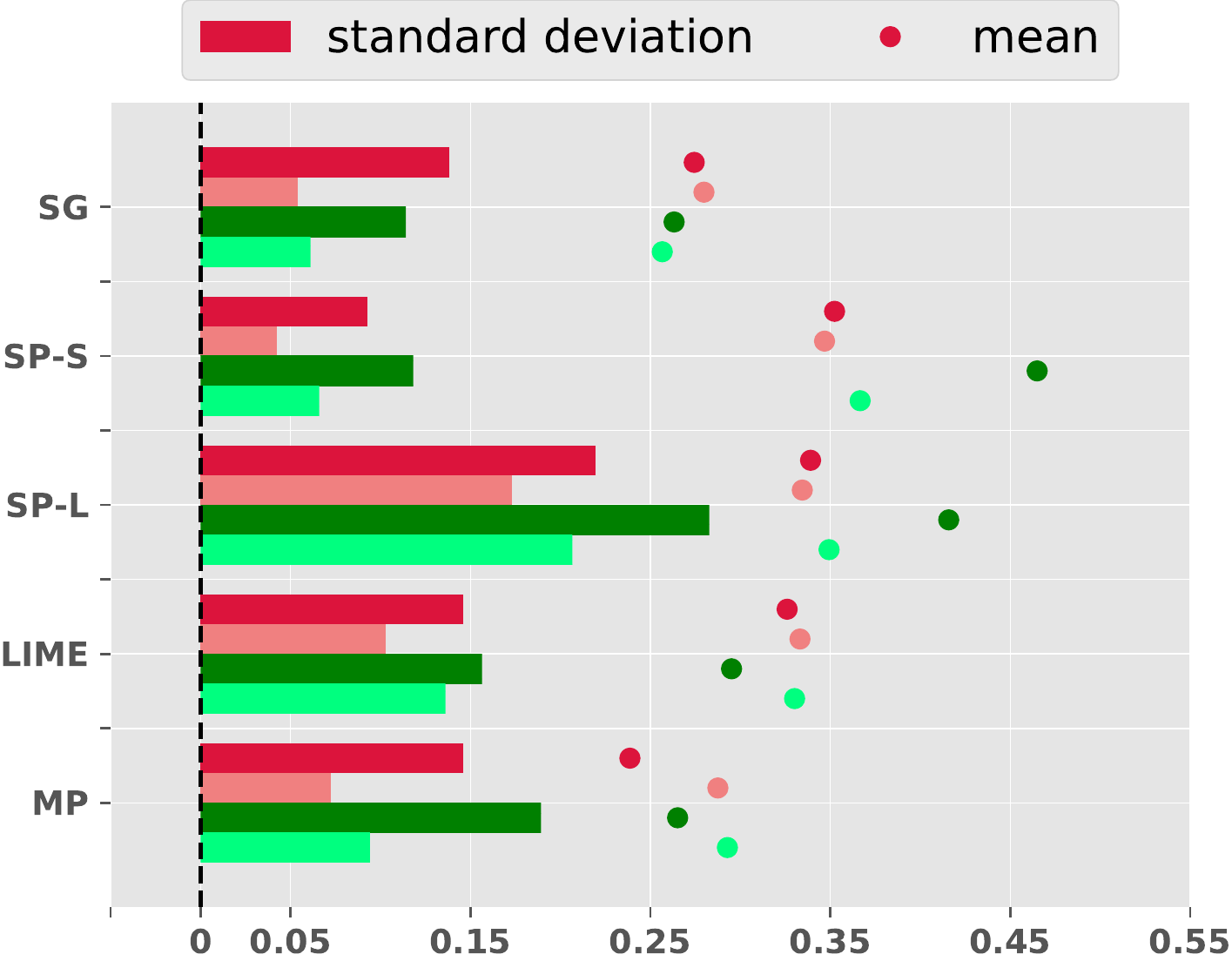}
	}
	\subcaptionbox{Deletion\label{fig:eval_DEL}}
	[0.24\textwidth]{
		\includegraphics[width=0.25\textwidth]{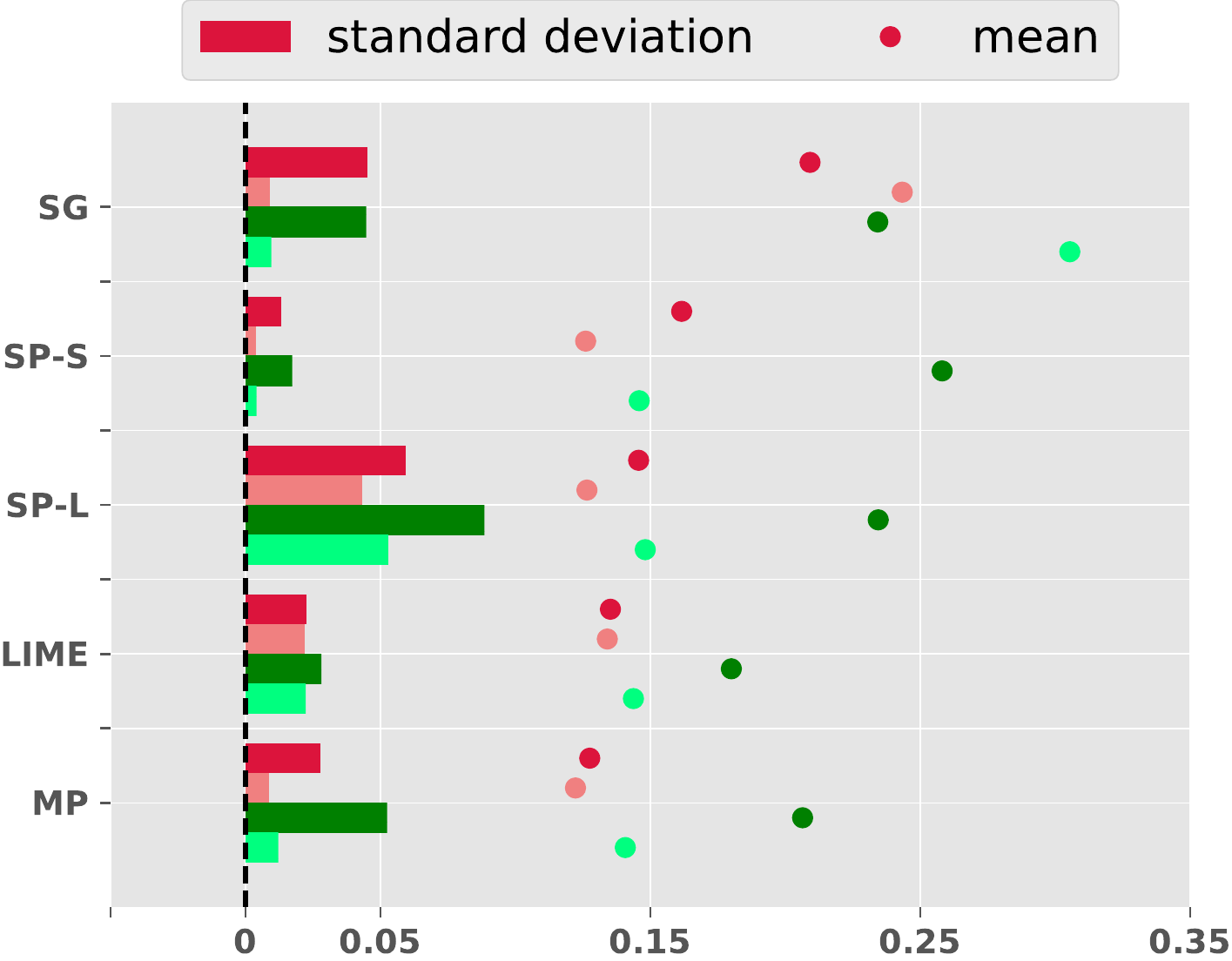}}
	\subcaptionbox{Insertion\label{fig:eval_INS}}
	[0.24\textwidth]{
		\includegraphics[width=0.25\textwidth]{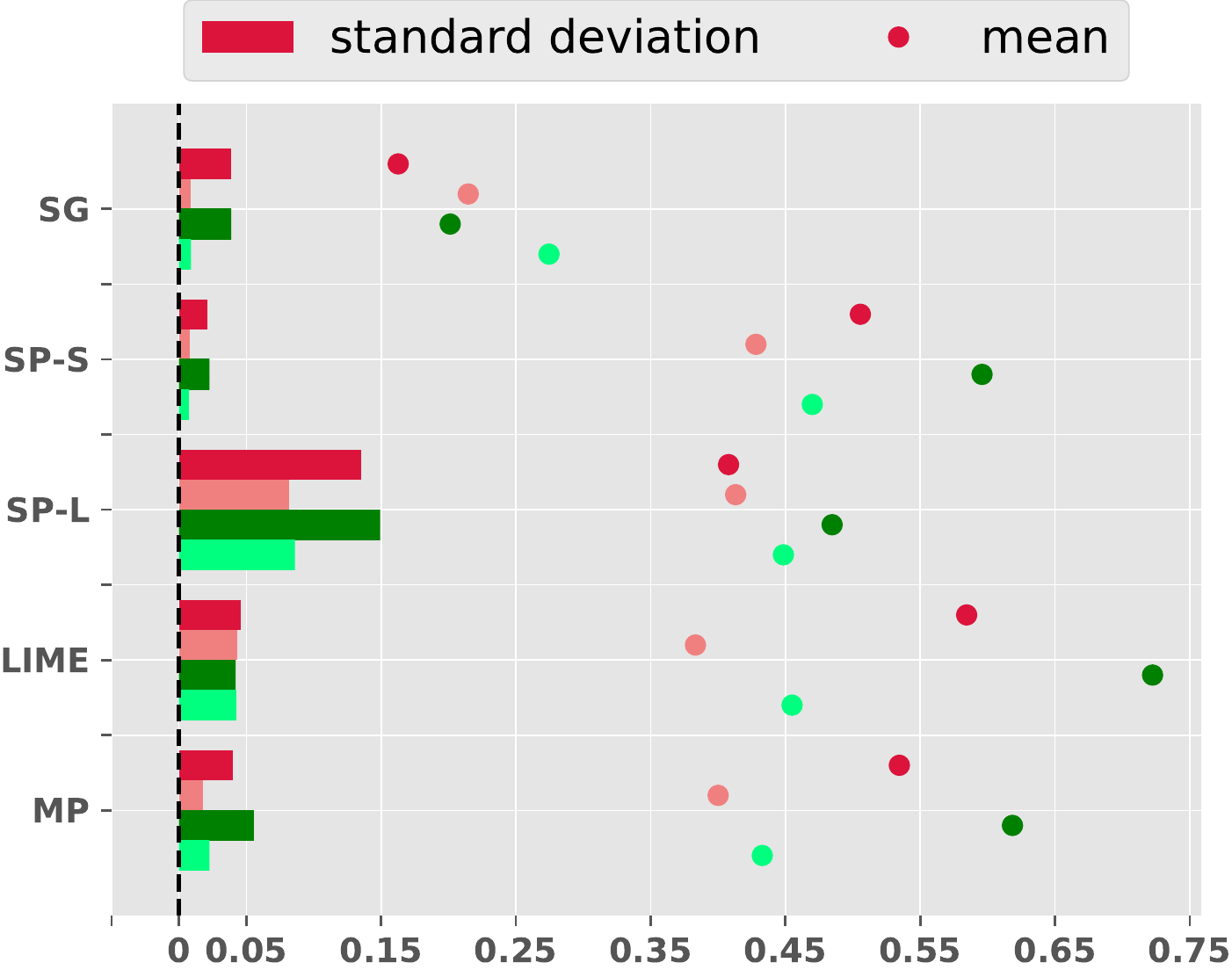}
	}
	\caption{
		Average sensitivity of an individual attribution map measured in the pixel space (a) and three accuracy metric spaces: the Localization error (b), Deletion (c) and Insertion (d) scores (Sec.~\ref{sec:eval_sensitivity}).
		The results were produced by varying the random seed of LIME and MP (bottom two rows), the patch size in SP (SP-S and SP-L), and the sample size of SG (top row).
%		\anh{Explain clearly that}
%		\anh{a} mean SSIM across the dataset
%		\anh{b--d Global mean and average std denoting the average performance and its average variance when changing the hyperparameter settings.}
%		Mean SSIM across the dataset (a) and global mean and average standard deviation (std) values of Localization error (b), Deletion (c) and Insertion (d) scores obtained from evaluating the sensitivity of explanation accuracy to hyperparameters (Sec.~\ref{sec:eval_sensitivity}). 
		SP-S and SP-L are two variants of the SP experiments (Sec.~\ref{sec:SP_exp}). 
%		SP-S represents small difference in patch sizes, $p\in\{52,53,54\}$ and SP-L represents large difference in patch sizes, $p\in\{5, 17, 29, 41, 53\}$.
		For Localization performance of SP-S (b), even a change of $\pm1$px in patch size results in a std of $\sim$10\% for GoogLeNet (dark red) and ResNet (dark green).
%		Higher SSIM scores (\ie less sensitivity among heatmaps across different hyperparameter settings) often also correlate with smaller stds in the accuracy metrics.
		Compared to regular models, robust models (here, GoogLeNet- and ResNet-R) cause the attribution maps to be more consistent pixel-wise under hyperparameter changes---\ie higher SSIM scores (a)---and also more consistent in the three accuracy metrics---\ie lower standard deviations (b--d).
		See Table~\ref{tab:eval_sensitivity} for the exact numbers.
	}
	\label{fig:eval_sensitivity}
%	\vspace*{-0.4cm}
\end{figure*}

\subsection{Non-gradient attribution maps are sensitive to hyperparameters}
\label{sec:non_grad_sensitivity}

\subsubsection{Sliding-Patch is sensitive to the patch size} 
\label{sec:SP_exp}

Sec.~\ref{sec:sensitivity} shows that \emph{gradient}-based explanation methods are sensitive to hyperparameters and their sensitivity depends on the robustness of the gradients with respect to the input changes (Sec.~\ref{sec:MP_exp}).
Here, we test whether methods that are \emph{not gradient-based} would have similar shortcomings.
We chose SP \cite{zeiler2014visualizing} which slides a square patch of size $p\times p$ across the input image and records the classification probability changes into the corresponding cells in the attribution map.
While SP has been widely used \cite{zeiler2014visualizing,MATLAB,ancona2018towards}, it remains unknown how to choose the patch size.

\begin{figure}[H]
	\centering
	{	
		\small
		\begin{flushleft}
			\hspace{0.01cm}Input image
			\hspace{0.01cm}$5 \times 5$
			\hspace{0.2cm}$17 \times 17$
			\hspace{0.2cm}$29 \times 29$
			\hspace{0.3cm}$41 \times 41$
			\hspace{0.1cm}$53 \times 53$
			\hspace{-8.1cm}\rotatebox{90}{\hspace{-3.55cm}zoomed-in\hspace{0.40cm}zoomed-out}			
		\end{flushleft}		
	}
	\vspace*{-0.15cm}
	\includegraphics[width=0.45\textwidth]{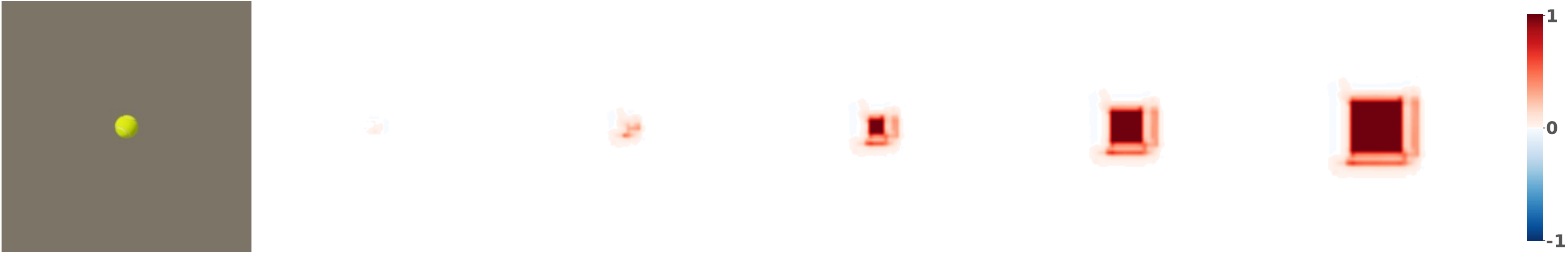}
	{	
		\small
		\vspace*{-0.2cm}
		\begin{flushleft}
			\hspace{0.01cm}\class{tennis~ball}: $0.985$
		\end{flushleft}		
	}
	\vspace*{-0.15cm}
	\includegraphics[width=0.45\textwidth]{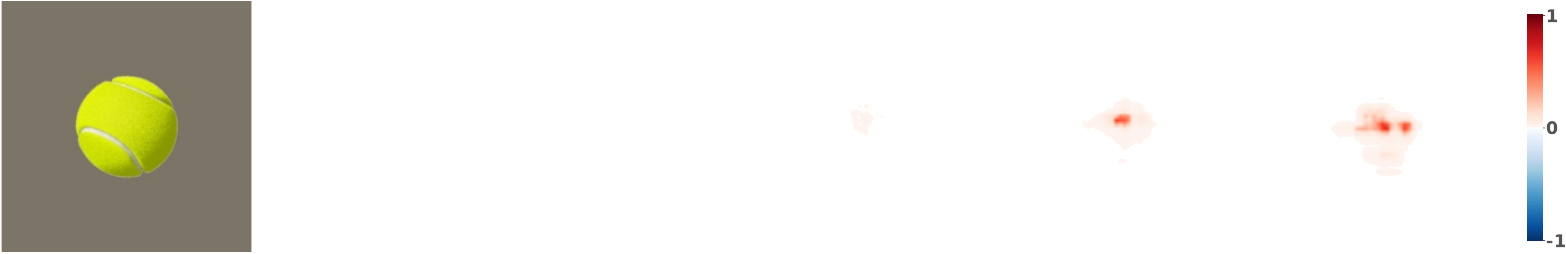}
	{	
		\small
		\vspace*{-0.2cm}
		\begin{flushleft}
			\hspace{0.01cm}\class{tennis~ball}: $0.998$
		\end{flushleft}		
	}	
	\caption{
		SP explanations are sensitive to patch sizes.
		\textbf{Zoomed-out:} SP attribution region (red squares) for a tennis ball of size $19 \times 19$ (rendered on a $224 \times 224$ zero image) grows as the patch size increases.
		\textbf{Zoomed-in:} SP outputs blank heatmaps at patch sizes of $5 \times 5$, $17 \times 17$, and $29 \times 29$, which are much smaller than the size of the tennis ball (here, $84\times84$) in a zoomed-in version of the top image.
		%		
		%		SP attribution maps for a zoomed-in \class{tennis~ball}, of size $84 \times 84$ in a $224 \times 224$ gray image. 
		%		No change in target class probability if the patch size is much smaller than the object size (blank heatmaps for patch sizes, $5 \times 5$ and $17 \times 17$).		
		%		\anremoved{
		%		\textbf{Row 1:} SP attribution maps for a zoomed-out \class{tennis~ball}, of size $19 \times 19$ in a $224 \times 224$ gray image.
		%		The attribution map enlarges as we increase the patch size.
		%		\textbf{Row 2:} SP attribution maps for a zoomed-in \class{tennis~ball}, of size $84 \times 84$ in a $224 \times 224$ gray image. 
		%		No change in target class probability if the patch size is much smaller than the object size (blank heatmaps for patch sizes, $5 \times 5$ and $17 \times 17$).		
		%		}
	}
	\label{fig:SP_patch}
%	\vspace*{-0.4cm}	
\end{figure}

To understand \textbf{the relation between SP patch size and the size of the object} in an input image, we generated two images, each containing a tennis ball of size $19\times19$ or $84\times84$ on a zero background of size $224\times224$ (Fig.~\ref{fig:SP_patch}).
We ran SP on these two images sweeping across 5 patch sizes of $p\times p$ where $p\in\{5, 17, 29, 41, 53\}$.
We observed that the heatmaps tend to be blank when the patch size is much smaller than the object size (Fig.~\ref{fig:SP_patch}; zoomed-in) because the occlusion patch is too small to substantially change the classification score.
In contrast, if the patch size is much larger than the object size (Fig.~\ref{fig:SP_patch}; zoomed-out), the attribution areas tend to be exaggerated \ie even larger than the object size (Fig.~\ref{fig:SP_patch}; the size of the red square increases from left to right).
Therefore, SP explanations are subject to errors as the size of the object in the image is unknown.

\subsec{Sensitivity to large changes} 
To quantify the sensitivity of SP explanations to the patch size, here, we measure the average similarity between a reference SP attribution map at $p = 29$ and each of the four attribution maps generated by sweeping across $p\in\{5, 17, 41, 53\}$ on the same input image. 
This set of patch sizes covers a large range of settings (hence, denoted by SP-L) used in the literature \cite{ancona2018towards,zeiler2014visualizing,MATLAB}.
We kept the stride constant at $3$.
We observed that across all classifiers, SP is highly sensitive to changes within the SP-L set.
In contrast to the case of gradient-based methods, SP explanations for robust classifiers are not significantly more consistent than those for regular models (Fig.~\ref{fig:SP_sensitivity}).
Compared to other methods, SP sensitivity to patch sizes is higher than the sensitivity of SG and MP
(Fig.~\ref{fig:eval_SSIM}; SP-L bars are the shortest on average).
See Fig.~\ref{fig:appendix_SP} for more examples on sensitivity to large changes in patch size.

\subsec{Sensitivity to small changes} We further repeated the previous experiment but comparing the similarity of SP explanations at $p = 53$ with those generated  at $p \in \{52, 54\}$ i.e. a small range (hence, denoted by SP-S).
We observed that SP explanations are not 100\% consistent even when the patch dimension changes within only $\pm1$px (Fig.~\ref{fig:eval_SSIM}; SSIM scores for SP-S are $< 1.0$).
\vspace{-0.2cm}
\subsubsection{LIME is sensitive to random seeds and sample sizes}
\label{sec:LIME_exp}

LIME \cite{ribeiro2016should} is a black-box explanation method.
Instead of masking out a single square patch (as in SP), which can yield the ``square artifact'' (Fig.~\ref{fig:SP_patch}; zoomed-out), LIME masks out a finite set of random \emph{superpixels}.

Our experiments show that LIME is highly sensitive to its two common hyperparameters.
First, LIME attribution maps interestingly often change as the random seed (which controls the random sampling of superpixel masks) changes!
Second, LIME is also sensitive to the changes in the number of perturbation samples.
See Sec.~\ref{sec:appendix_LIME} for more details.
Aligned with the results with SP (Sec.~\ref{sec:SP_exp}), here, we did not find robust classifiers to yield more stable LIME heatmaps than regular classifiers consistently under all three similarity metrics.
An explanation is that GoogLeNet-R and ResNet-R are robust to pixel-wise changes but not patch-wise or superpixel-wise changes (as done by SP and LIME) in the input image.
See Fig.~\ref{fig:worst_case} for a list of the most sensitive cases across all the LIME sensitivity experiments.

\begin{figure*}[t]
	\centering
	\subcaptionbox{Localization error\label{fig:v3_wsl}}
	[0.31\textwidth]{
		\includegraphics[width=0.29\textwidth]{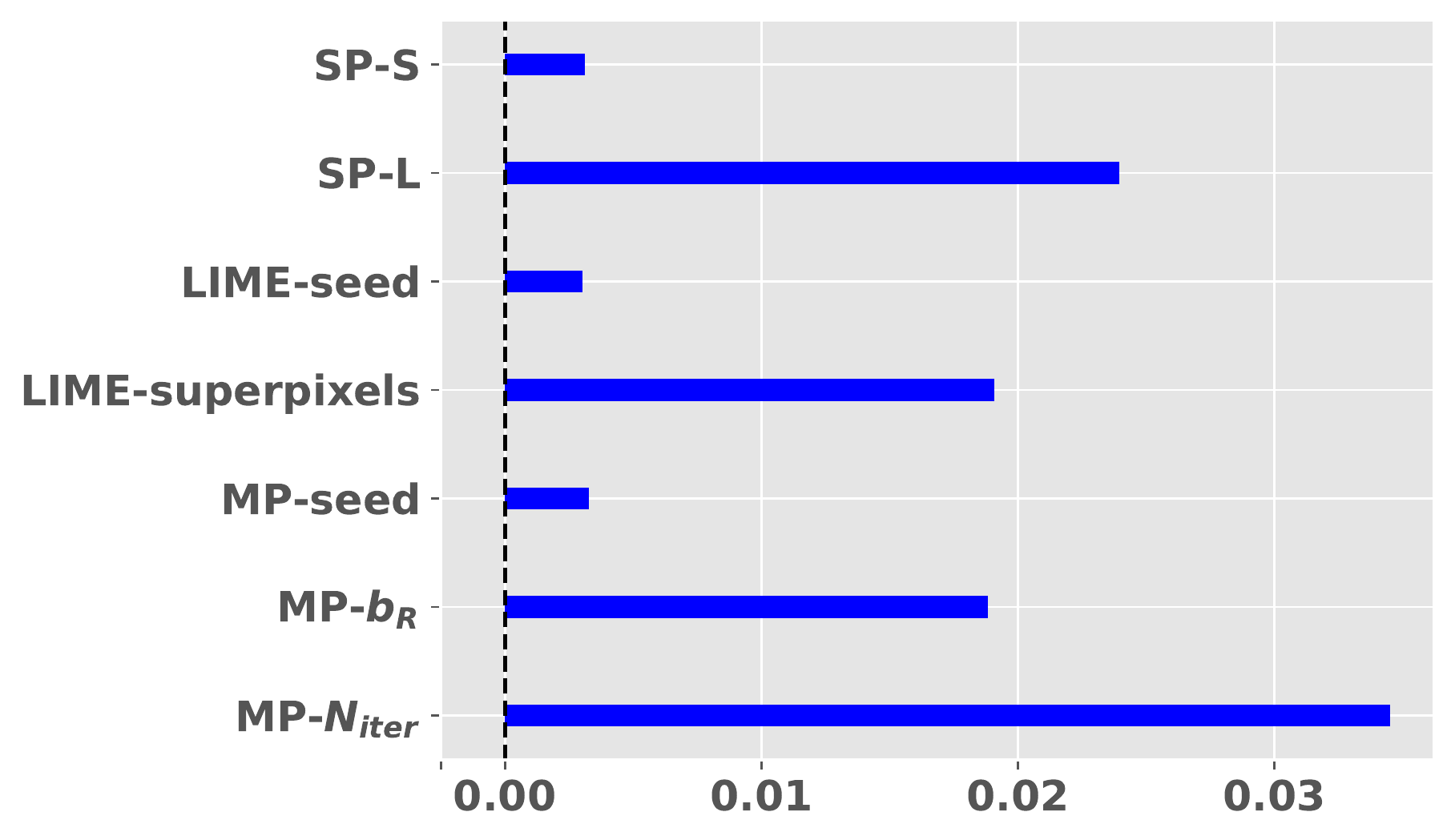}
	}	
	\subcaptionbox{Deletion\label{fig:v3_Deletion}}
	[0.31\textwidth]{
		\includegraphics[width=0.3\textwidth]{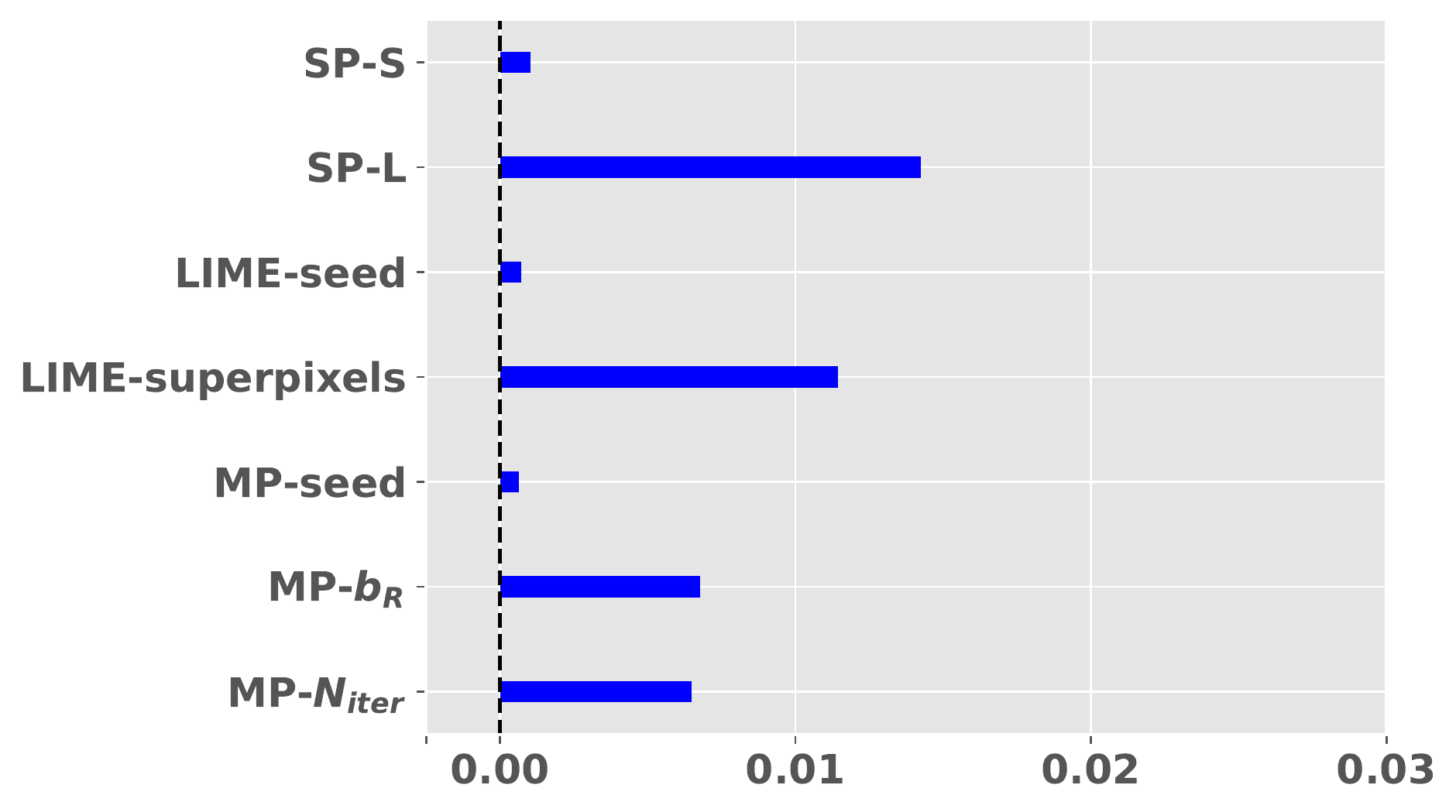}
	}
	\subcaptionbox{Insertion\label{fig:v3_insertion}}
	[0.3\textwidth]{
		\includegraphics[width=0.3\textwidth]{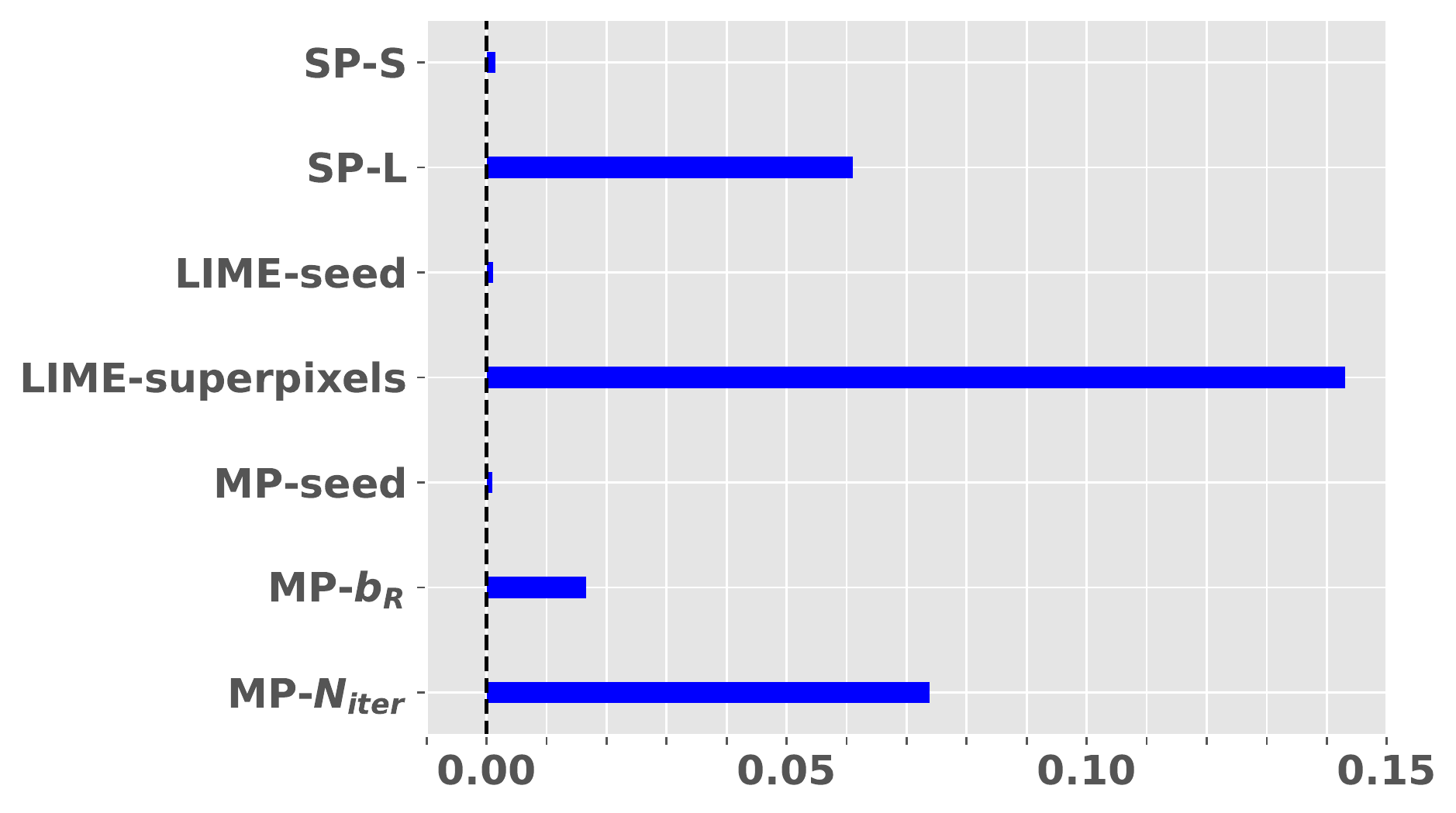}
	}
	\caption{
		Comparisons of the variation in three accuracy scores of attribution methods when 	changing different hyperparameters.
		Here, the horizontal bars show standard deviations (std) for the Localization error (a), Deletion (b) and Insertion (c) scores obtained by marginalizing over all images and classifiers (see Sec.~\ref{sec:eval_sens_hyper}).
		Changing the number of superpixels (in LIME) and the number of iterations (in MP) causes the largest sensitivity to the accuracy of the two methods, respectively.
		%		We observe higher deviation in the Insertion metric (c) as compared to Localization error (a) and Deletion metric (b). 
		%		Note that, the deviations in (a) and (b) may appear small but they are significant, \eg a $0.03$ deviation in Localization errors for MP-$N_{iter}$ (a) represents 3\% variation in localization (\ie ~450 images were wrongly localized due to the hyperparameter changes \naman{how come it is 450. It should be 3*1750/100, no?}).
	}
	\label{fig:v3_sensitivity}
	\vspace*{-0.2cm}
\end{figure*}

\subsection{How do the accuracy scores of an explanation vary when a hyperparameter changes?}
%\subsection{Sensitivity of the correctness of an individual attribution map to hyperparameter changes}
\label{sec:eval_sensitivity}

In Sec.~\ref{sec:sensitivity} and Sec.~\ref{sec:non_grad_sensitivity}, we have shown that many attribution methods are highly sensitive to changes in their common hyperparameters.
For example, under SSIM, the average explanation consistency is often far from the maximum (Fig.~\ref{fig:eval_SSIM}; GoogLeNet and ResNet scores are far below 1.0).
However, there is still a need to quantify how the variation in pixel-wise heatmaps translates into the variation in accuracy scores.
That is, two heatmaps that are different pixel-wise may have the same accuracy score.
%Because there may be \emph{multiple} similarly accurate explanations for the same prediction, it is unclear whether an attribution map remains accurate when some hyperparameter settings change.
Therefore, it is important for users to understand: \emph{How much does the correctness of an explanation varies, on average, when a given hyperparameter changes?}
To answer that, here, we quantify the variance of three explanation accuracy scores (\ie the Localization error, Insertion, and Deletion scores described in Sec.~\ref{sec:dataset_network}) upon varying the \emph{most common} hyperparameters of the considered attribution methods: (1) the sample size in SG (Sec.~\ref{sec:SG_exp}); 
(2) the patch size in SP (Sec.~\ref{sec:SP_exp}; both sweeping across a small range \ie SP-S and a large range \ie SP-L); 
(3) the random seed in LIME (Sec.~\ref{sec:LIME_exp});
and (4) the random seed in MP (Sec.~\ref{sec:MP_exp}). 
%\chirag{Sec.~\ref{sec:MP_exp}? since we added random seed in MP}).

\subsec{Experiment}
For each hyperparameter, we swept across $N$ values to generate the corresponding $N$ explanations for each input image.
Using an accuracy metric, we evaluated each set of $N$ attribution maps per image to produce $N$ accuracy scores.
From the $N$ scores, we then obtained a mean and a std, for each image.
From the per-image means and standard deviations, we then calculated the global mean and average std across the dataset (Fig.~\ref{fig:eval_sensitivity}).
We repeated the same procedure for each accuracy metric and each classifier.

\subsec{Results} 
First, we found that changing the tested hyperparameters (\ie which are the most common) does not only change the explanations (Fig.~\ref{fig:eval_SSIM}; average SSIM scores are under 1.0) but also their three downstream accuracy scores (Fig.~\ref{fig:eval_sensitivity}b--d; the average std bars are above 0).
However, explanation accuracy varies differently between the metrics. 
That is, compared to the mean scores (Fig.~\ref{fig:eval_sensitivity}; circles), the score variation (in std) are higher for object localization (Fig.~\ref{fig:eval_sensitivity}b) and lower for deletion and insertion metrics (Fig.~\ref{fig:eval_sensitivity}c--d).
Notably, the localization scores are highly sensitive---the average stds of regular and robust models are $0.51\times$ and $0.31\times$ of their respective mean accuracy scores.

Second, varying the patch size of SP by only 1px caused a small variation in the explanation (Fig.~\ref{fig:eval_SSIM}; mean SSIM scores are $\approx$ 1 for SP-S) but a large variation in object localization performance (Fig.~\ref{fig:eval_WSL}; for SP-S, the stds are $\sim$10\% of the mean statistics).

Third, across all four tested hyperparameters and three accuracy metrics, the correctness of explanations for robust models is on average 2.4$\times$ less variable than that for regular models.
In sum, we found that explanations for robust classifiers are not only more consistent but also more similarly accurate upon varying the common hyperparameters (compared to the darker bars \ie regular classifiers, lighter bars are longer in Fig.~\ref{fig:eval_sensitivity}a and shorter in Fig.~\ref{fig:eval_sensitivity}b--d).

\subsection{Which hyperparameter when changed causes a higher variation in explanation accuracy?}
%\subsection{Sensitivity of the accuracy scores of an attribution method over dataset and all classifiers}
%\chirag{Sensitivity importance of hyperparameters}}
\label{sec:eval_sens_hyper}

In Sec.~\ref{sec:eval_sensitivity}, we show that the accuracy of an \emph{individual} explanation, on average, can vary substantially as we change a hyperparameter.
Here, we ask a different important question: \emph{Which hyperparameter when varied causes a higher variation in explanation accuracy?}
That is, we attempt to compare hyperparameters by computing the marginal effects of changing each hyperparameter to the variation in accuracy scores (when marginalizing over all images and four classifiers).

\subsec{Experiment}
As a common practice in the literature, for each classifier, we computed an accuracy score for each generated explanation and took a mean accuracy score over the entire dataset.
Repeating the computation for $N$ values of each hyperparameter (\eg $N$ random seeds of LIME), we obtained $N$ mean accuracy scores from which we computed an std $s$.
For each hyperparameter, we averaged over $\{s\}_4$ \ie four such stds, each computed for a classifier, yielding one global std, which is used for comparing hyperparameters.
Here, we compare the global stds for different hyperparameters within and between methods (see Fig.~\ref{fig:v3_sensitivity}): 
(1) the patch size in SP (Sec.~\ref{sec:SP_exp}; SP-S and SP-L); 
(2) the random seed and the number of superpixels in LIME (Sec.~\ref{sec:LIME_exp}); 
(3) the random seed, the blur radius, and the number of iterations of MP (Sec.~\ref{sec:MP_exp}).

\subsec{SP results} 
Within SP, we found that varying the patch size across a larger range yields a higher variation in accuracy scores (Fig.~\ref{fig:v3_wsl}; SP-L vs. SP-S).
%\naman{Isn't this related to line 2. Should be after line 2}
%\todo{Compare LIME-seed and MP-seed}

\subsec{LIME results} 
Our results enable quantitatively comparing the effects of changing different hyperparameters.
In LIME, varying the number of superpixels causes far more sensitivity in the correctness of explanations compared to varying the LIME random seed (Fig.~\ref{fig:v3_sensitivity}; row 3 vs.~4).
Specifically, the std of Insertion scores when changing the number of superpixels was 130.5$\times$ higher as compared to the std when changing the random seed (Fig.~\ref{fig:v3_insertion}).

\subsec{MP results} 
In MP, changing the number of optimization iterations causes the largest sensitivity in explanation accuracy (among the three MP hyperparameters).
Precisely, the std of Insertion scores, when changing the blur radius $b_{R}$ and the number of iterations $N_{iter}$, was 16.6$\times$ and 74$\times$ higher than that when changing the random seed (Fig.~\ref{fig:v3_insertion}; bottom three rows).
%Similarly for MP, the std of Insertion score was 16.6$\times$ higher on changing $b_{R}$ and $N_{iter}$ respectively as compared to changing the random seed for the mask initialization.

\subsec{Across methods}
Changing the random seed in LIME vs. in MP (two different methods) interestingly causes a similar variation in all three accuracy metrics (Fig.~\ref{fig:v3_sensitivity}; row 3 vs. 5).

\section{Discussion and Conclusion}
\label{sec:discuss}
%\chirag{Editing here}\\

We present the first thorough study on the sensitivity of attribution methods to changes in their input hyperparameters.
%The aim of our systematic study is to provide interpretability researchers different insights of sensitivity in explanation methods. 
Our findings show that the attribution maps for many gradient-based and perturbation-based interpretability methods can change radically upon changing a hyperparameter, causing their accuracy scores to vary as well.
We propose to evaluate the sensitivity to hyperparameters as an evaluation metric for attribution methods.
It is important to carefully evaluate the pros and cons of interpretability methods with no hyperparameters and those that have.

%In sum, first, the gradient heatmaps from robust classifiers are less sensitive to input perturbations and exhibit clear visible structures as compared to the gradients of regular classifiers.
%Second, the output heatmaps of many attribution methods are sensitive to hyperparameters.
%Third, the sensitivity of the attribution maps also reflects in their evaluation performance.

%\begin{figure*}[]
%	\subcaptionbox{Random seed\label{fig:MP_seed}}
%	[0.32\linewidth]{
%		\includegraphics[width=0.32\textwidth]{Figures/plots_combined/MP/A23/time_15702996786088188_Error_Plot_Combined_Method_MP}}
%	\subcaptionbox{Different mask inits for $28 \times 28$ mask\label{fig:MP_mask_init_28}}
%	[0.32\linewidth]{
%		\includegraphics[width=0.32\textwidth]{Figures/plots_combined/MP/A20/Mask_Size_028/time_1569359273810118_Error_Plot_Combined_Method_MP}}
%%	\subcaptionbox{MP sensitivity across different mask inits for $224 \times 224$ mask\label{fig:MP_mask_init_224}}
%%[1\linewidth]{
%%	\includegraphics[width=0.49\textwidth]{Figures/plots_combined/MP/A20/Mask_Size_224/time_1569359288493401_Error_Plot_Combined_Method_MP}}\\
%	\subcaptionbox{Different number of iterations \label{fig:MP_iteration}}
%	[0.32\linewidth]{
%		\includegraphics[width=0.32\textwidth]{Figures/plots_combined/MP/A21/time_15693593094659355_Error_Plot_Combined_Method_MP}}
%	\caption{MP sensitivity}
%	\label{fig:MP_sensitivity}
%\end{figure*}

\clearpage
{\small
\bibliographystyle{ieee_fullname}
\bibliography{references}
}

\newpage
%%%%%%%%%%%% Supplementary Information %%%%%%%%%%%%
% Supplementary Information hacks from http://jshodges.com/index.php?qs=kb_001

%%%%%%%%%%%%%%%%%%%%%%%%%%%%%%%%%%%%%%%%%%%%
%%%%%%%%%%%%%%%%%%%%%%%%%%%%%%%%%%%%%%%%%%%%
%%%%%%%%%%%%%%%%%%%%%%%%%%%%%%%%%%%%%%%%%%%%
% THIS ISSUE WAS IN ANH'S COMMAND
% Issue: Resetting counters breaks the refs to the sections/figures in SI.

% IT HAS BEEN SOLVED BY INCLUDING THE 
% \appendix COMMAND AND REORDERING THE OTHER COMMANDS
% PLEASE DON'T CHNAGE ANYTHING. 
% USE AS IS (UNLESS YOU FACE SOME OTHER ERROR)

\newcommand{\beginsupplementary}{%
    \setcounter{table}{0}
    \renewcommand{\thetable}{S\arabic{table}}%
    \setcounter{figure}{0}
    \renewcommand{\thefigure}{S\arabic{figure}}%
    \setcounter{section}{0}
}

\beginsupplementary%
\appendix

% For section headers starting with S
\renewcommand{\thesection}{S\arabic{section}}
\renewcommand{\thesubsection}{\thesection.\arabic{subsection}}

% rules for title box at top of first page
\newcommand{\toptitlebar}{
    \hrule height 4pt
    \vskip 0.25in
    \vskip -\parskip%
}
\newcommand{\bottomtitlebar}{
    \vskip 0.29in
    \vskip -\parskip%
    \hrule height 1pt
    \vskip 0.09in%
}

\newcommand{\suptitle}{Supplementary materials for:\\\papertitle}

% Make a 1-column title
\newcommand{\maketitlesupp}{
    \newpage
    \onecolumn%[
%        \begin{@twocolumnfalse}
        \null%x
        \vskip .375in
        \begin{center}
            {\Large \bf \suptitle\par}
            % additional two empty lines at the end of the title
            \vspace*{24pt}
            {
                \large
                \lineskip=.5em
                %			\begin{tabular}[t]{c}
                %				\ifcvprfinal\@author\else Anonymous CVPR submission\\
                %				\vspace*{1pt}\\%This space will need to be here in the final copy, so don't squeeze it out for the review copy.
                %                Paper ID \cvprPaperID
                %			\end{tabular}
                \par
            }
            % additional small space at the end of the author name
            \vskip .5em
            % additional empty line at the end of the title block
            \vspace*{12pt}
        \end{center}
%    \end{@twocolumnfalse}
%    ]
}

\maketitlesupp%

\section{Method Description and Implementation Details}
\label{sec:method_description_implementation_details}
 We now provide a detailed description of the interpretability methods that we have used in our proposed experiments.
% As we described in the main text, a deep learning model is a function $F$, mapping a coloured image $x$ of spatial dimension $D \times D$ onto a softmax probability of a target class i.e $F: \mathbb{R}^{D\times D\times 3} \rightarrow \mathbb{R}$. The model $F$ can also be considered as composition of function as $F(x) = Softmax(L(x))$ where $L$ represents the logit score. 
 As described in Sec.~\ref{sec:related_work}, a deep learning model is a function $f$, mapping a coloured image $\vx$ of spatial dimension $d \times d$ onto a softmax probability of a target class, i.e $f: \mathbb{R}^{d\times d\times 3} \rightarrow \mathbb{R}$. 
% The model $F$ can also be considered as composition of function as $F(x) = softmax(L(x))$ where $L$ represents the logit score.
 The model $f$ can also be represented as composition of functions i.e $f(\vx) = softmax(L(x))$, where $L$ represents the logit score.
 An attribution method $A$, maps the model $f$, an image $\vx$ and the respective set of hyperparameters $\mathcal{H}$ to an attribution map $\va \in [-1, 1]^{d \times d}$ \footnote{Following Adebayo et al. \cite{adebayo2018sanity}, we normalized the attribution maps of all explanation methods to the range [-1.0, 1.0] except for SP and MP. The attribution maps for SP and MP, by default, have a fixed range of [-1.0, 1.0] and [0.0, 1.0] respectively. For other explanation methods, the attribution maps were normalized by dividing the heatmaps by the maximum of their absolute values.}, 
 \ie $A(f, \vx, \mathcal{H}) = \va$. 
 The attribution score $a_i \in [-1, 1] $ corresponding to a pixel $x_i$, is an indication of how much a pixel contributes for or against the model prediction, $f(\vx)$, depending on the sign of $a_i$. 
%  \vskip 0.1in
 Most explanation methods, particularly the perturbation-based methods, inadvertently introduce their own hyperparameters, $\mathcal{H}$, but the set $\mathcal{H}$ can be empty for some explanation methods. 
 \\
 \\
 Now we describe different gradient and perturbation-based explanation algorithms used in our experiments.

\begin{itemize}
    \item \textbf{Gradient} - Model gradients for a given image and a target class represent how a small change in input pixels values affects the classification score and thus, serves as a common attribution map. Mathematically, Gradient attribution map, $\va^{Grad}$, is defined as:
    \begin{equation*}
    \va^{Grad} = \frac{\partial L}{\partial \vx}
    \end{equation*}
    
%    As such there is no specified range for gradients. Thus, we normalize it in the range $[-1, 1]$ to represent them as attribution maps. 
    \item \textbf{Gradient $\odot$ Input (GI)} - It is the Hadamard product of the input and the model gradients with respect to the input. Mathematically, GI attribution map, $\va^{GI}$, is defined as:
    \begin{equation*}
    \va^{GI} = \vx \odot \frac{\partial L}{\partial \vx}
    \end{equation*}
%    In this case, the image prior serves the function of smoothing the noisy gradient heatmaps. 
    
%    Again, because of its unbounded nature, we normalize it in the range $[-1, 1]$.

    \item \textbf{Integrated Gradients (IG)} - IG tackles the gradient saturation problem by averaging the gradients over $N_{IG}$ interpolated inputs  derived using input and ``baseline'' image. Here, ``baseline image" is the featureless image for which model prediction is neutral. Mathematically, IG attribution map, $\va^{IG}$ is defined as:
    \begin{equation*}
    \va^{IG} = (\vx - \bm{\bar x}) \times \int_{\alpha=0}^{1} \frac{\partial f(\bm{\bar x} + \alpha \times (\vx - \bm{\bar x}))}{\partial \vx} d\alpha
    \end{equation*}

    where $\bm{\bar x}$ is the baseline image. In practice, the integral above is approximated as follows:
    \begin{equation*}
         \va^{IG} = \frac{1}{N_T} \sum_{j=1}^{N_T}\Bigg((\vx - \bm{\bar x_j}) \times \int_{\alpha=0}^{1} \frac{\partial f(\bm{\bar x_j} + \alpha \times (\vx - \bm{\bar x_j}))}{\partial \vx} d\alpha \Bigg)
    \end{equation*}
    with $N_T$ being the number of trials.\\
    In our experiments, we only consider the number of trials, $N_T$, as a hyperparameter and fix the number of interpolated samples $N_{IG}$ to $100$. Our PyTorch implementation of IG follows the original implementation by the authors \cite{ankurtal85:online}. 
    
    \item \textbf{SmoothGrad (SG)} - To create smooth and potentially robust heatmaps (to input perturbations), SG averages the gradients across a large number of noisy inputs. Mathematically, SG attribution map, $\va^{SG}$, is defined as:
    \begin{equation*}
    \va^{SG} = \frac{1}{N_{SG}} \sum_{n=1}^{N_{SG}} \frac{\partial L(\bm{x+ \epsilon_n})}{\partial \bm{x}}
    \end{equation*}
    
    where $\epsilon$ are i.i.d samples drawn from a Gaussian distribution of mean $\mu$ and std $\sigma$. \\
    In our experiments, we consider two major hyperparameters of SG, namely the std, $\sigma$ and $N_{SG}$ samples. The mean for the i.i.d. samples were fixed to $0$. Our PyTorch implementation of SG follows the original implementation by the authors \cite{PAIRcode82:online}.
    
    \item \textbf{Sliding Patch (SP)} - SP, or Occlusion as it is simply called, is one of the simplest perturbation-based methods where the authors use a gray patch to slide across the image and the change in probability is treated as an attribution value at the corresponding location. 
    Concretely, given a binary mask, $\vm \in \{0, 1\}^{d \times d}$ (with $1$'s for the pixels in the patch and $0$'s otherwise), and a filler image, $\vz$, a perturbed image $\bm{\bar{x}} \in \mathbb{R}^{d \times d\times 3}$ is defined as follows: 
    
    \begin{equation} \label{eq:masking_equation}
	    \bm{\bar x} = \vx \odot (\mathds{1}_{D \times D} - \vm) + \vz \odot \vm
    \end{equation}
    
    where $\vz$ is a zero image or gray image\footnote{In the ImageNet dataset, the mean pixel value is ($0.485, 0.456, 0.406$).} before input-pre-processing. 
%    Thus, the explanation map at the pixel location $i$ SP explanation $E^{SP}$ at a pixel location $i$ is given as 
    Thus, the SP explanation map, $\va^{SP}$, at the pixel location $i$ is defined as:
    \begin{equation*} 
        a^{SP}_i = f(\vx) - f(\bm{\bar x}^i)
    \end{equation*} 
    
    where $\bm{\bar x}^i $ is the corresponding perturbed image generated by setting the patch centre at $i$.
    Due to computational complexity, the square patch (size $p \times p$ where $p \in \mathbb{N}$)) is slid using a stride value of $s$ greater than 1 ($s \in \mathbb{N}$), resulting in an attribution map $\va^{SP} \in \mathbb{R}^{d' \times d'}$ where $d' = \floor{\frac{d-p}{s} + 1}$ with $\floor.$ being the greatest integer.
    We use bilinear upsampling to scale $\va^{SP}$ back to the full image resolution.\\
    In our experiments, we fix the stride $s$ to be 3 and only change the patch side $p$. We implemented SP from scratch using PyTorch based on a MATLAB implementation \cite{MATLAB}.
    
    \item \textbf{LIME} - Similar to SP, it is another perturbation-based method which occludes the input image randomly. 
%    using a set of  $S$ random superpixels. 
    The input image is first segmented into a set of $S$ non-overlapping superpixels. 
    Then it generates $N_{LIME}$ perturbed samples by graying out a random set of superpixels out of all the $2^{S}$ possible combinations, \ie it generates a random superpixel mask $\bm{m'} \in \{0, 1\}^{S}$, to mask out the image as in Eq.~\ref{eq:masking_equation}. 
    For each perturbed sample $\bm{\bar x}^i$, LIME distributes the model prediction $f(\bm{\bar x}^i)$ among the superpixels, inversely weighted by the $L_{2}$ distance of $\bm{\bar x}^i$ from the original image $\vx$.
%     using an exponential kernel. 
     Finally, the weights of the superpixels are averaged over $N_{LIME}$ perturbed samples. 
     The final weight $a_k$ for the $k$\textsuperscript {th} superpixel is assigned to all the pixels in it, thus, resulting in LIME attribution map $\va^{LIME}$.\\
    We use SLIC algorithm \cite{achanta2012slic} for generating the superpixels and consider the number of samples, $N_{LIME}$, number of superpixels $S$ and the random seed as hyperparameters in our experiments.
    All the other parameters are set to their default value as given in the author's implementation \cite{marcotcr75:online}.
    
    \item \textbf{MP} - Instead of perturbing the image with a fixed mask, MP learns the minimal continuous mask, $\vm \in [0, 1]^{d \times d}$, which could maximally minimize the model prediction. 
    MP proposes the following optimization problem: 
    
    \begin{equation*}
        \bm{m}^* = \argmin_\vm \lambda \norm{\vm}_1 + f(\bm{\bar x})
    \end{equation*}

    where the perturbed input, $\bm{\bar x}$, is given by Eq.~\ref{eq:masking_equation} and the filler image, $\vz$, is obtained by blurring $\vx$ with a Gaussian blur of radius $b_R$. 
    In order to avoid the generation of adversarial samples, MP learns a small mask of size $d'' \times d''$
    %d'' is used to avoid repetition with SP%
     which is upsampled to the original image size, $d \times d$, in every optimization step. 
    To learn a robust and smooth mask, the  authors further change the objective function as follows:    
    
    \begin{equation*}
        \vm^* = \argmin_\vm \lambda_1 \norm{\vm}_1 + \lambda_2 TV(\vm) + E_{\tau \sim \mathcal{U}(0, a)}  f(\Phi( \bm{\bar x}, \tau))
    \end{equation*}
    
    where $TV(\vm)$ 
%    = \sum_i \norm{\nabla \vm_i}^3_3, 
    is the TV-norm used to obtain a smooth mask. The third term is the expectation over randomly jittered samples.
%    , $\bm{\bar x}$. 
    The jitter operator $\Phi(.)$ translates the perturbed sample by $\tau$ pixels in both horizontal and vertical direction, where $\tau$ is uniformly sampled from the range $[0, a]$ with $a \in \mathbb{R}$.
    In practice, the above equation is implemented by gradient-descent for a number of iterations $N_{iter}$

    Notably, MP introduces many hyperparameters and the model explanation map, $\va^{MP} = \vm$, learnt by MP is entangled with these hyperparameters. 
    We perform sensitivity experiments with various setting of iterations $N_{iter}$, Gaussian blur radius $b_R$, and the random seed for mask initialization. Our MP implementation in PyTorch is based on the Caffe implementation given by the authors \cite{ruthcfon43:online}. \\
%    \naman{here we have to introduce how come random seed comes into picture for mask init because in the original paper, authors use circular mask.}

\end{itemize}

\section{Adversarial training}
\label{sec:googLeNet_training}
Madry et al. \cite{madry2017towards} proposed training robust classifiers using adversarial training. Engstrom et al. \cite{engstrom2019adversarial} adversarially trained a ResNet-50 model using Projected Gradient Descent (PGD) \cite{madry2017towards} attack with a normalized step size. We followed \cite{engstrom2019adversarial} and trained robust GoogLeNet model, denoted as GoogLeNet-R, for our sensitivity experiments. We used adversarial perturbation in $l_{2}$-norm for generating adversarial samples during training. Additionally, we used $\epsilon=3$, a step size of $0.5$ and the number of steps as $7$ for PGD. The model was trained end-to-end for $90$ epochs using a batch-size of $256$ on 4 Tesla-V100 GPU's. We used SGD optimizer with a learning rate ($lr$) scheduler starting with $lr=0.1$ and dropping the learning rate by $10$ after every $30$ epochs. The standard accuracy for off-the-shelf GoogLeNet model \cite{torchvis88:online} on 50k ImageNet validation dataset was $68.862\%$. Our adversarially trained GoogLeNet-R achieved an accuracy of $50.938\%$ on the same 50k images. 

\section{Similarity between IG heatmaps for regular classifiers and GI heatmaps for robust classifiers}
\label{sec:compare_IG_inpGrad}

IG generates a smooth attribution map by averaging gradients  over a large collection of interpolated inputs. 
Intuitively, both IG and GI are computed using the element-wise product of an input and its respective gradient. 
Hence, similar to Sec.~\ref{sec:question_smoothing}, we evaluate the similarity between the IG of regular models with the GI of robust models.

\subsec{Experiment}  For each image, we generated IG explanations for regular models by sweeping across the number of trials $N_T \in \{0, 10, 50, 100\}$. Here, $N_T = 0$ represents vanilla GI. We computed the similarity between each IG heatmap of a regular model (\eg ResNet) and the vanilla GI of their robust counterparts (\eg ResNet-R).

%We ran IG using different $N_{T}$, where $N_{T}\in\{0, 10, 50, 100\}$, on our dataset.
%Here, $N_{T}=0$ represents the GI of ResNet which we use for comparison. 
%We also generate the GI of the respective images for comparison.

%To test the sensitivity to the number of interpolated samples $N_{T}$, we measure the average pair-wise similarity between a reference heatmap at $N_{T}=0$ and each of the four heatmaps generated by sweeping across $N_{T}\in\{10, 50, 100\}$ on the same input image.
%$N_{T}=0$ represents the GI of ResNet which we use for comparison.
%\naman{This is wrong, confirmed with Chirag. >- For each image, we generated IG explanations for regular models by sweeping across the number of trials $N_T \in \{0, 10, 50, 100\}$. Here, $N_T = 0$ represents vanilla GI. We computed the similarity between each IG heatmap of a regular model and the vanilla  GI of their robust counterparts. -< More or less similar to main text} 

\begin{figure}[H]
	\centering
%	{	
%		\small
%		\vspace*{-0.15cm}
%		\begin{flushleft}
%			\hspace{0.8cm}(a) Real
%			\hspace{2cm}(b) GI
%			\hspace{1.3cm}(c) $N_{T}$=$10$
%			\hspace{1.0cm}(d) $N_{T}$=$50$
%			\hspace{1.1cm}(e) $N_{T}$=$100$
%			\hspace{1.2cm}(f) Input$\times$Grad
%		\end{flushleft}		
%	}
	{	
		\small
		\vspace*{-0.15cm}
		\begin{flushleft}
			\hspace{0.2cm}(a) Input image
			\hspace{0.4cm}(b) IG for ResNet
			\hspace{0.3cm}(c) $N_{T}$=$10$
			\hspace{0.9cm}(d) $N_{T}$=$50$
			\hspace{0.8cm}(e) $N_{T}$=$100$
			\hspace{0.6cm}(f) GB \cite{springenberg2014striving}
			\hspace{0.5cm}(g) GI for ResNet-R
		\end{flushleft}		
	}
	\vspace*{-0.3cm}
	\includegraphics[width=0.99\textwidth]{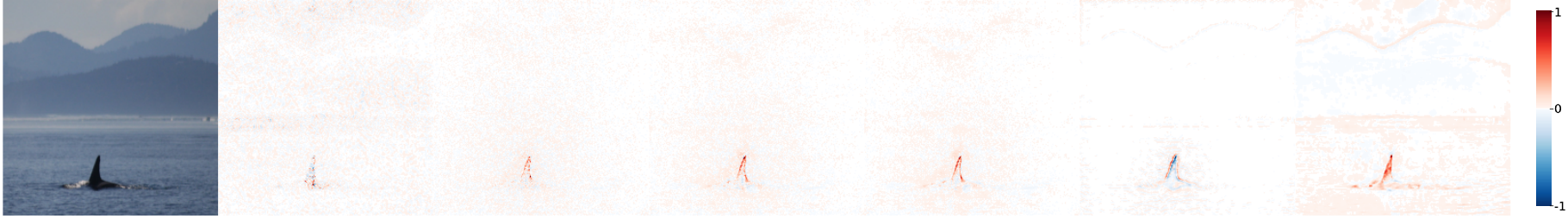}
%	\includegraphics[width=0.99\textwidth]{images/IG_trend_ResNet_ILSVRC2012_val_00025069.pdf}
%	{	
%		\small
%		\vspace*{-0.15cm}
%		\begin{flushleft}
%			\hspace{1.0cm}SSIM:
%			\hspace{1.8cm}0.7909
%			\hspace{1.5cm}0.8648
%			\hspace{1.7cm}0.8714
%			\hspace{1.9cm}0.8765
%			\hspace{2.3cm}1.0
%		\end{flushleft}		
%	}
	{	
		\small
		\vspace*{-0.15cm}
		\begin{flushleft}
			\hspace{1.0cm}SSIM:
			\hspace{1.2cm}0.7909
			\hspace{1.3cm}0.8648
			\hspace{1.4cm}0.8714
			\hspace{1.4cm}0.8765
			\hspace{1.5cm}0.8833			
			\hspace{1.7cm}1.0
		\end{flushleft}		
	}
	\caption{
		The Integrated Gradient (IG) \cite{sundararajan2017axiomatic} explanations (c--e) for a prediction of ResNet are turning into the explanation of a different prediction of a different classifier \ie ResNet-R as we increase $N_{T}$---the hyperparameter that governs the smoothness of IG explanations.
		Similarly, under GuidedBackprop (GB) \cite{springenberg2014striving}, the explanation appears substantially closer to that of a different model (f vs. g) compared the original heatmaps (f vs. b).
		Below each heatmap is the SSIM similarity score between that heatmap and the heatmap in (g).
%		Qualitative trend showing the increase in similarity between GI of ResNet-R (g) and IG of ResNet (c-e) on increasing the number of trials, $N_{T}$.
%		GuidedBackprop attribution map of the \class{killer~whale} image using ResNet is shown in (f).
%		All SSIM scores were calculated with respect to the GI of the robust model (g).
%		Starting from the GI of ResNet (b) we observe the increase in SSIM scores with different number of trials in IG.
%		The \class{killer~whale} image is the case where the SSIM score is maximum between the IG, using $N_{T}=100$ (e), of ResNet and the GI of ResNet-R (g).
%		The GuidedBackprop attribution map (f) is most similar (SSIM=0.8833) to the Input$\times$Grad of ResNet-R.
	}
	\label{fig:IG_trend_qual}
\end{figure}

\begin{figure}[H]
	\centering
	\subcaptionbox{GoogLeNet vs. GoogLeNet-R\label{fig:IG_trend_quan_googLeNet}}
	[0.40\linewidth]{
		\includegraphics[width=0.4\textwidth]{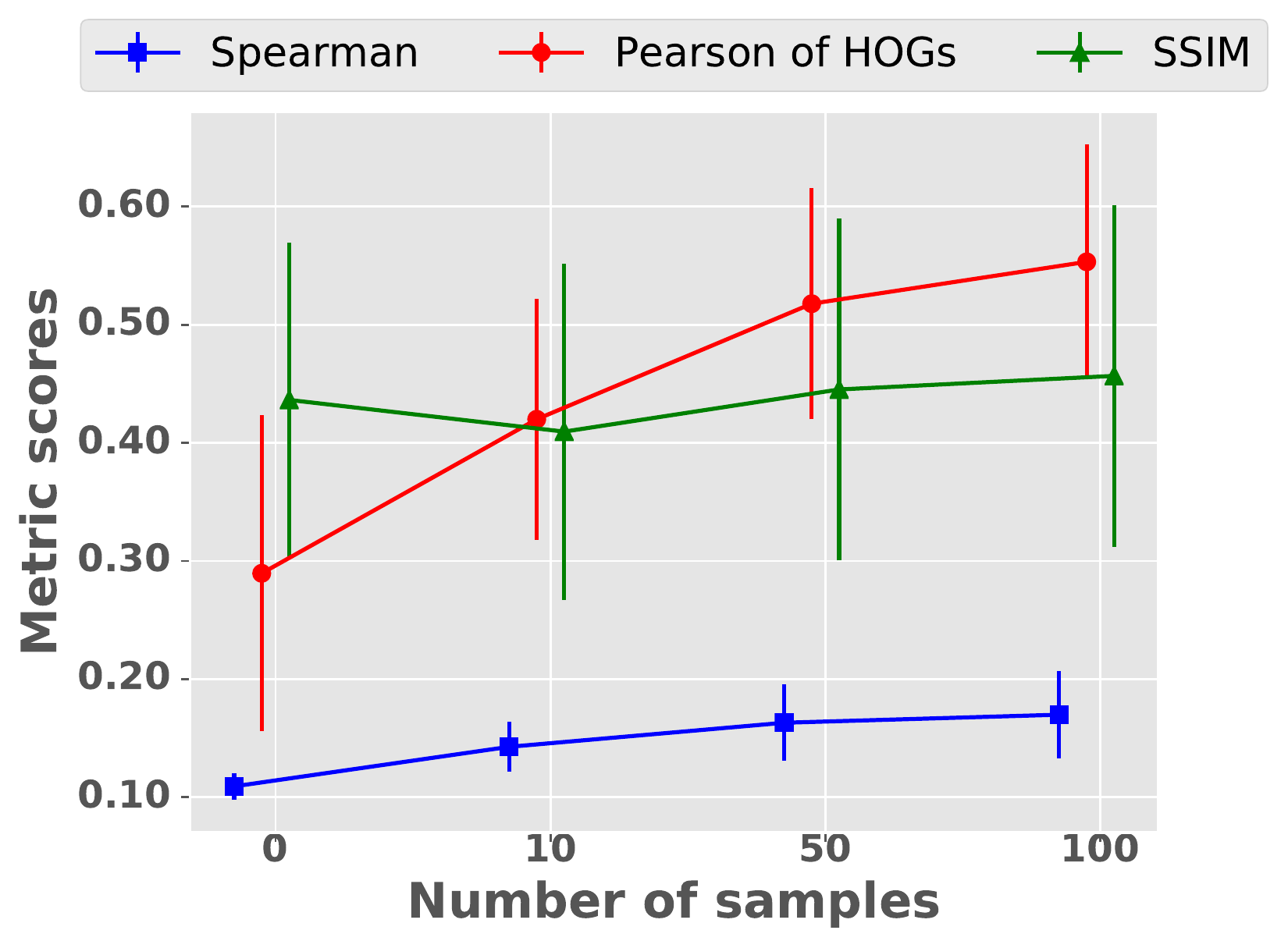}
	}
	\hspace{1.8cm}
	\subcaptionbox{ResNet vs. ResNet-R\label{fig:IG_trend_quan_resNet}}
	[0.40\linewidth]{
		\includegraphics[width=0.4\textwidth]{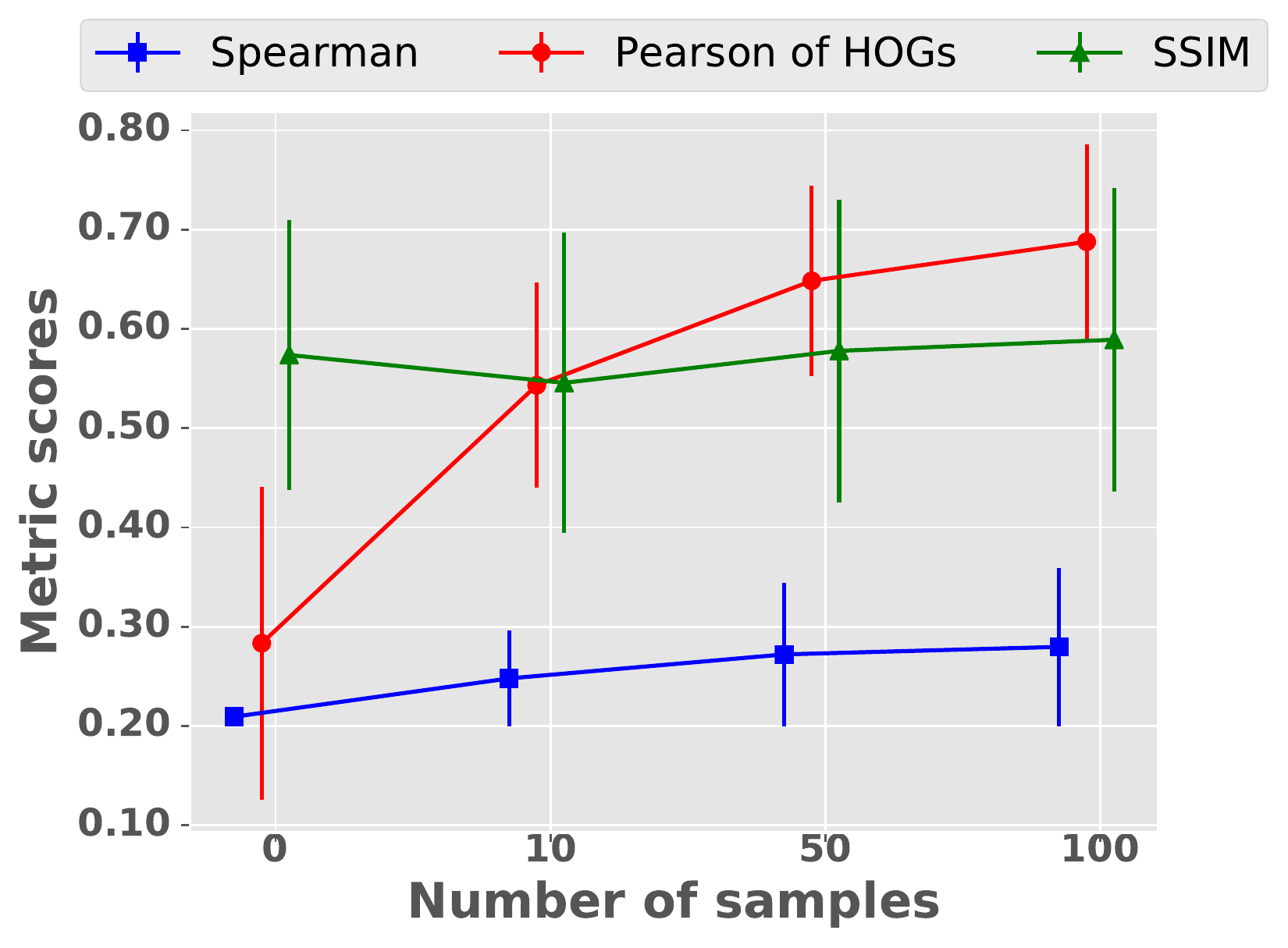}
	}
	\caption{
		Error plots showing the similarity between the Gradient $\odot$ Input \cite{shrikumar2016not} of a robust model (GoogLeNet-R or ResNet-R) and the Integrated Gradient \cite{sundararajan2017axiomatic} of the respective regular model (GoogLeNet or ResNet) across all metrics as we increase \boldsymbol{$N_{T}$} --- a hyperparameter that governs the smoothness of IG explanations.
		Here, $N_{T}=0$ represents the GI of the regular model.
		The scores represent the average similarity scores across $1,735$ images.
%		The SSIM scores, particularly, have high std across all samples.
	}
	\label{fig:IG_trend_quant}
\end{figure}

\subsec{Results} We observed that, on increasing the $N_T$, the IG becomes increasingly similar to the GI of the robust model (Fig.~\ref{fig:IG_trend_qual}). 
The same trend holds for the average similarity scores across the $1735$ images for both GoogLeNet and ResNet (Fig.~\ref{fig:IG_trend_quant}).
Similar to Sec. \ref{sec:question_smoothing}, the observed similarity scores give a false sense of assurance to the end-users about the model robustness.
%These results point out to our observations in Sec. \ref{sec:question_smoothing} that this similarity might give false sense of assurance to the user about the robustness of their models.

%The heatmaps become cleaner as we start increasing $N_{T}$, in IG algorithm (Fig.~\ref{fig:SG_trend_qual}c-e). 
%For a given image, we found that the gradients from IG, using large number of trials, becomes similar to the GI of the robust version of the same model (Fig.~\ref{fig:IG_trend_qual}f). 
%In Fig.~\ref{fig:IG_trend_quant}, we calculate the similarity between attribution maps from IG, where $N_{T}\in\{0, 10, 50, 100\}$, of ResNet and GI of ResNet-R (Fig.~\ref{fig:IG_trend_quan_resNet}). 

%We observe that the similarity increases as we increase $N_{T}$ in IG. 
%Similar trend was observed for GoogLeNet model (Fig.~\ref{fig:IG_trend_quan_googLeNet}).

	\begin{figure}[H]
	\centering
	\vspace{-0.3cm}
	\subcaptionbox{GoogLeNet vs. GoogLeNet-R\label{fig:SG_trend_quan_googLeNet}}
	[0.4\linewidth]{
	\includegraphics[width=0.4\textwidth]{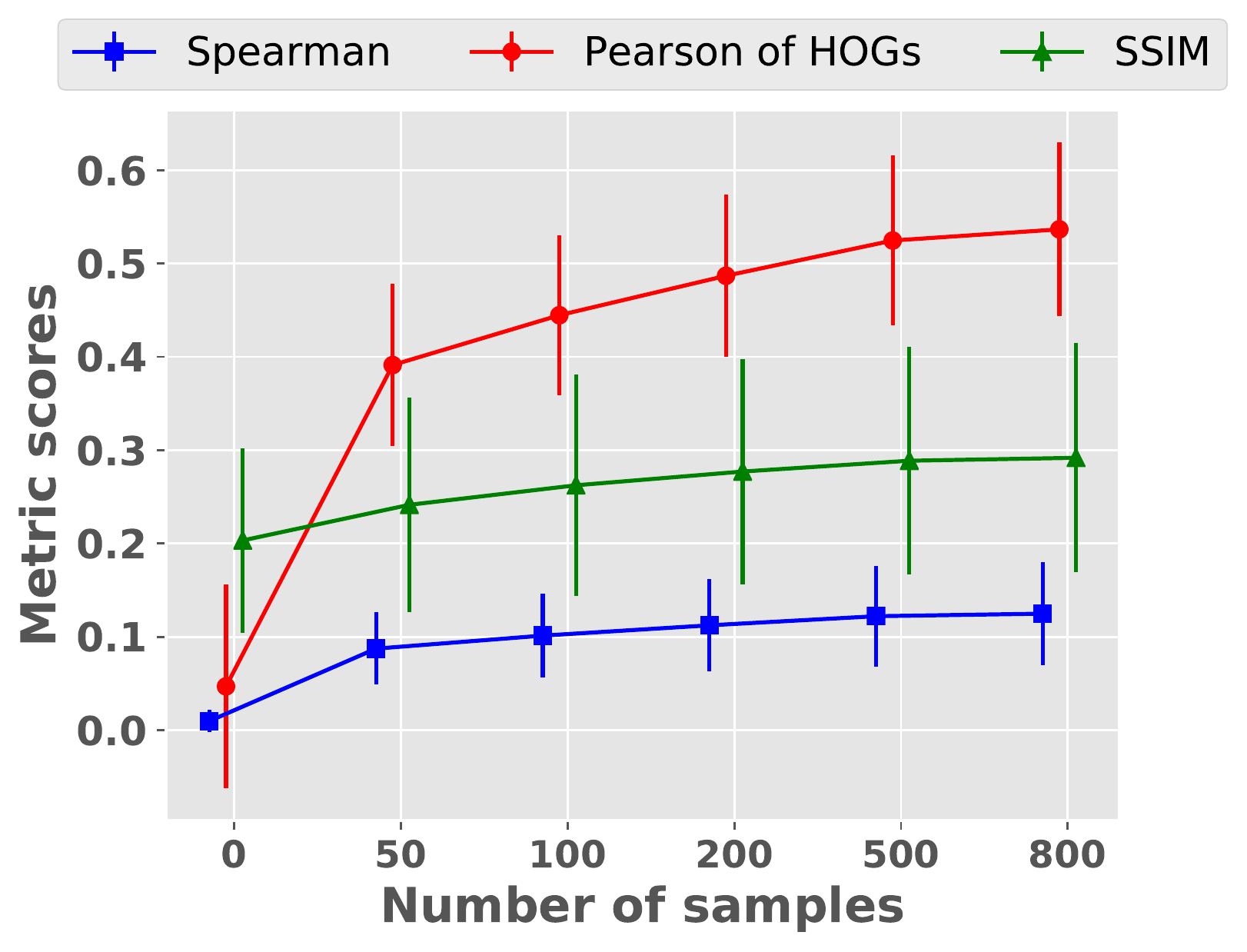}
	}
	\hspace{1.8cm}
	\subcaptionbox{ResNet vs. ResNet-R\label{fig:SG_trend_quan_resNet}}
	[0.4\linewidth]{
	\includegraphics[width=0.41\textwidth]{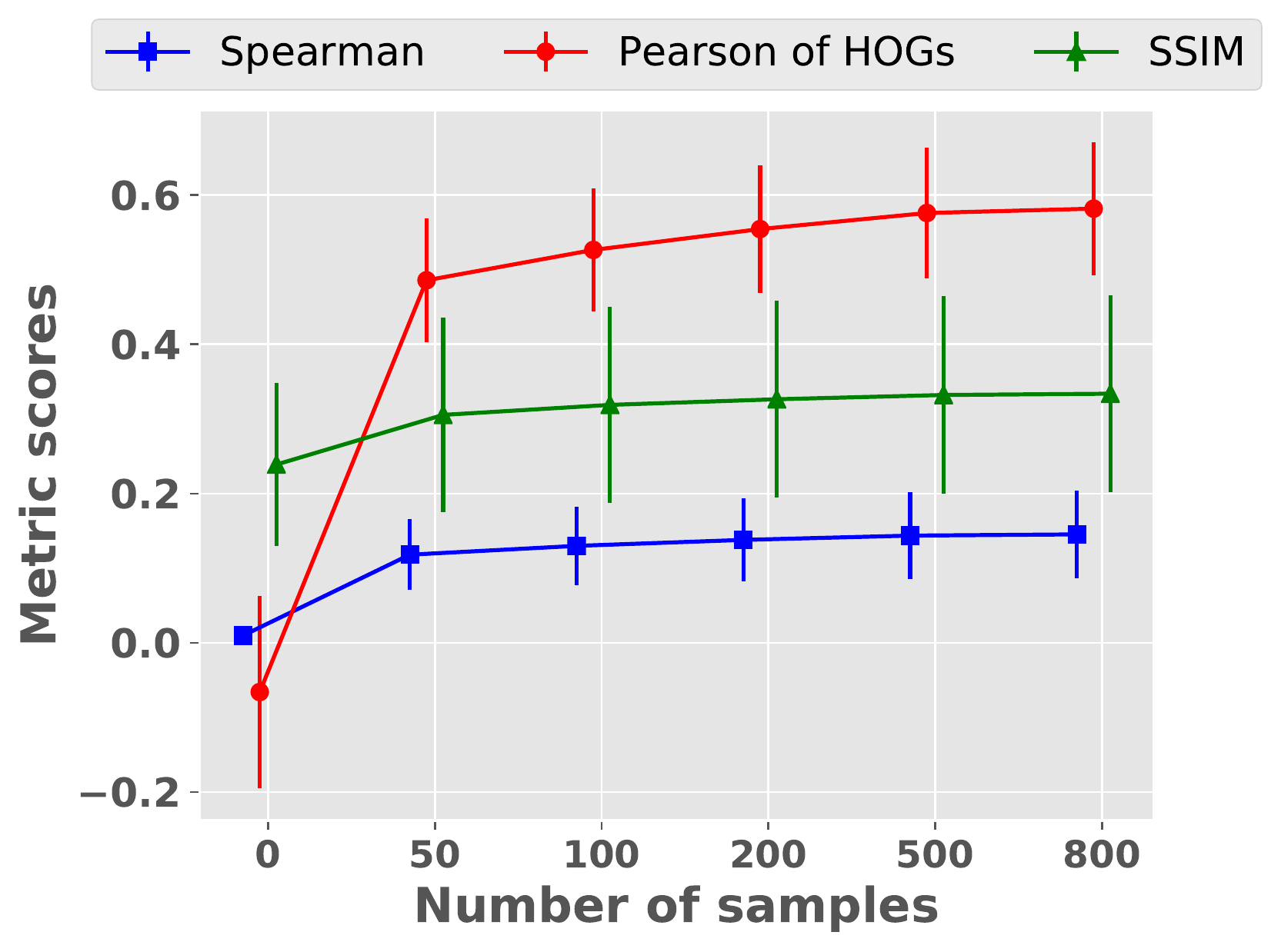}
	}
	\caption{
		Error plots showing the similarity between the gradients of a robust model (GoogLeNet-R or ResNet-R) and the SmoothGrad heatmaps \cite{smilkov2017smoothgrad} of the respective regular model (GoogLeNet or ResNet) across all metrics as we increase \boldsymbol{$N_{SG}$} --- a hyperparameter that governs the smoothness of SG explanations.
		Here, $N_{SG}=0$ represents the gradient of the regular model.
		The scores are the mean similarity scores taken over $1,735$ images.
%		Error plot showing the similarity between the SmoothGrad (SG) attribution maps of regular model and gradient of robust models for different number of SG samples, where (a) GoogLeNet vs. GoogLeNet-R and (b) ResNet vs. ResNet-R.
%		%			This figure represents the quantitative results of Sec.~\ref{sec:question_smoothing}.
%		%			The VG of GoogLeNet \& GoogLeNet-R (a) and ResNet \& ResNet-R (b) are very different (lowest scores across all metrics at point $0$ in the x-axis).
%		We observe an increase in scores across all metrics as we increase the number of samples in SG for both models.
	}
	\label{fig:SG_trend_quant}
\end{figure}

\section{Additional sensitivity experiments}
\label{sec:ext_sens_expt}

\subsection{SmoothGrad sensitivity to the std of Gaussian noise}
\label{sec:appendix_SG}
SmoothGrad (SG) generates the attribution map by averaging the gradients from a number of noisy images. 
The std of Gaussian noise $\sigma$ is a heuristically chosen parameter which, ideally, should not change the resultant attribution map. 
On the contrary, we found that changing $\sigma$ causes a large variation in the SG attribution maps.

\subsec{Experiment} To test the sensitivity to the std of Gaussian noise, we measure the average similarity between a reference heatmap at $\sigma=0.2$ and each of the heatmaps generated by sweeping across $\sigma\in\{0.1, 0.3\}$ on the same input image.
Other than the aforementioned changes, we used all default hyperparameters as in \cite{smilkov2017smoothgrad}.

\begin{figure*}[t]
	\centering
	{	
		\small
		\vspace*{-0.15cm}
		\begin{flushleft}
			\hspace{8.45cm}Input image
			\hspace{0.6cm}$\sigma$=$0.1$
			\hspace{0.8cm}$\sigma$=$0.2$
			\hspace{0.8cm}$\sigma$=$0.3$
			\hspace{-6.7cm}\rotatebox{90}{\hspace{-3.75cm}ResNet-R\hspace{1.1cm}ResNet}			
		\end{flushleft}		
	}
	\vspace{-0.6cm}
	\subcaptionbox{Average robustness across the dataset when changing std of Gaussian noise $\sigma$.\label{fig:SG_noise_quant}}
	[0.3\linewidth]{
		\includegraphics[width=0.3\textwidth]{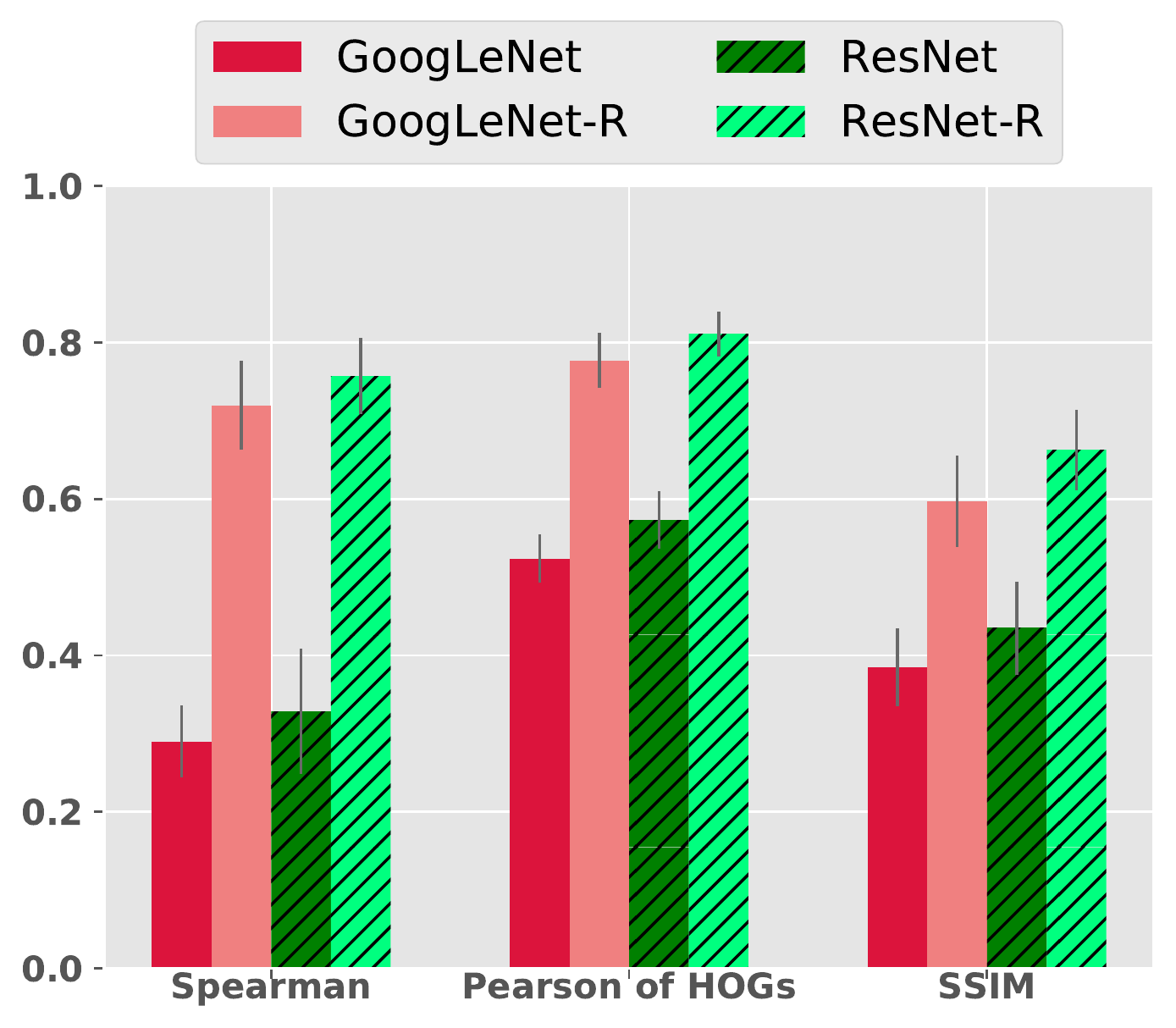}}
	\hspace{1cm}
	\subcaptionbox{SG heatmaps for ResNet-R are more consistent compared to those for ResNet.\label{fig:SG_noise_qual}}
	[0.40\linewidth]{
		\includegraphics[width=0.4\textwidth]{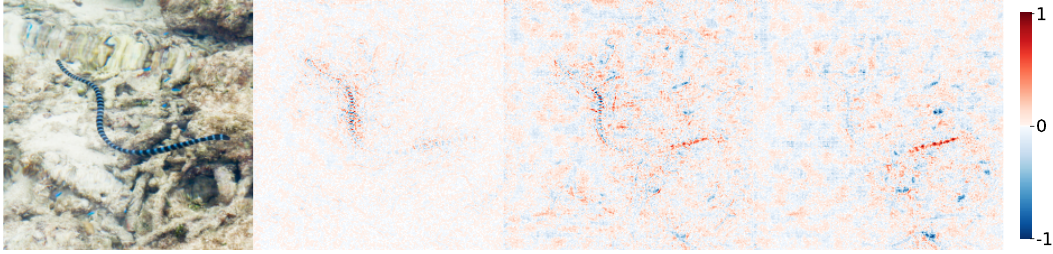}
		\text{SSIM:$0.2307 $}
		\includegraphics[width=0.4\textwidth]{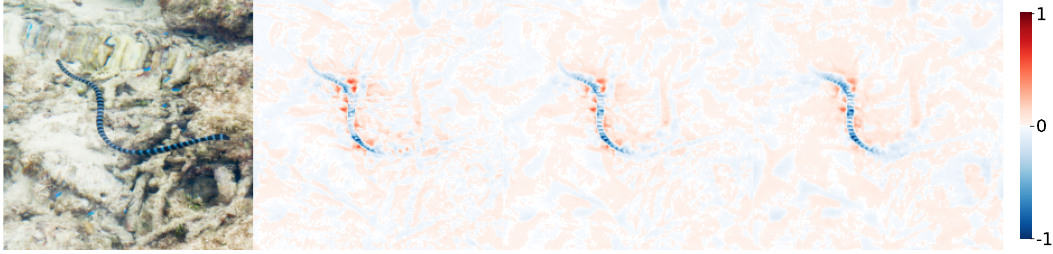}
		\text{SSIM:$0.8509 $}
	}
	\caption{
		Quantitative (a) and qualitative (b) figures showing the sensitivity of SmoothGrad (SG) \cite{smilkov2017smoothgrad} attribution maps when the \textbf{std of the Gaussian noise (\boldsymbol{$\sigma$})} changes. 
		\newline\textbf{Left panel:} 
%		SG maps are sensitive to $\sigma$ of Gaussian noise across all models. 
		Compared to regular models (GoogLeNet and ResNet), heatmaps generated for robust models (GoogLeNet-R and ResNet-R) are substantially more consistent to when the Gaussian std hyperparameter changes (a).
%		Lower sensitivity of robust models across all three metrics as compared to regular models (a).
%		is because they were adversarially trained using noisy inputs.
		\newline\textbf{Right panel:} Across the dataset, the reference image caused the largest difference between the SSIM scores of ResNet heatmaps vs. ResNet-R heatmaps (b).
		As $\sigma$ increases, the attribution maps of ResNet become noisier while ResNet-R heatmaps become smoother (b; row 1).
%		A higher SSIM (0.8509) is observed between the ResNet-R attribution maps (row 2) for different $\sigma 's$ as compared to the SSIM (0.2307) in ResNet (row 1) case.
		%		SG Gaussian noise \{0.1, 0.2, 0.3\} sensitivity  | SSIM: ResNet:0.6901  ResNet-R: 0.8938. 50 samples.
		%		\todo{Find the bottom rows showing minimum variations}
		%		\todo{Choose matchstick}
	}
	\label{fig:SG_noise_sensitivity}
\end{figure*}

\subsec{Results} We found that the SG attribution maps of regular models are more sensitive as compared to that of robust models (Fig.~\ref{fig:SG_noise_qual}).
Quantitatively, high sensitivity was observed in the average similarity scores across the dataset (Fig.~\ref{fig:SG_noise_quant}).
Notably, the average Spearman correlation score, across the dataset, for GoogLeNet-R is 2.5$\times$ than that of GoogLeNet.
% ($0.2897$). 
%Similar increases were also observed for other Similarity metrics (Fig.~\ref{fig:SG_noise_quant}).

\subsection{LIME sensitivity to changes in the random seed and number of perturbed samples}
\label{sec:appendix_LIME}
%LIME generates an attribution map by linearly fitting randomly perturbed images. 
The most common hyperparameter setting for LIME is the random seed for sampling different superpixel combinations.
We quantify the sensitivity of LIME across different random seeds as one can expect a minimum change in the output attribution map on changing the algorithm seed.

\subsec{Experiment}
To test the sensitivity to random seed, we measure the average similarity between a reference heatmap at $seed=0$ and each of the heatmaps generated by sweeping across $seed\in\{1, 2, 3, 4\}$ on the same input image.
%We choose five random seeds, $\{0, 1, 2, 3, 4\}$, with seed$=0$ as our reference and calculate the average similarity scores using all four pairs.
Notably, the number of intermediate samples for the linear regression fitting in LIME is an important factor for the resultant heatmap. 
Hence, we also quantify the sensitivity of LIME across the number of perturbed images, \ie $N_{LIME} \in \{500, 1000\}$, to generate two heatmaps and calculate the average similarity metric scores between them. 

\begin{figure*}[h]
	\centering
	{	
		\small
%		\vspace*{-0.15cm}
		\begin{flushleft}
			\hspace{7.1cm}Input image
			\hspace{0.35cm}seed=$0$
			\hspace{0.55cm}seed=$1$
			\hspace{0.55cm}seed=$2$
			\hspace{0.55cm}seed=$3$
			\hspace{0.55cm}seed=$4$
			\hspace{-9.3cm}\rotatebox{90}{\hspace{-3.5cm}ResNet-R\hspace{1.0cm}ResNet}
		\end{flushleft}		
	}
	\vspace{-0.9cm}
	\subcaptionbox{Average similarity scores across the dataset when changing random seed.\label{fig:LIME_seed_quant}}
	[0.3\linewidth]{
		\includegraphics[width=0.3\textwidth]{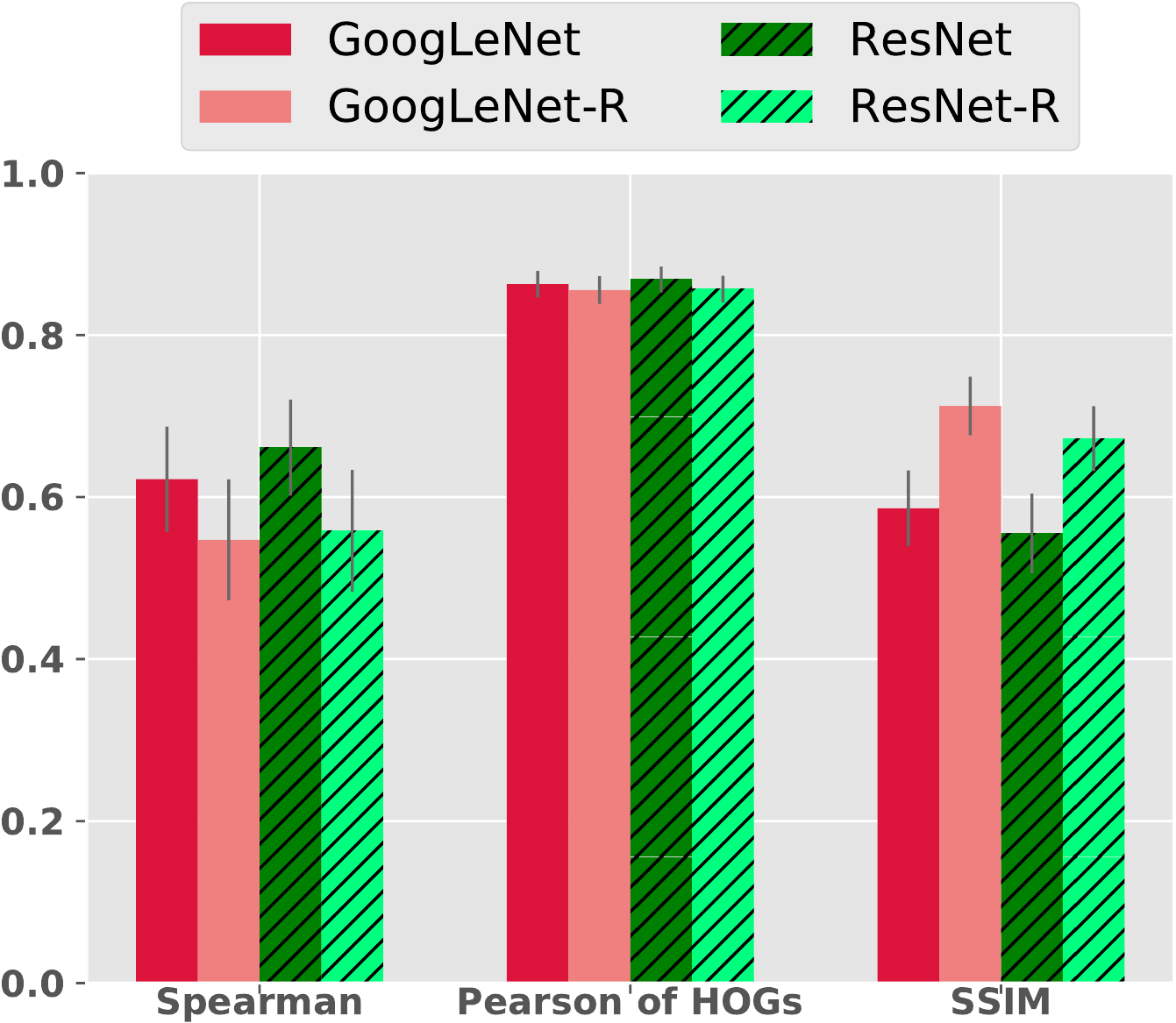}}
	\hspace{1.0cm}	
	\subcaptionbox{LIME heatmaps for ResNet-R are more consistent (under SSIM similarity score) compared to those for ResNet.\label{fig:LIME_seed_qual}}
	[0.55\linewidth]{
		\includegraphics[width=0.55\textwidth]{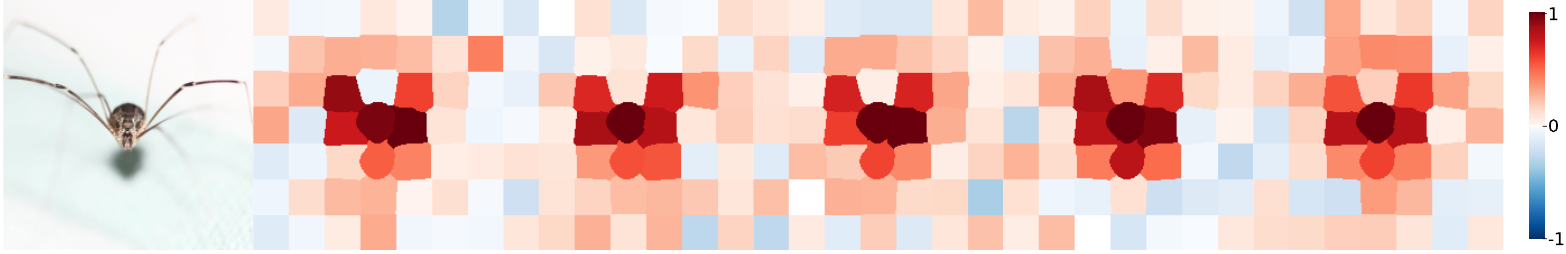}
		\text{SSIM: $0.1822$}
		\includegraphics[width=0.55\textwidth]{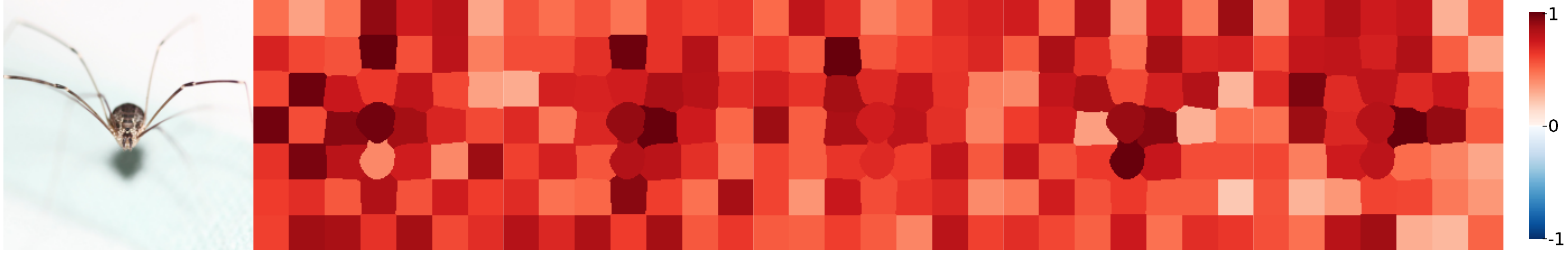}
		\text{SSIM: $0.7793$}
	}
	
	\caption{
		Quantitative (a) and qualitative (b) figures showing the sensitivity of LIME attribution maps when \textbf{the random seed} of LIME (which governs the random selection of LIME superpixel masks) changes. 
		%		The number of superpixels and the total number of intermediate samples were fixed to $50$ and $1000$ for all seeds.
		\newline\textbf{Left panel:} For both regular and robust models, LIME attribution maps are similarly sensitive to the random seed (similarity scores well below 1.0).
		The high Pearson of HOGs scores are hypothesized to be because the SLIC superpixel segmentation imposes a consistent visual structure bias across LIME attribution maps (before and after the random seed changes).
		Under SSIM, LIME heatmaps of robust models are more consistent than those of regular models.
		\newline\textbf{Right panel:} Across the dataset, the reference image causes the largest difference between the SSIM scores of ResNet heatmaps and those of ResNet-R heatmaps (b; top row vs. bottom row).
%		LIME attribution maps of ResNet (b; top row) gives different weight to most superpixels across the five random seeds.
	}
	%		\todo{choose the lowest SSIM score image} \todo{remove bottom row}}
	\label{fig:LIME_seed_sensitivity}
\end{figure*}

\begin{figure*}[]
	\centering
	{	
		\small
		\vspace*{-0.15cm}
		\begin{flushleft}
			\hspace{9.35cm}Input image
			\hspace{0.01cm}$N_{LIME}$=$500$
			\hspace{0.01cm}$N_{LIME}$=$1000$
			\hspace{-5.7cm}\rotatebox{90}{\hspace{-3.75cm}ResNet-R\hspace{1.1cm}ResNet}
		\end{flushleft}		
	}
	\vspace{-0.7cm}
	\subcaptionbox{Average robustness across the dataset when changing $N_{LIME}$.\label{fig:LIME_sample_quant}}
	[0.3\linewidth]{
		\includegraphics[width=0.3\textwidth]{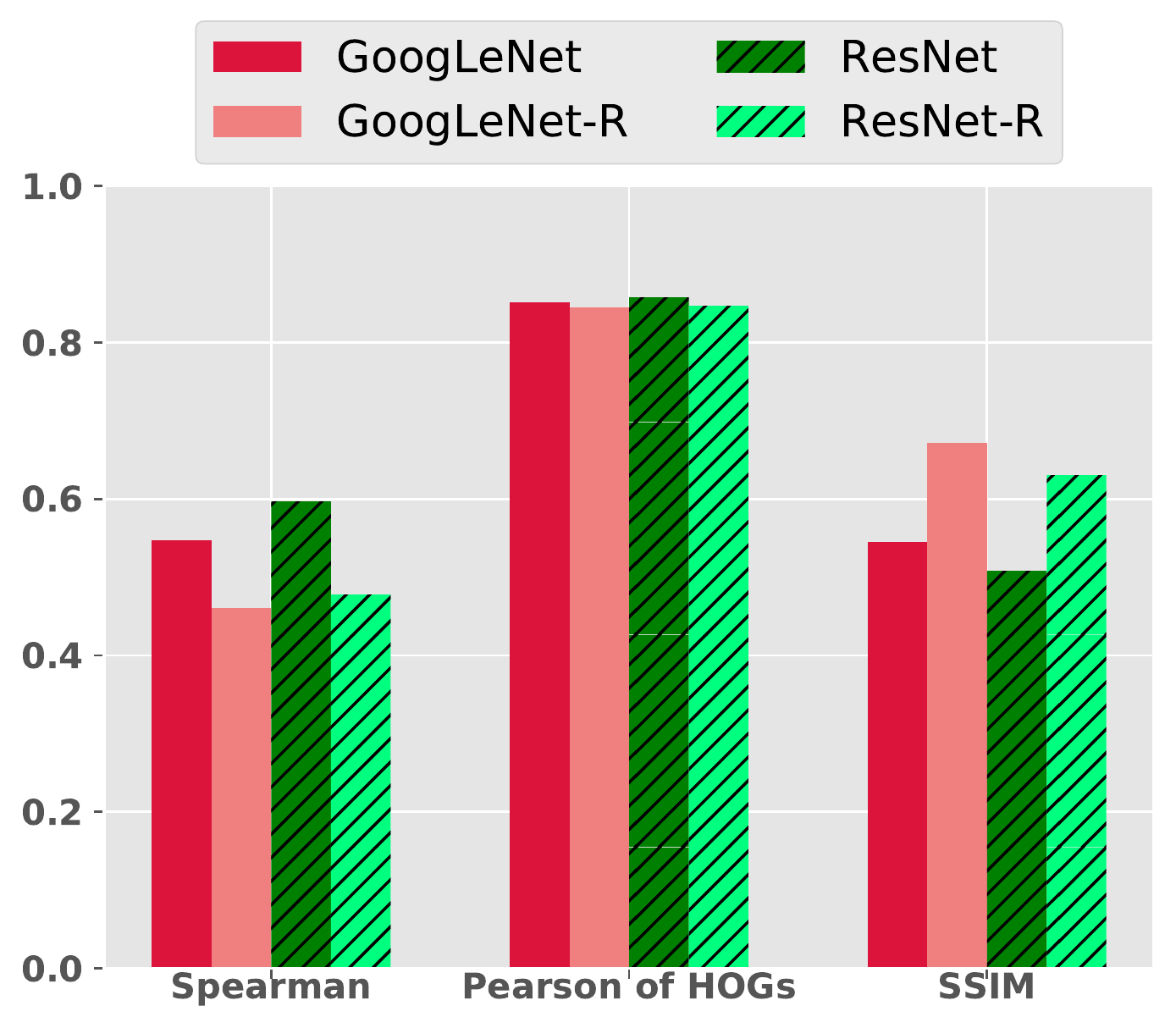}}
	\hspace{1cm}
	\subcaptionbox{LIME heatmaps for ResNet-R are more consistent compared to those for ResNet.\label{fig:LIME_sample_qual}}
	[0.3\linewidth]{
		\includegraphics[width=0.3\textwidth]{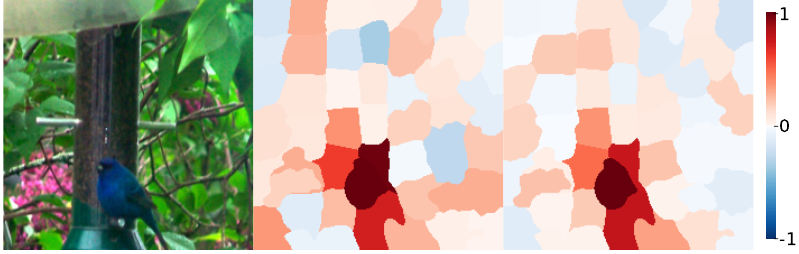}
		\text{SSIM: $0.0918$}
		\includegraphics[width=0.3\textwidth]{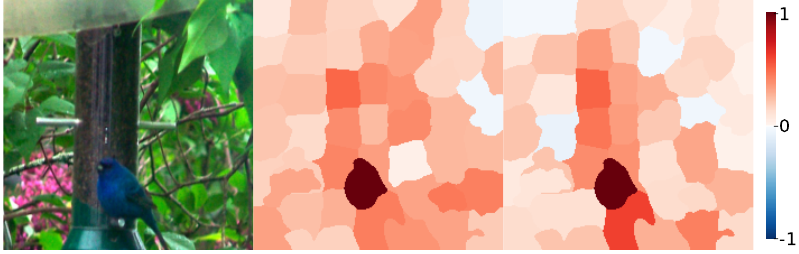}
		\text{SSIM: $0.5655$}
	}
	\caption{Quantitative (a) and qualitative (b) figures showing the sensitivity of LIME attribution maps when \textbf{the number of perturbed samples \boldsymbol{$N_{LIME}$}} changes.
		\newline\textbf{Left panel}: Both robust and regular models are similarly sensitive to the $N_{LIME}$ under Pearson correlation of HOGs while the heatmaps for robust models are more consistent under SSIM (a).
		\newline\textbf{Right panel}: Across the dataset, the reference input image causes the largest difference between the SSIM scores of ResNet heatmaps vs. the SIM scores of ResNet-R heatmaps (b; top row vs. bottom row).}
	\label{fig:LIME_sample_sensitivity}
\end{figure*}

\subsec{Results} We did not observe any significant difference between similarity scores of robust and regular models across both experiments (Fig.~\ref{fig:LIME_seed_quant}, \ref{fig:LIME_sample_quant}). 
%All models were found to be sensitive. 
Note that the robust models were adversarially-trained on pixel-wise noise whereas, LIME operates at the superpixel level. 
We hypothesize this to be a reason for insignificant differences found between robust vs. regular models when changing the random seed. 
The previous experiments were performed at the number of superpixels $S = 50$.
Additionally, we also repeated the same experiments at $150$ superpixels but observed no significant improvement in the robustness of robust models (data not shown).

Strikingly, the Pearson correlation value for HOG features are high in both experiments (Fig.~\ref{fig:LIME_seed_quant} \& Fig.~\ref{fig:LIME_sample_quant}). 
An explanation for that is because the SLIC superpixel segmentation step of LIME imposes a strong structural bias in LIME attribution maps.

\newpage
\subsection{Meaningful Perturbation sensitivity to changes in the random seed}
\label{sec:appendix_MP}
For MP mask optimization, Fong et al. \cite{fong2017interpretable} used a circular mask initialization that suppresses the score of the target class by $99\%$ when compared to that of using a completely blurred image. We argue that this circular mask acts as a strong bias towards ImageNet images (\ie they may not work for other datasets) since ImageNet mostly contains object-centric images. 
Hence, we evaluate the sensitivity of MP attribution maps by initializing masks with different random seeds (corresponding to different mask initializations).

\subsec{Experiment} 
Similar to Sec.~\ref{sec:appendix_LIME}, we calculate the average pairwise similarity between a reference heatmap using $seed=0$ and each of the heatmaps generated by sweeping across $seed\in\{1, 2, 3, 4\}$ on the same input image.
All the other hyperparameters are the same as in \cite{fong2017interpretable}.

\begin{figure*}[h]
	\centering
	{	
		\small
%		\vspace*{-0.15cm}
		\begin{flushleft}
			\hspace{7.1cm}Input image
			\hspace{0.2cm}seed~=~0
			\hspace{0.4cm}seed~=~1
			\hspace{0.4cm}seed~=~2
			\hspace{0.4cm}seed~=~3
			\hspace{0.4cm}seed~=~4
			\hspace{-9.3cm}\rotatebox{90}{\hspace{-3.5cm}ResNet-R\hspace{0.9cm}ResNet}	
		\end{flushleft}		
	}
	\vspace{-0.9cm}
	\subcaptionbox{Average robustness across the dataset when changing random seed.\label{fig:MP_seed_quant}}
	[0.3\linewidth]{
		\includegraphics[width=0.3\textwidth]{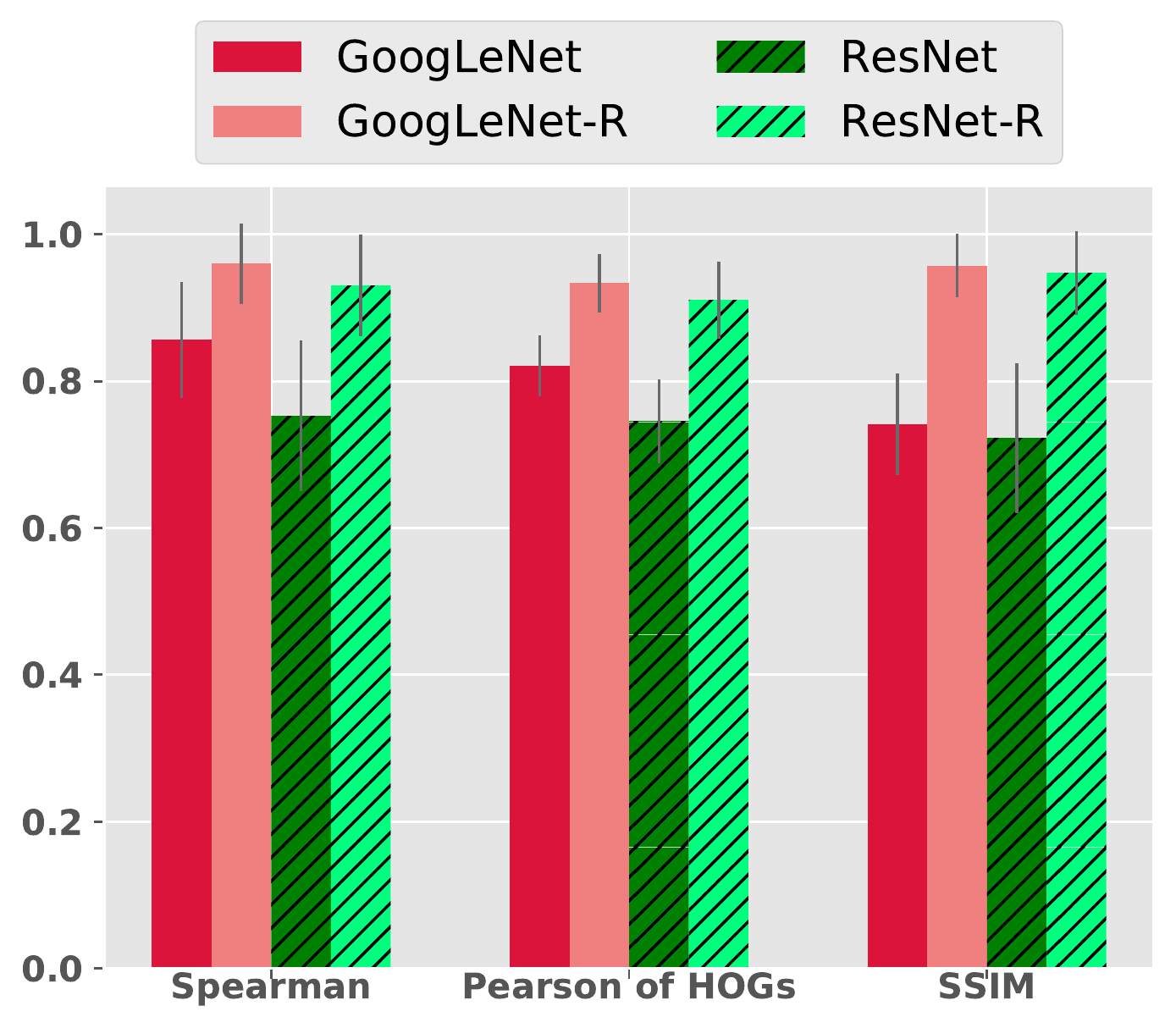}}
	\hspace{1cm}	
	\subcaptionbox{MP heatmaps for ResNet-R are more consistent compared to those for ResNet.\label{fig:MP_seed_qual}}
	[0.55\linewidth]{
		\includegraphics[width=0.55\textwidth]{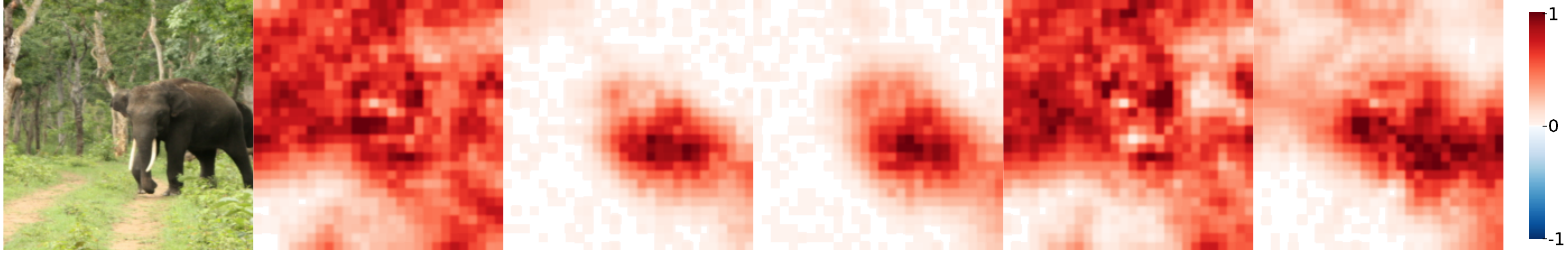}
		\text{SSIM: $0.3305$}
		\includegraphics[width=0.55\textwidth]{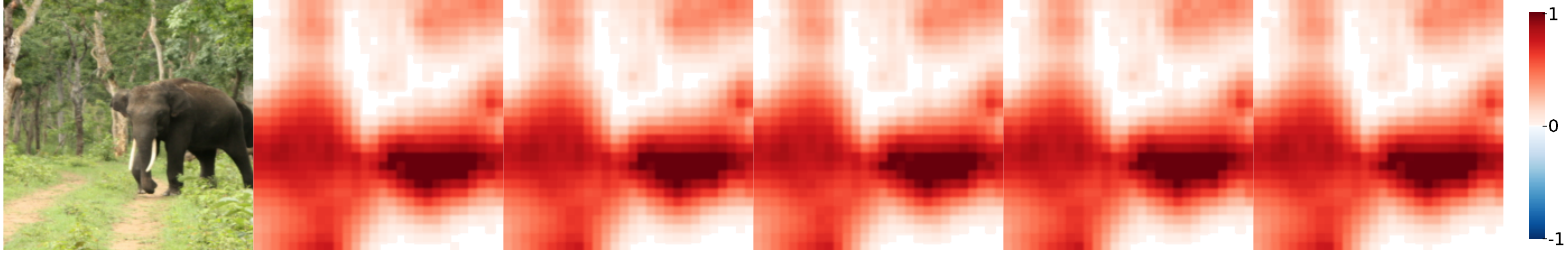}
		\text{SSIM: $0.9871$}
	}
	\caption{
		%		Across the dataset, the average similarity scores between the MP attribution maps for regular models is lower than robust models.
		Robust classifiers cause heatmaps to be more consistent (\ie higher SSIM scores) when \textbf{the random seed} changes, both quantitatively (a) and qualitatively (b).
		\newline\textbf{Right panel:} Across the dataset, the reference image causes the largest difference between the SSIM scores of ResNet heatmaps vs. ResNet-R heatmaps (b; top row vs. bottom row).
		%		ResNet (row 1) gives different attribution maps for differet random seeds.
		%		\todo{maximim difference criteria}
		%		Quantitative (a) and Qualitative (b) figures show the sensitivity of MP attribution maps on changing the mask initialization seed. 
		%		All the other MP hyperparameters were fixed to their default values for all seeds.
		%		\textbf{(a)}: High sensitivity (low metric scores) for the MP attribution maps of regular models. 
		%		The adversarial trained robust models show less sensitivity (high metric scores) both quantitatively and qualitatively.
		%		\textbf{(b)}: Across the dataset, the reference image caused the largest difference between the SSIM scores of ResNet heatmaps vs. ResNet-R heatmaps.
		%		ResNet (row 1) gives different attribution maps for differet random seeds. 
		%		It is interesting to see how for some seeds MP gives completely noisy heatmaps (lower SSIM). 
		%		However, for ResNet-R we get consistent heatmaps (higher SSIM) across all five random seeds.
	}
	\label{fig:MP_seed_sensitivity}
\end{figure*}

\subsec{Results} We found that robust models are less sensitive to random initialization of masks (Fig.~\ref{fig:MP_seed_qual}).
The average similarity scores for robust models are consistently higher than their regular counterparts (Fig.~\ref{fig:MP_seed_quant}). 
%This can be observed qualitatively in Fig.~\ref{fig:MP_seed_qual}.
%Interestingly, MP completely fails for some seeds (seed=0 and seed=3 of ResNet in Fig.~\ref{fig:MP_seed_qual}).

\begin{figure*}[h]
	\centering
	\subcaptionbox{Average similarity of heatmaps across the dataset under three metrics when the blur radius $b_{R}$ changes.\label{fig:MP_blur_quant}}
	[0.3\linewidth]{
		\includegraphics[width=0.3\textwidth]{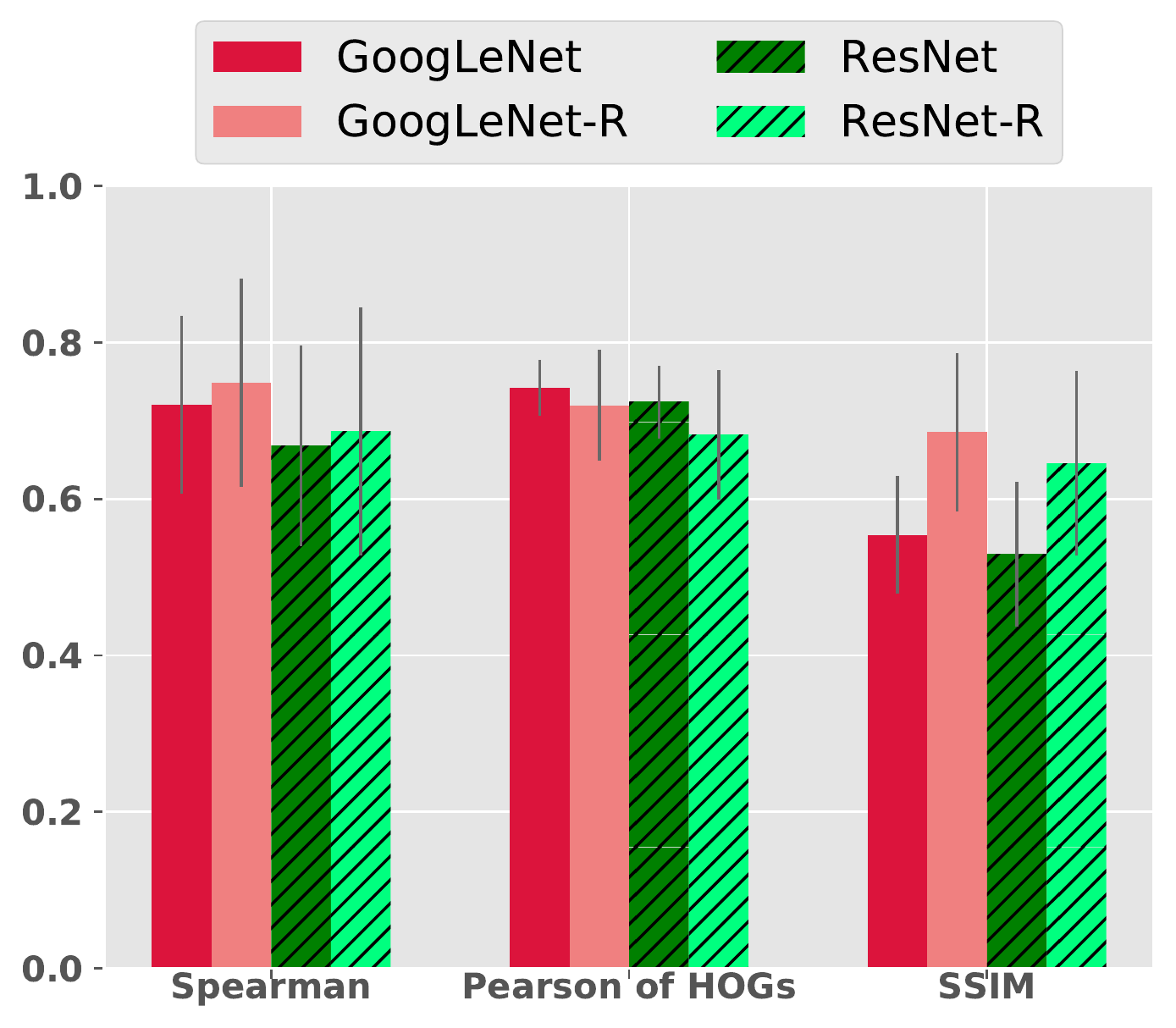}}
	\hspace{1.0cm}
	\subcaptionbox{Average similarity of heatmaps across the dataset under three metrics when the number of iterations $N_{iter}$ changes.\label{fig:MP_iter_quant}}
	[0.5\linewidth]{
		\includegraphics[width=0.3\textwidth]{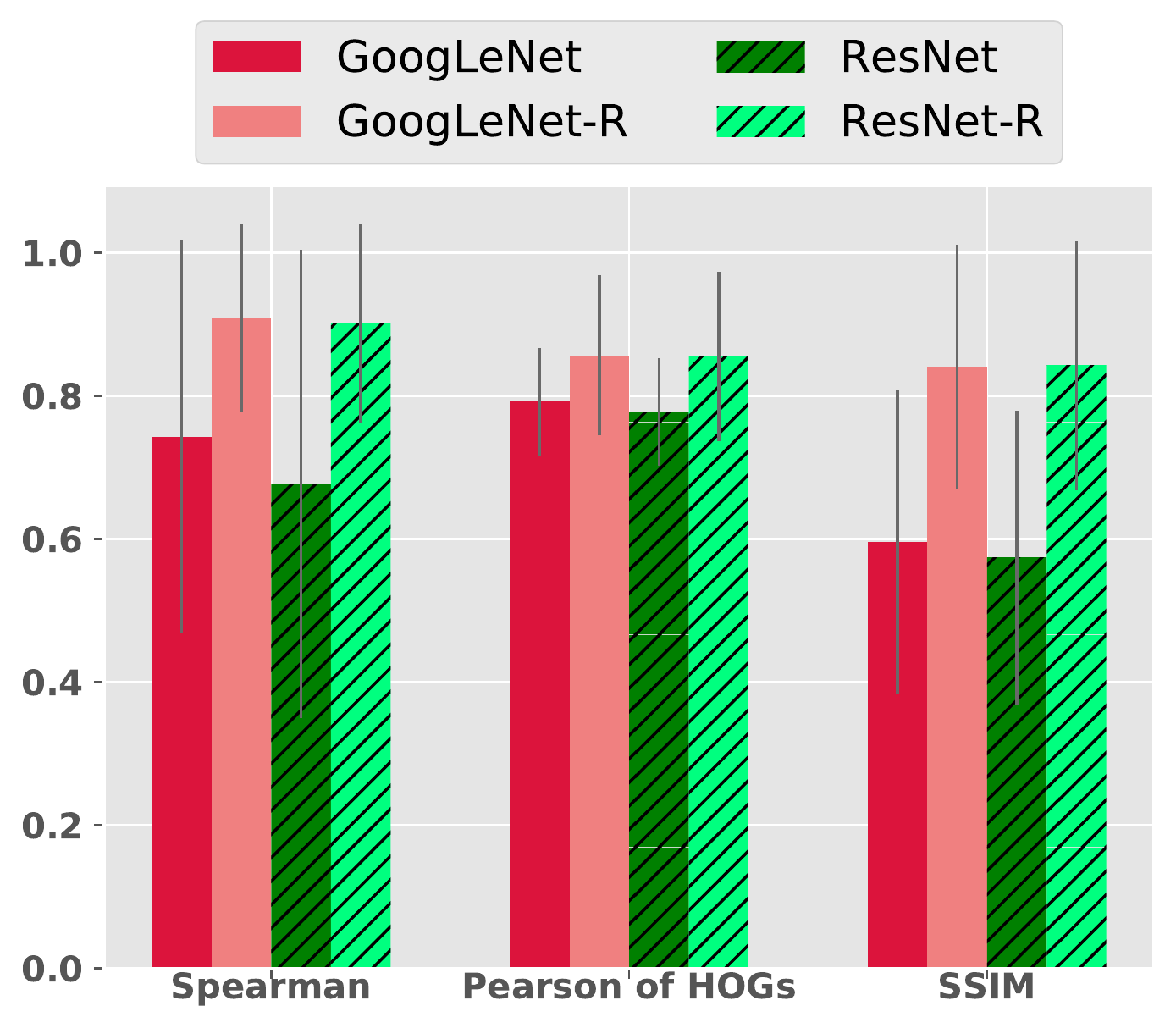}}
\caption{
	Error bar plots showing the similarity of \textbf{Meaningful-Perturbation (MP)} attribution maps when a hyperparameter---here \textbf{Gaussian blur radius \boldsymbol{$b_R$}} (a), and \textbf{the number of iterations \boldsymbol{$N_{iter}$}} (b)---changes.
	These figures represent the quantitative results for the experiments in Sec.~\ref{sec:MP_exp}.
	\newline\textbf{Left panel}: Changing $b_R$ caused the heatmaps for regular classifiers (GoogLeNet and ResNet) to vary more, under Spearman rank correlation and SSIM, than those for robust classifiers (see Figs.~\ref{fig:MP_blur_qual} \&~\ref{fig:appendix_MP_blur} for qualitative results).
	\newline\textbf{Right panel}: Heatmaps generated for regular models (dark red \& dark green) are consistently more variable than those generated for robust models (light red \& light green) across all metrics (b).
}
\label{fig:MP_blur_iter_sensitivity}
\end{figure*}

\setlength{\tabcolsep}{1.5pt}
\begin{table*}[]
	\centering
	\def\arraystretch{1.3}%
	\begin{tabular}{|c|c|c|c|c|c|c|}
		\hline
		\textbf{Algorithm}             & \textbf{Models}      & \textbf{SSIM}  & \textbf{\begin{tabular}[c]{@{}c@{}}Localization\\ Error\end{tabular}} & \textbf{Insertion} & \textbf{Deletion} \\ \hline
		\multirow{4}{*}{\textbf{SG}}   & \textbf{GoogLeNet}   & 0.6422$\pm$0.3197 & 0.2744$\pm$0.1382                                                                      & 0.1627$\pm$0.0386     & 0.2091$\pm$0.0453    \\ \cline{2-6} 
		& \textbf{GoogLeNet-R} & 0.9648$\pm$0.0051 & 0.2798$\pm$0.0539                                                                      & 0.2146$\pm$0.0085     & 0.2433$\pm$0.0090    \\ \cline{2-6} 
		& \textbf{ResNet}   & 0.7854$\pm$0.0238 & 0.2632$\pm$0.1140                                                                     & 0.2012$\pm$0.0388     & 0.2342$\pm$0.0447    \\ \cline{2-6} 
		& \textbf{ResNet-R} & 0.9780$\pm$0.0034 & 0.2566$\pm$0.0611                                                                    & 0.2745$\pm$0.0089     & 0.3054$\pm$0.0095    \\ \hline
		\multirow{4}{*}{\textbf{SP-S}} & \textbf{GoogLeNet}   & 0.9221$\pm$0.0321 & 0.3524$\pm$0.0926                                                                      & 0.5056$\pm$0.0208     & 0.1616$\pm$0.0132    \\ \cline{2-6} 
		& \textbf{GoogLeNet-R} & 0.9894$\pm$0.0069 & 0.3468$\pm$0.0424                                                                      & 0.4281$\pm$0.0082     & 0.1260$\pm$0.0039    \\ \cline{2-6} 
		& \textbf{ResNet}   & 0.9633$\pm$0.0188 & 0.4649$\pm$0.1182                                                                       & 0.5959$\pm$0.0226     & 0.2581$\pm$0.0173    \\ \cline{2-6} 
		& \textbf{ResNet-R} & 0.9891$\pm$0.0073 & 0.3666$\pm$0.0660                                                                      & 0.4699$\pm$0.0075     & 0.1459$\pm$0.0041    \\ \hline
		\multirow{4}{*}{\textbf{SP-L}} & \textbf{GoogLeNet}   & 0.6210$\pm$0.1021 & 0.3390$\pm$0.2194                                                                      & 0.4078$\pm$0.1354     & 0.1456$\pm$0.0595    \\ \cline{2-6} 
		& \textbf{GoogLeNet-R} & 0.6540$\pm$0.1361 & 0.3344$\pm$0.1729                                                                      & 0.4130$\pm$0.0817     & 0.1265$\pm$0.0434    \\ \cline{2-6} 
		& \textbf{ResNet}   & 0.8239$\pm$0.0718 & 0.4158$\pm$0.2827                                                                     & 0.4846$\pm$0.1493     & 0.2344$\pm$0.0885    \\ \cline{2-6} 
		& \textbf{ResNet-R} & 0.6867$\pm$0.1276 & 0.3493$\pm$0.2066                                                                      & 0.4485$\pm$0.0861     & 0.1481$\pm$0.0528    \\ \hline
		\multirow{4}{*}{\textbf{LIME}} & \textbf{GoogLeNet}   & 0.5862$\pm$0.0467 & 0.3260$\pm$0.1458                                                                      & 0.5844$\pm$0.0458     & 0.1352$\pm$0.0227    \\ \cline{2-6} 
		& \textbf{GoogLeNet-R} & 0.7125$\pm$0.0363 & 0.3331$\pm$0.1030                                                                      & 0.3832$\pm$0.0432     & 0.1340$\pm$0.0220    \\ \cline{2-6} 
		& \textbf{ResNet}   & 0.5552$\pm$0.0491 & 0.2951$\pm$0.1565                                                                      & 0.7224$\pm$0.0421     & 0.1800$\pm$0.0281    \\ \cline{2-6} 
		& \textbf{ResNet-R} & 0.6722$\pm$0.0401 & 0.3301$\pm$0.1361                                                                      & 0.4549$\pm$0.0424     & 0.1437$\pm$0.0223    \\ \hline
		\multirow{4}{*}{\textbf{MP}}   & \textbf{GoogLeNet}   & 0.7412$\pm$0.0697 & 0.2386$\pm$0.1458                                                                      & 0.5345$\pm$0.0402     & 0.1275$\pm$0.0278    \\ \cline{2-6} 
		& \textbf{GoogLeNet-R} & 0.9572$\pm$0.0432 & 0.2875$\pm$0.0725                                                                      & 0.4001$\pm$0.0176     & 0.1222$\pm$0.0086    \\ \cline{2-6} 
		& \textbf{ResNet}   & 0.7221$\pm$0.1019 & 0.2651$\pm$0.1892                                                                      & 0.6184$\pm$0.0556     & 0.2064$\pm$0.0524    \\ \cline{2-6} 
		& \textbf{ResNet-R} & 0.9476$\pm$0.0572 & 0.2928$\pm$0.0941                                                                      & 0.4328$\pm$0.0226     & 0.1407$\pm$0.0121    \\ \hline
	\end{tabular}
	\caption{
		The results in this table are the numeric format of Fig.~\ref{fig:eval_sensitivity}. Compared to regular models, robust classifiers (GoogLeNet-R and ResNet-R) are more robust in the attribution space (\ie higher SSIM scores) and also more robust in the downstream accuracy space (\ie smaller stds across three different accuracy metrics: Localization error, Deletion, and Insertion).
	}
	\label{tab:eval_sensitivity}
\end{table*}

\begin{figure}[h]
	\centering
	\includegraphics[width=0.5\linewidth]{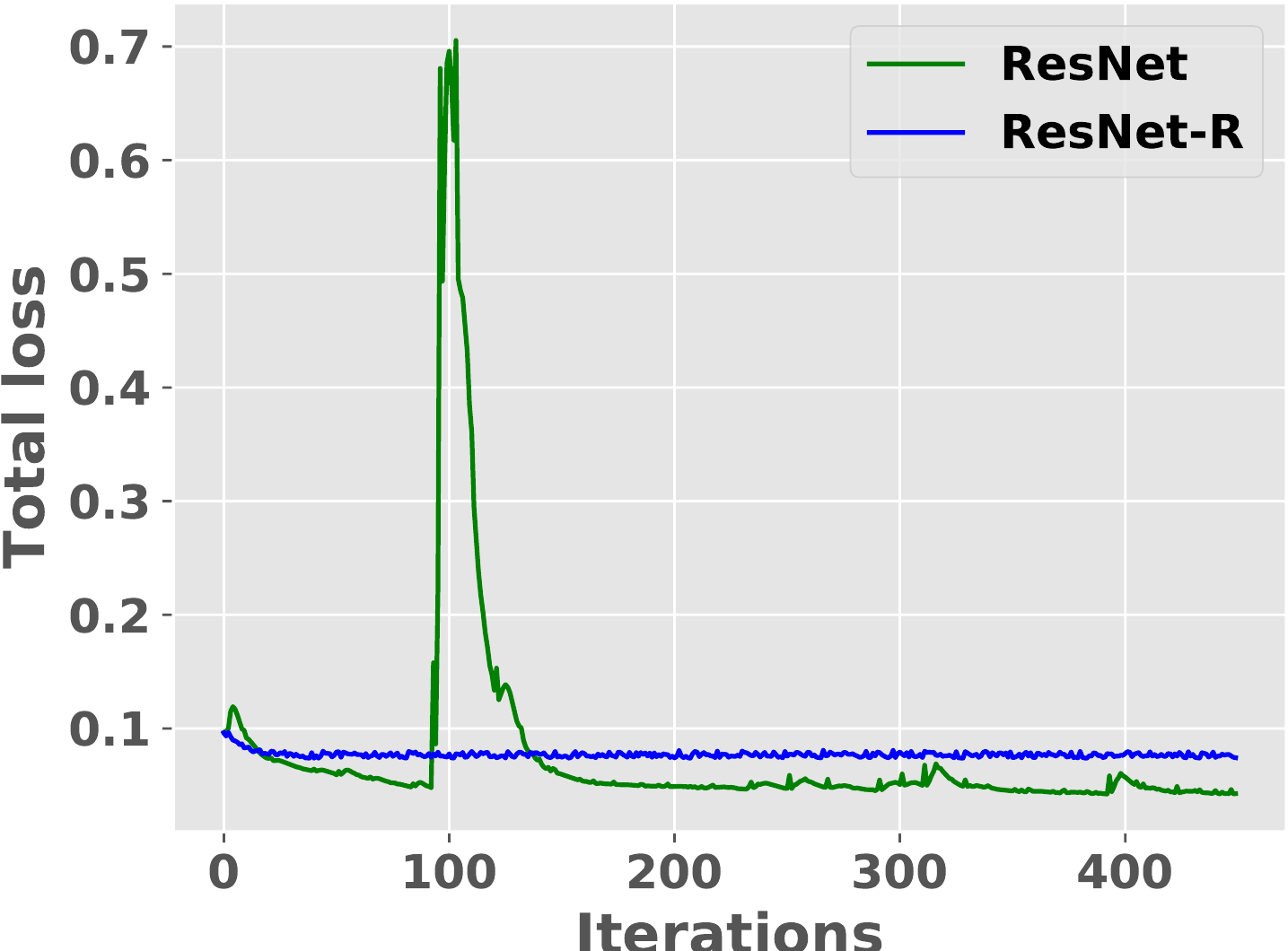}\\
	\caption{
		The total-loss plots ($L_{1}$ + $TV$ + $softmax$) when running MP optimization algorithm (using a ResNet and ResNet-R classifier) on the reference \class{studio~couch} image in Fig.~\ref{fig:MP_iter_qual}.
		The loss curve for ResNet-R converges quickly after 10 steps while MP loss curve often fluctuates (here, peaked at around step 100).
%		Total loss plot ($L_{1}$ + $TV$ + $softmax$), of the reference \class{studio~couch} image from Fig.~\ref{fig:MP_iter_qual}, in MP optimization for ResNet and ResNet-R model.
%		The volatile optimization of ResNet is oberved in the figure above whereas the optimization for ResNet-R is smoother.
	}
	\label{fig:MP_loss}
\end{figure}

\begin{figure*}
	\begin{subfigure}{1.0\linewidth}
		\centering
		\includegraphics[width=0.8\linewidth]{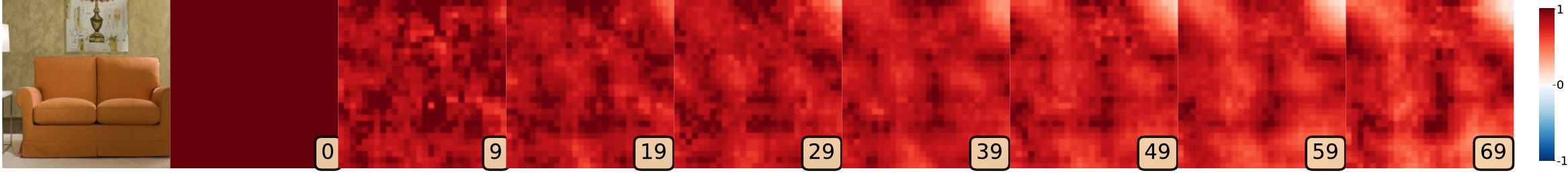}\\
		\includegraphics[width=0.8\linewidth]{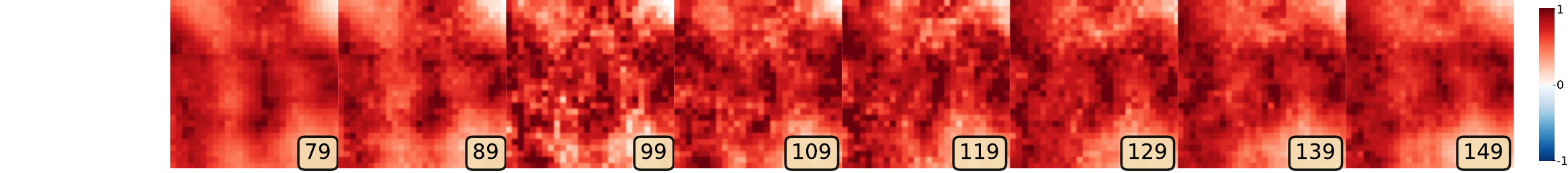}\\
		\includegraphics[width=0.8\linewidth]{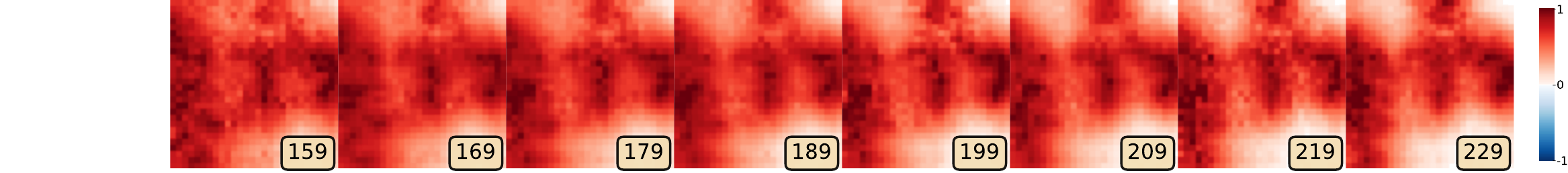}\\
		\includegraphics[width=0.8\linewidth]{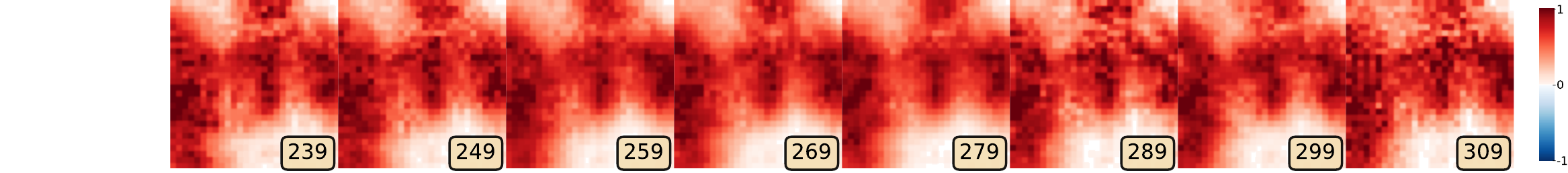}\\
		\includegraphics[width=0.8\linewidth]{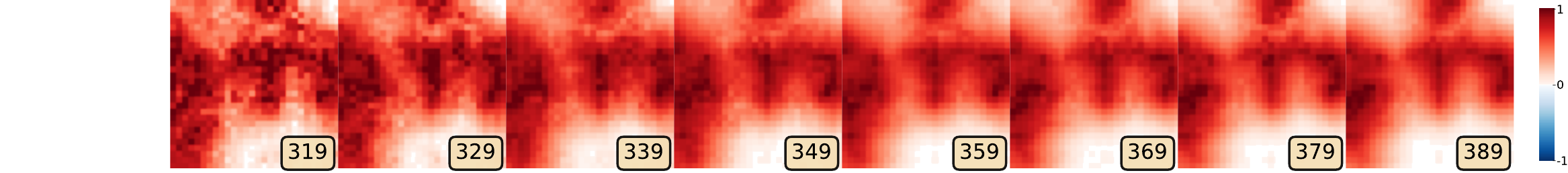}\\
		\includegraphics[width=0.8\linewidth]{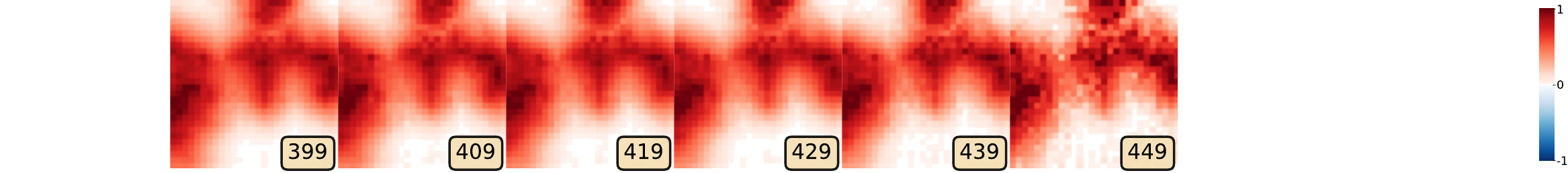}\\	
		\caption{
			ResNet
		}
		\label{fig:appendix_MP_iteration_ResNet_1}
	\end{subfigure}
	
	\begin{subfigure}{1.0\linewidth}
		\centering
		\includegraphics[width=0.8\linewidth]{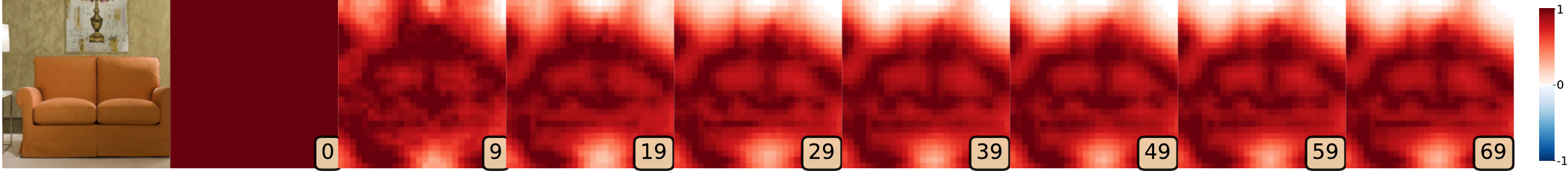}\\
		\includegraphics[width=0.8\linewidth]{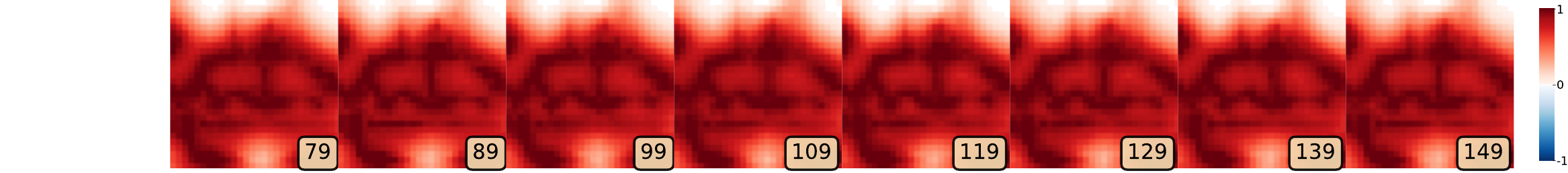}\\
		\includegraphics[width=0.8\linewidth]{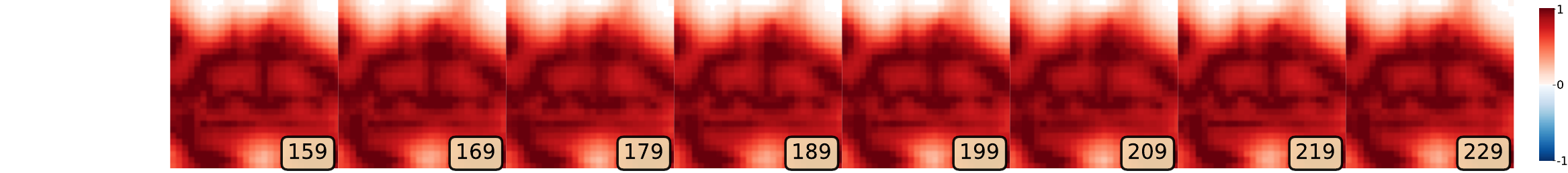}\\
		\includegraphics[width=0.8\linewidth]{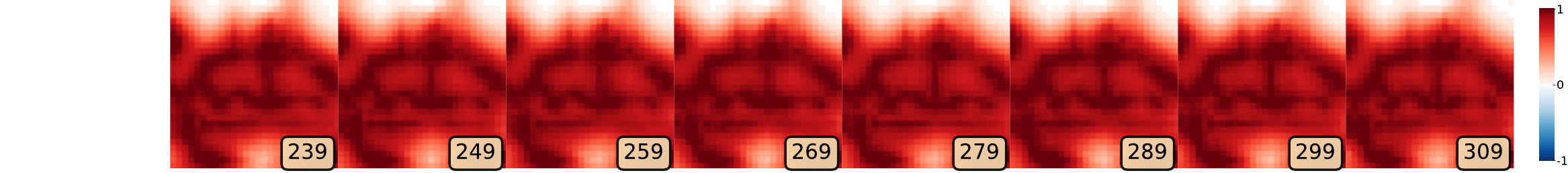}\\
		\includegraphics[width=0.8\linewidth]{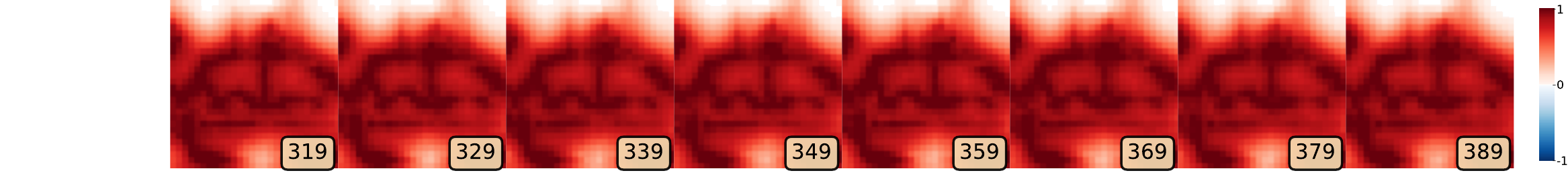}\\
		\includegraphics[width=0.8\linewidth]{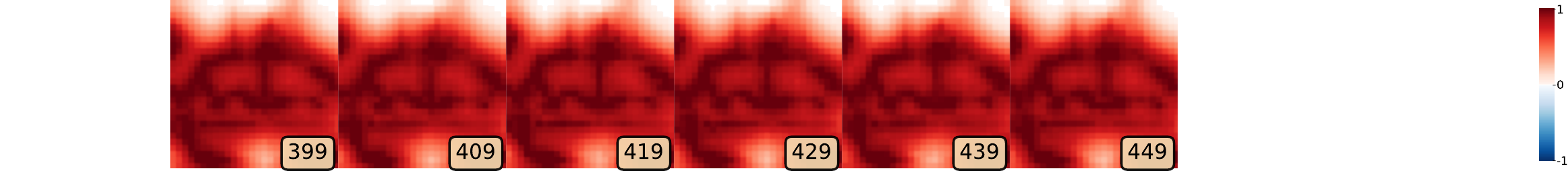}\\	
		\caption{
			ResNet-R
		}
		\label{fig:appendix_MP_iteration_ResNet_R_1}
	\end{subfigure}
	\caption{
		Evolution of attribution maps generated from a 450-step MP optimization run for a \class{studio~couch} image using ResNet (a) and ResNet-R (b) models.
		This figure is an extension of Fig.~\ref{fig:MP_iter_qual}.
		The attribution maps for ResNet-R model (b) converges to the optimum mask in just $\sim10$ iterations whereas the mask in the ResNet model are very inconsistent and keep fluctuating among differernt iterations. 
		For instance, the ResNet (a) masks becomes noisy iteration $289$, $309$, $319$, and $449$ despite being stable at $209$, $379$ and $409$ iterations.
        These qualitatively heatmaps are consistent with the quantitative loss-over-iteration plots (Fig.~\ref{fig:MP_loss}) where the ResNet loss curve oscillates while the ResNet-R curve converges early.
%        \naman{I think we need to put some sort of footnote here saying that 300 ones won't match because of the replication issues due to upsampling/pytorch on cuda}
	}
	\label{fig:appendix_MP_iteration}
\end{figure*}

\begin{figure*}[h]
	\centering
	{	
		\small
		\begin{flushleft}
			\hspace{7.1cm}Input image
			\hspace{0.4cm}$5\times5$
			\hspace{0.5cm}$17\times17$
			\hspace{0.4cm}$29\times29$
			\hspace{0.4cm}$41\times41$
			\hspace{0.4cm}$53\times53$
			\hspace{-9.3cm}\rotatebox{90}{\hspace{-3.6cm}ResNet-R\hspace{1.0cm}ResNet}				
		\end{flushleft}		
	}
	\vspace{-0.9cm}
	\subcaptionbox{Average similarity of heatmaps across the entire dataset when the patch size changes.\label{fig:SP_sens_quant}}
	[0.3\linewidth]{
		\includegraphics[width=0.3\textwidth]{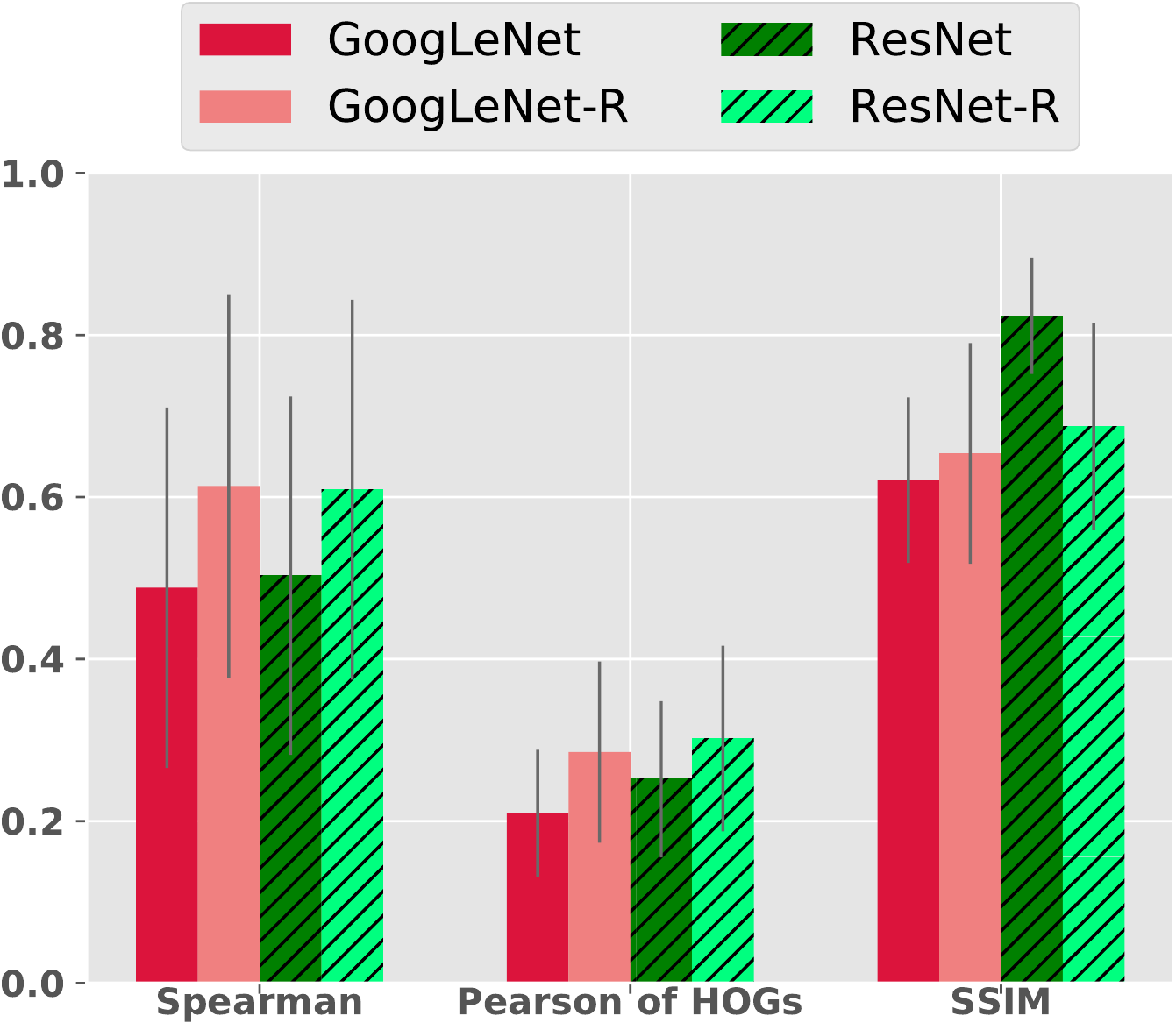}}
	\hspace{1.0cm}
	\subcaptionbox{SP heatmaps for ResNet \& ResNet-R change entirely for different patch sizes.\label{fig:SP_sens_qual}}
	[0.55\linewidth]{
		\includegraphics[width=0.55\textwidth]{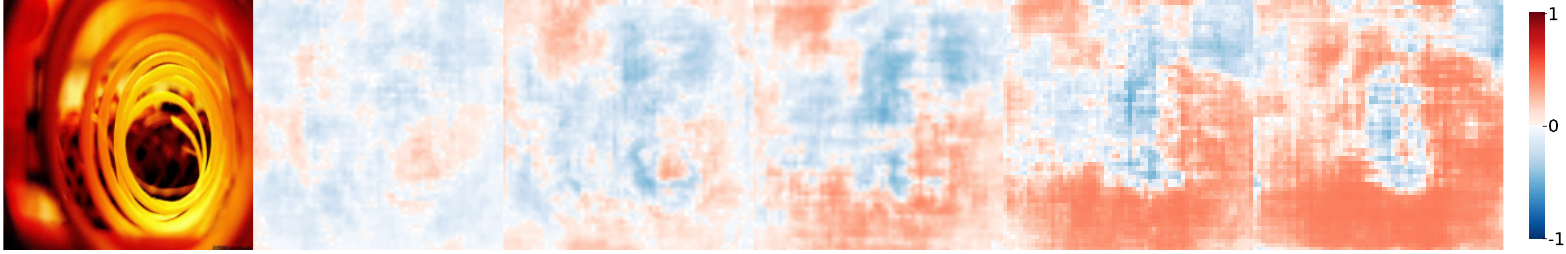}
		\text{HOG:$0.1095$}
		\includegraphics[width=0.55\textwidth]{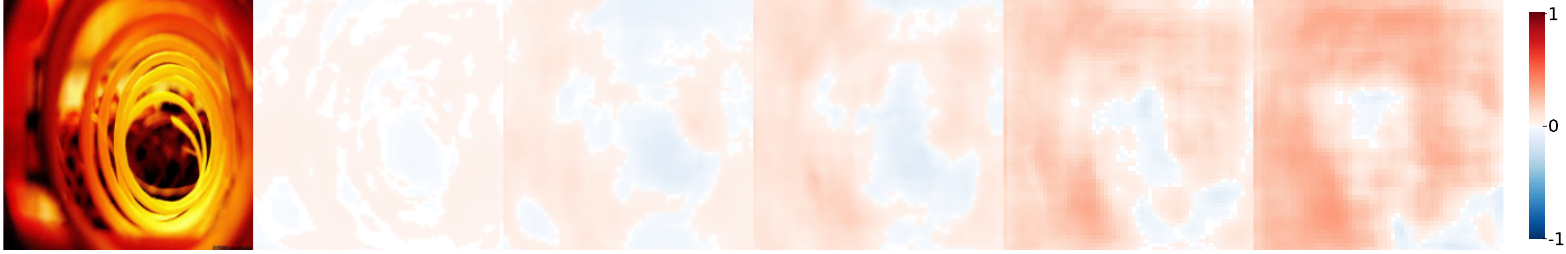}
		%		\text{HOG:$0.1095 $}
		\text{HOG:$0.2203$}
	}
	
	\caption{
		Sliding Patch (SP) attribution maps vary largely---both quantitatively (a) and qualitatively (b)---when \textbf{the patch size} changes.
		The stride was $3$ for all cases.
		\newline\textbf{Right panel:} As the patch size increases, we observe the attribution values are higher (\ie higher-intensity heatmaps), for both ResNet and ResNet-R (b).
		\newline\textbf{Left panel:}
		On average, across the dataset, we observe low similarity, under all three metrics, across the generated heatmaps (for both ResNet and ResNet-R) when the patch size changes (a).
%		On increasing patch sizes, we observe more important pixels in the attribution maps (b).
		See Fig.~\ref{fig:appendix_SP} for more examples of this behavior.
%        \naman{From a reading point of view, S11 should be S12 and S12 should be S11. I had changed it like this previously but you can decide.}
%		Quantitative (a) and Qualitative (b) figures show the sensitivity of Sliding Patch (SP) attribution maps on changing the patch size, $p \times p$. 
%		%		The stride, $s$, was fixed to $3$ for all cases.
%		\textbf{(a)}: SP maps are highly sensitive to the patch size across all models and metrics.
%		\textbf{(b)}: Across the dataset, the reference image caused the largest difference between the Pearson correlation of HOG features of ResNet heatmaps vs. ResNet-R heatmaps.
%		For increasing patch sizes, we observe larger regions as important pixels in the attribution maps for both ResNet \& ResNet-R.
%		See Fig.~\ref{fig:appendix_SP} for more examples of this behavior.
		%		\todo{If time permits, add two rows like the the row above, but one for an image of a small tennis ball, one for an image of a close-up tennis ball. This would show that the ``right'' patch size depends on the object size.}
	}
	%		\todo{remove bottom row} \todo{add SSIM scores to all qualitative figures but in SP add HOG} \todo{add figure for smalled set of patch sizes} \todo{add HOG edge images in bottom row}}
	\label{fig:SP_sensitivity}
\end{figure*}

\begin{figure*}[t]
	\centering
	{	
		\small
		\begin{flushleft}
			\hspace{7.2cm}(i)
			\hspace{1.15cm}(ii)
			\hspace{0.8cm}(iii)
			\hspace{1.0cm}(iv)
			\hspace{1.0cm}(v)
			\hspace{0.9cm}(vi)
			\hspace{0.9cm}\vspace{0.5cm}(vii)
		\end{flushleft}		
	}
	\vspace{-1.05cm}
	{	
		\small
		\vspace*{-0.10cm}
		\begin{flushleft}
			\hspace{6.37cm}\rotatebox{90}{\hspace{-3.35cm}ResNet-R\hspace{0.8cm}ResNet}				
			\hspace{0.05cm}Input image
			\hspace{0.01cm}$N_{SG}$=$50$
			\hspace{0.05cm}$N_{SG}$=$100$
			\hspace{0.05cm}$N_{SG}$=$200$
			\hspace{0.05cm}$N_{SG}$=$500$
			\hspace{0.05cm}$N_{SG}$=$800$
			\hspace{0.1cm}Diff (vi -- ii)
		\end{flushleft}		
	}	
	\vspace{-1.0cm}
	\subcaptionbox{Average robustness across the dataset\label{fig:SG_sample_quant}}
	[0.30\linewidth]{
		\includegraphics[width=0.30\textwidth]{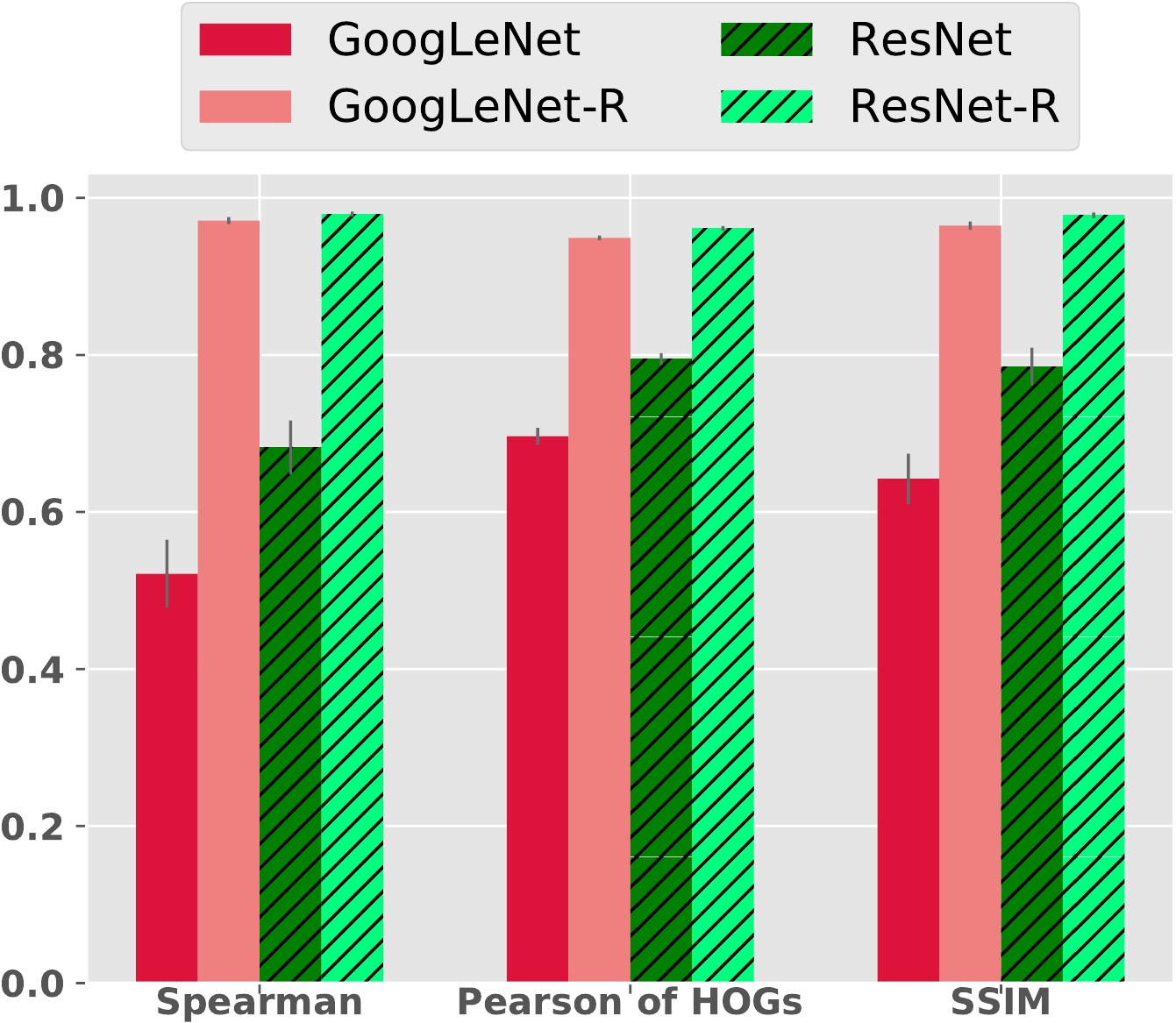}}
	\hspace{1.0cm}
	\subcaptionbox{SG heatmaps for ResNet-R are more consistent compared to those for ResNet. \label{fig:SG_sample_qual}}
	[0.6\linewidth]{
		\includegraphics[width=0.6\textwidth]{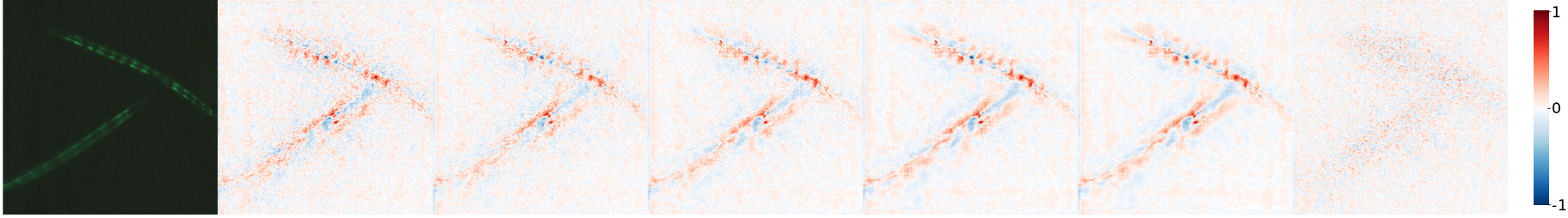}
		%			\text{SSIM: $0.3633 $}
		\text{\hspace{4cm}SSIM: $0.3633 $\hspace{2.5cm}\hfill $L_1: 2686.75$}
		\includegraphics[width=0.6\textwidth]{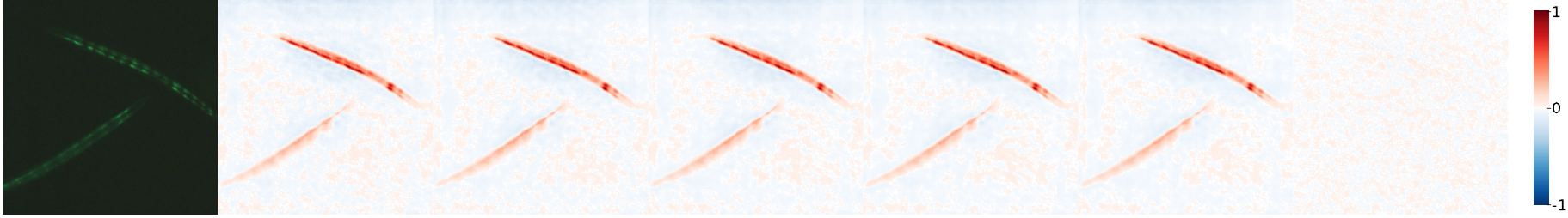}
		\text{\hspace{4cm}SSIM: $0.9167 $\hspace{2.8cm}\hfill $L_1: 480.13$}
	}
	\caption{
		On average, across the dataset, SmoothGrad explanations for robust classifiers are almost perfectly consistent upon varying the sample size $N_{SG} \in \{50, 100, 200, 500, 800\}$ \ie GoogLeNet-R and ResNet-R similarity scores are near 1.0 (a).
		However, the same heatmaps for regular classifiers are substantially more sensitive (a).
		We show here the input image (i) that yields the largest difference (among the dataset) between the SSIM score for ResNet-R heatmaps (0.9167) and that for ResNet heatmaps (0.3633).
		While SG heatmaps may appear qualitatively consistent, the pixel-wise variations (\eg see column vii---the results of subtracting ii from vi) may cause issues for applications that require pixel-level precision.
		%
		%			Quantitative (a) and Qualitative (b) figures show the sensitivity of SmoothGrad (SG) attribution maps on changing $N_{SG}$ samples hyperparameter. 
		%		The standard deviation, $\sigma$, of the Gaussian noise was fixed to 0.15 for all cases.
		%			\textbf{(a)}: SG maps are sensitive to $N_{SG}$ across both regular models. 
		%			Interestingly, the sensitivity scores for the robust models are $\sim1.0$. 
		%			\textbf{(b)}: Across the dataset, the reference image caused the largest difference between the SSIM scores of ResNet heatmaps vs. ResNet-R heatmaps.
		%			A higher SSIM (0.9167) is observed between the ResNet-R attribution maps (row 2; (ii)-(vi)) for different $N_{SG}$'s as compared to the SSIM (0.3633) in ResNet (row 1; (ii)-(vi)) case. The last column (vii) shows the difference heatmap i.e the difference between the attribution map of $N_{SG}=800$ (vi) and $N_{SG}=50$ (ii). 
		%			$L_1$ score of the difference heatmap is comparatively higher for ResNet ($L_{1}=2686.75$) than ResNet-R ($L_{1}=480.13$).
		%		\todo{add the difference images between $N_{SG}=800$ and $N_{SG}=50$}
		%		\todo{add vertical model names}
		%		\todo{Find the bottom rows showing minimum variations}
		%		\todo{Choose matchstick}
	}
	\label{fig:SG_sample_sensitivity}
%	\vspace*{-0.4cm}
\end{figure*}

\begin{figure}[t]
	\centering
	\includegraphics[width=0.35\textwidth]{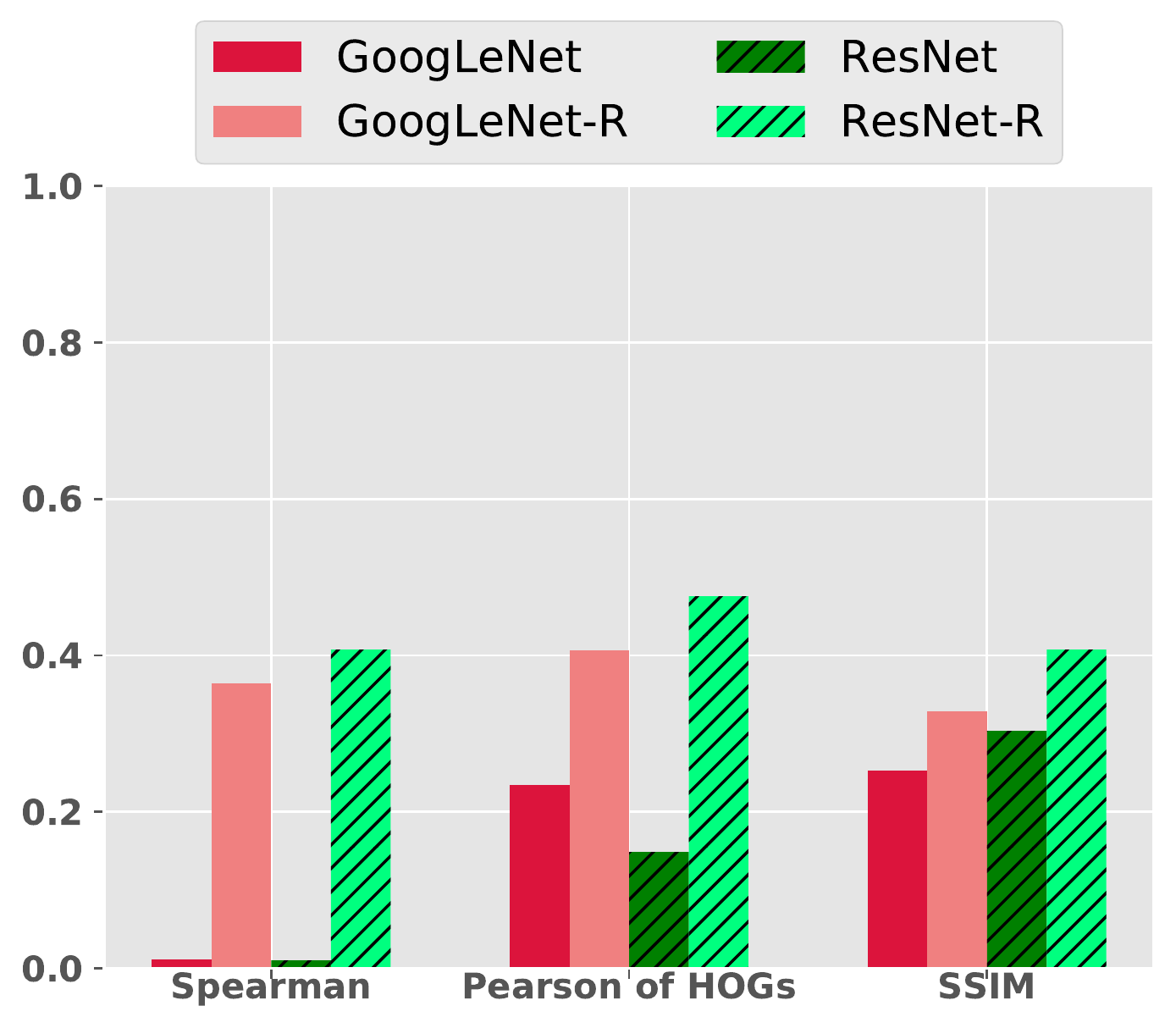}
	\caption{
		Compared to the gradients of regular classifiers (darker red and green), the gradients of robust classifiers (lighter red and green) are consistently more invariant before and after the addition of noise to the input image under all three similarity metrics (higher is better).}
	\label{fig:grad_sec_1_1_quant}
%	\vspace*{-0.4cm}	
\end{figure}

\begin{figure*}
	\centering
	{	
		\scriptsize
		\vspace*{-0.15cm}
		\begin{flushleft}
			\hspace{1.7cm}(a) Input image
			\hspace{0.2cm}(b) Grad
			\hspace{0.1cm}(c) $N_{SG}$=$50$
			\hspace{0.1cm}(d) $N_{SG}$=$100$
			\hspace{0.1cm}(e) $N_{SG}$=$200$
			\hspace{0.1cm}(f) $N_{SG}$=$500$
			\hspace{0.1cm}(g) $N_{SG}$=$800$
			\hspace{0.1cm}(h) GB \cite{springenberg2014striving}					
			\hspace{0.1cm}(i) Robust Grad	
		\end{flushleft}		
	}
	\vspace{-0.3cm}
	\includegraphics[width=0.8\textwidth]{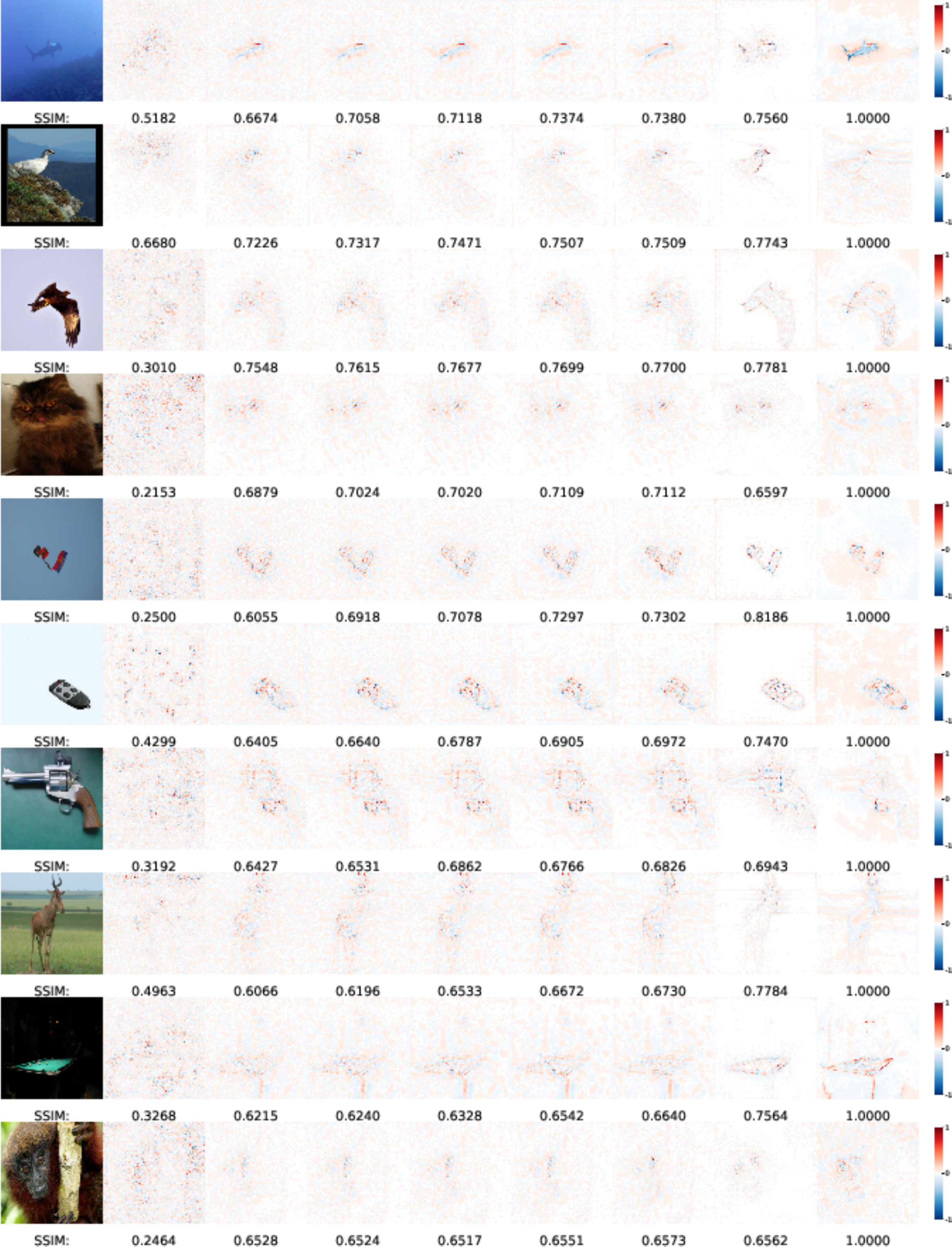}
	\caption{
		This figure is an extension to Fig.~\ref{fig:SG_trend_qual}.
		Qualitative trend showing the increase in similarity between the attribution maps from SmoothGrad (SG) of ResNet (c---g) and vanila gradient (Grad) of ResNet-R (i) as \textbf{the number of samples \boldsymbol{$N_{SG}$} increases}.
		Below each heatmap is the SSIM similarity score between that heatmap and the heatmap in column (i).
		As the sample size $N_{SG}$ increases, SG attribution maps of ResNet become increasingly more similar, under SSIM, to the gradient heatmaps of ResNet-R, a completely different network.
		Additionally, by comparing column (h) and (i), one might conclude that ResNet and ResNet-R behave similarly (because the heatmaps are similar both qualitatively and quantitatively under SSIM).
		However, these are two completely distinct networks with different training regimes and their differences can be seen by comparing column (b) and (i).
		In sum, de-noising heatmaps, \eg using SG or GB, may cause misinterpretation.
%		
%		GuidedBackprop (GB) \cite{springenberg2014striving} attribution map for the respective images using ResNet (h).
%		
%		Starting from the attribution map from VG of ResNet (b) we observe the increase in SSIM scores with increasing $N_{SG}$.
%		Across the dataset, the reference images had the maximum SSIM score between the attribution map from SG, using 800 samples (g), of ResNet and VG of ResNet-R (i).
%		On average, the SSIM score between the attribution map from GuidedBackprop of ResNet and VG of ResNet-R is the highest. \naman{I think, we should mention the avg value as well. 0.377 is the value}
	}
	\label{fig:appendix_guided_top10}
\end{figure*}

\begin{figure*}
	\centering
	{	
		\small
		\vspace*{-0.15cm}
		\begin{flushleft}
			\hspace{2.2cm}(a) Input image
			\hspace{0.4cm}(b) $5 \times 5$
			\hspace{0.6cm}(c) $17 \times 17$
			\hspace{0.7cm}(d) $29 \times 29$
			\hspace{0.5cm}(e) $41 \times 41$
			\hspace{0.6cm}(f) $53 \times 53$
		\end{flushleft}		
	}
 	\vspace{-0.3cm}
	\includegraphics[width=0.75\textwidth]{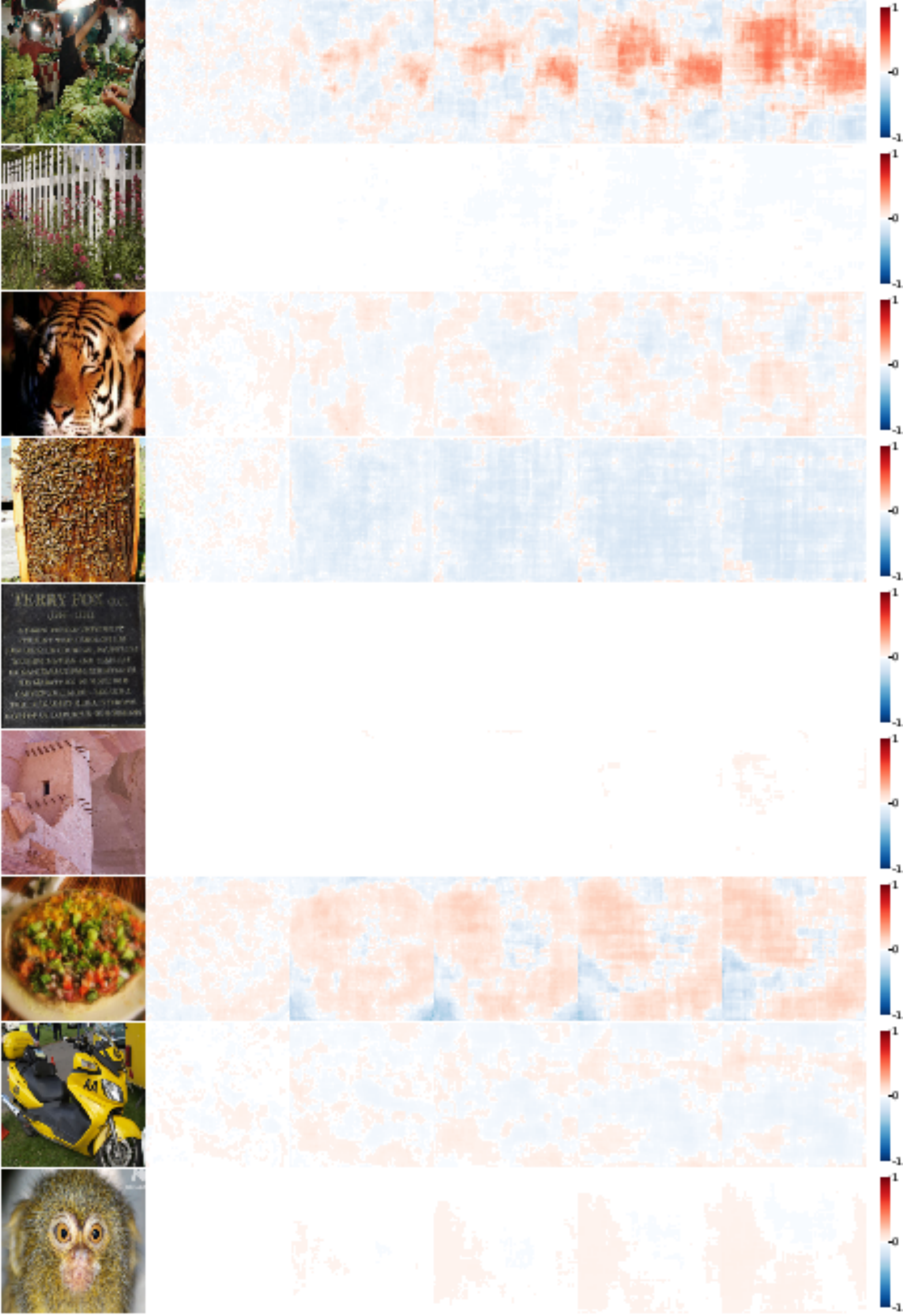}
	\caption{
		Sliding-Patch (SP) attribution maps are very sensitive to different \textbf{patch sizes} (Sec.~\ref{sec:SP_exp}).
		Across the dataset, the reference images had the lowest Pearson correlation of HOG features among the ResNet heatmaps.
%		Patches at multiple location outside the target object drop the target class probability resulting in unexplainable and noisy heatmaps. 
		For some images with huge objects (\eg the image of a white fence in row 2), we do not observe any significant probability drop even for a patch size of $53 \times 53$ (f) and hence the attribution values are almost zero.
		This observation underlines an important challenge of choosing the right patch size when using SP.
	}
	\label{fig:appendix_SP}
\end{figure*}

%\begin{figure*}
%	\centering
%	{	
%		\small
%		\vspace*{-0.15cm}
%		\begin{flushleft}
%			\hspace{4.7cm}(a) Input image
%			\hspace{0.3cm}(b) $b_R$=$5$
%			\hspace{0.6cm}(c) $b_R$=$10$
%			\hspace{0.6cm}(d) $b_R$=$30$
%%			\hspace{-12.1cm}\rotatebox{90}{\hspace{-18.8cm}ResNet-R\hspace{0.8cm}ResNet\hspace{0.8cm}ResNet-R\hspace{0.8cm}ResNet
%%			\hspace{0.8cm}ResNet-R\hspace{0.8cm}ResNet\hspace{0.8cm}ResNet-R\hspace{0.8cm}ResNet\hspace{0.8cm}ResNet-R\hspace{0.8cm}ResNet}
%		\end{flushleft}		
%	}
%	\vspace{-0.3cm}
%	\includegraphics[width=0.95\textwidth]{images/blur_radius_MP_resnet.pdf}
%	\text{ResNet}\\
%	\includegraphics[width=0.95\textwidth]{images/blur_radius_MP_resnet_r.pdf}
%	\text{ResNet-R}
%%	\includegraphics[width=0.45\textwidth]{images/appendix_MP_blur_comparison.pdf}
%	\caption{
%		Attribution maps become more scattered as we increase the Gaussian blur radius ($b_R$) in the MP sensitivity experiment.
%		The effect is explicitly seen for the ResNet attribution maps whereas the ResNet-R maps are comparatively stable.
%		Across the dataset, the reference images were randomly chosen for the ResNet \& ResNet-R model pairs.
%	}
%	\label{fig:appendix_MP_blur}
%\end{figure*}

\begin{figure*}
	\centering
		{	
			\small
			\vspace*{-0.15cm}
			\begin{flushleft}
				\hspace{0.75cm}Input
				\hspace{0.6cm}$b_R$=$5$
				\hspace{0.4cm}$b_R$=$10$
				\hspace{0.5cm}$b_R$=$30$
				\hspace{0.6cm}Input
				\hspace{0.6cm}$b_R$=$5$
				\hspace{0.4cm}$b_R$=$10$
				\hspace{0.5cm}$b_R$=$30$
				\hspace{0.7cm}Input
				\hspace{0.6cm}$b_R$=$5$
				\hspace{0.4cm}$b_R$=$10$
				\hspace{0.5cm}$b_R$=$30$
			\end{flushleft}
		}
	\vspace{-0.3cm}
	\begin{subfigure}{1.0\linewidth}
		\centering
		\includegraphics[width=0.95\linewidth]{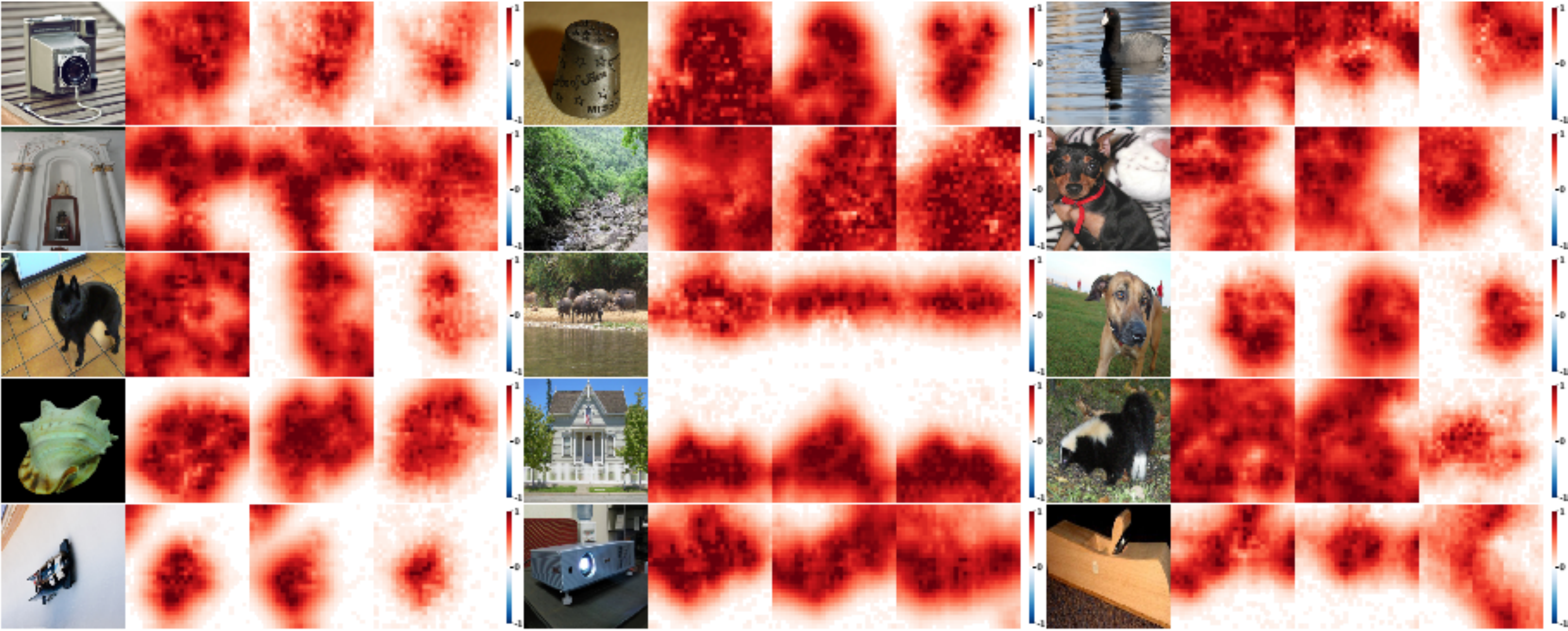}\\
		\caption{
			ResNet
		}
		\label{fig:appendix_MP_blur_ResNet}
	\end{subfigure}
	
	\begin{subfigure}{1.0\linewidth}
		\centering
		\includegraphics[width=0.95\linewidth]{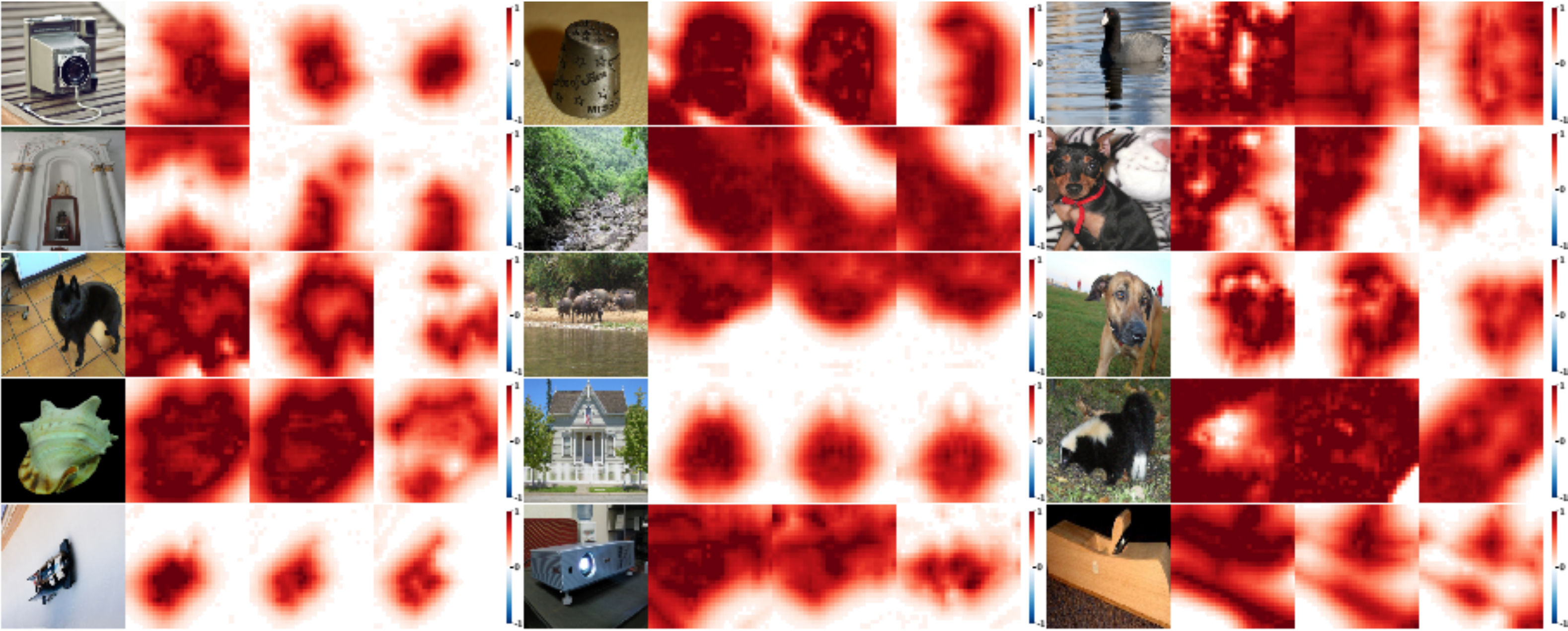}\\			
		\caption{
			ResNet-R
		}
		\label{fig:appendix_MP_blur_ResNet_R}
	\end{subfigure}
	\caption{
		Attribution maps of ResNet (a) become more scattered as we increase the \textbf{Gaussian blur radius \boldsymbol{$b_R$}} (from left to right) in the MP sensitivity experiment (Sec.~\ref{sec:MP_exp}).
		In contrast, for ResNet-R, the attribution maps become smoother as the blur radius increases.
		The reference images here were randomly chosen.
	}
	\label{fig:appendix_MP_blur}
\end{figure*}

\begin{figure*}
	\centering
	\begin{subfigure}[b]{0.49\textwidth}
		{	
			\small
			\begin{flushleft}
				\hspace{0.35cm}Input
				\hspace{0.7cm}500
				\hspace{0.7cm}1000
			\end{flushleft}
		}
		\vspace{-0.3cm}	
		\includegraphics[width=0.51\textwidth]{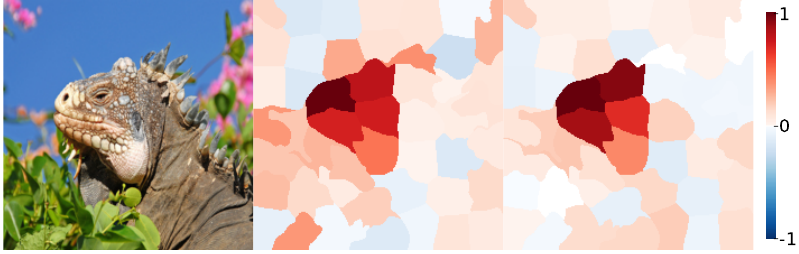}
		{	
			\small
			\vspace{-0.3cm}			
			\begin{flushleft}
				\hspace{0.35cm} LIME sample size | SSIM: 0.0602
			\end{flushleft}
			\vspace{-0.3cm}
		}
		{	
			\small
			\vspace*{-0.3cm}
			\begin{flushleft}
				\hspace{0.35cm}Input
				\hspace{0.6cm}seed=0
				\hspace{0.45cm}seed=1
				\hspace{0.45cm}seed=2
				\hspace{0.45cm}seed=3
				\hspace{0.45cm}seed=4												
			\end{flushleft}
		}
		\vspace{-0.3cm}
		\includegraphics[width=\textwidth]{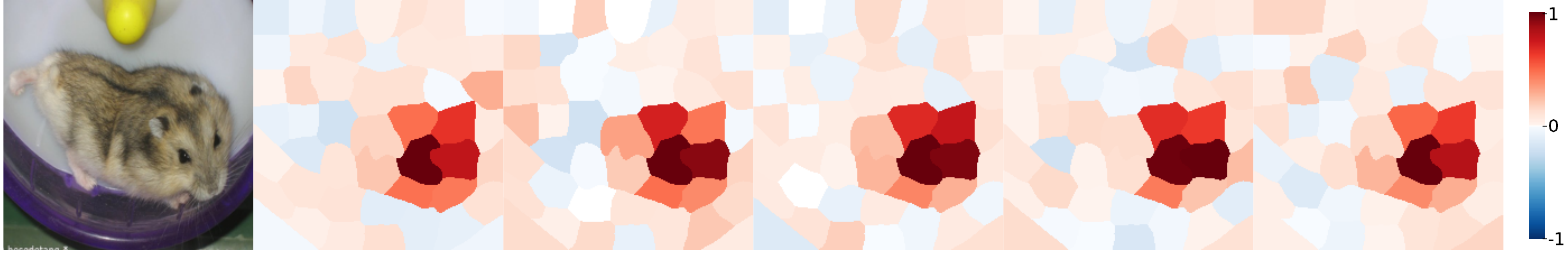}
		{	
			\small
			\vspace{-0.6cm}			
			\begin{flushleft}
				\hspace{1.6cm} LIME random seed | SSIM: 0.2148
			\end{flushleft}
		}
		{	
			\small
			\vspace*{-0.65cm}
			\begin{flushleft}
				\hspace{0.35cm}Input
				\hspace{0.3cm}$N_{iter}$=$10$
				\hspace{0.1cm}$N_{iter}$=$150$
				\hspace{0.01cm}$N_{iter}$=$300$
				\hspace{0.01cm}$N_{iter}$=$450$
			\end{flushleft}
		}
		\vspace{-0.3cm}
		\includegraphics[width=0.84\textwidth]{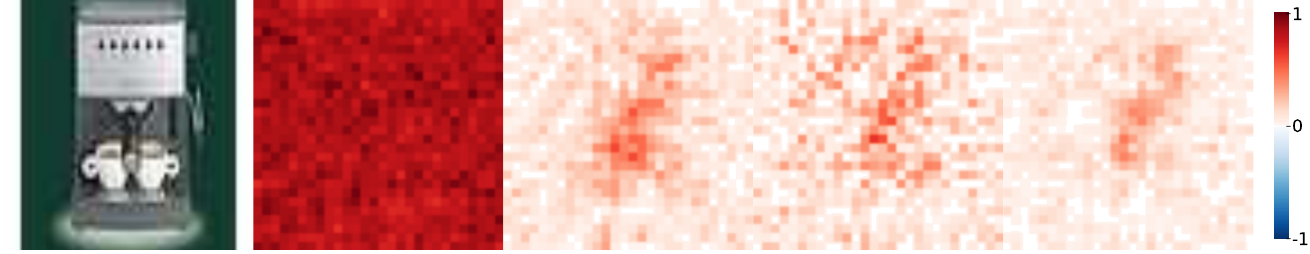}
		{	
			\small
			\vspace{-0.3cm}
			\begin{flushleft}
				\hspace{1.3cm} MP number of iterations | SSIM: 0.2869
			\end{flushleft}
			\vspace{-0.3cm}		
		}		
		{	
			\small
			\vspace*{-0.3cm}
			\begin{flushleft}
				\hspace{0.3cm}Input
				\hspace{0.75cm}$b_R$=$5$
				\hspace{0.55cm}$b_R$=$10$
				\hspace{0.6cm}$b_R$=$30$
			\end{flushleft}
		}
		\vspace{-0.3cm}	
		\includegraphics[width=0.68\textwidth]{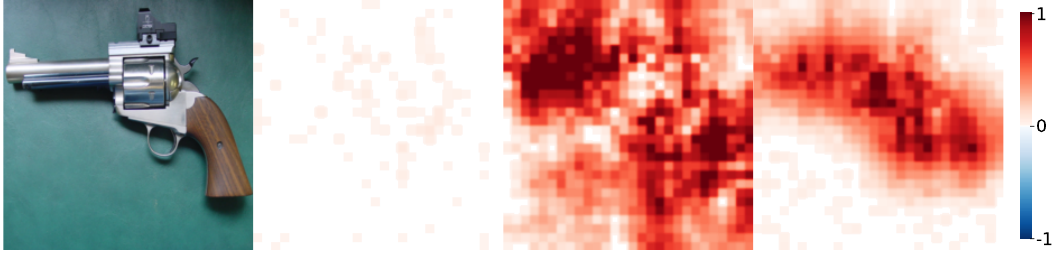}
		{	
			\small
			\vspace{-0.3cm}
			\begin{flushleft}
				\hspace{1.2cm} MP blur radius | SSIM: 0.1411
			\end{flushleft}
			\vspace{-0.3cm}		
		}
		{	
			\small
			\vspace*{-0.3cm}
			\begin{flushleft}
				\hspace{0.35cm}Input
				\hspace{0.5cm}seed=0
				\hspace{0.45cm}seed=1
				\hspace{0.45cm}seed=2
				\hspace{0.45cm}seed=3
				\hspace{0.45cm}seed=4												
			\end{flushleft}
		}
		\vspace{-0.3cm}
		\includegraphics[width=\textwidth]{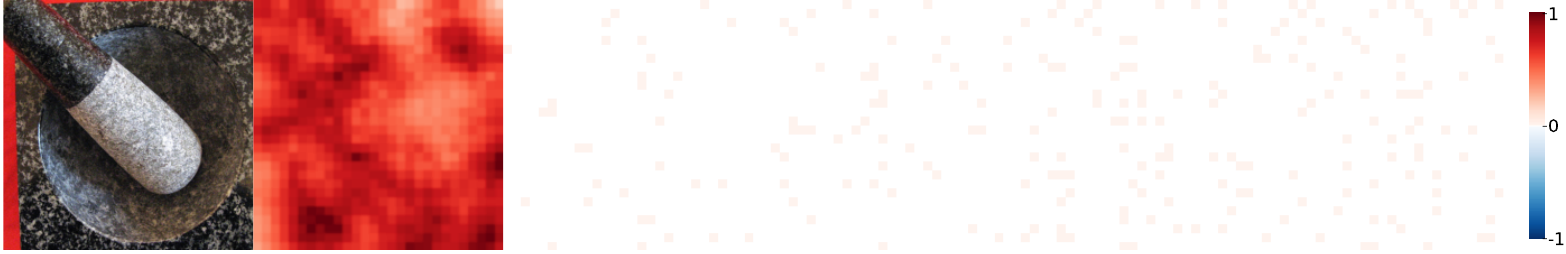}
		{	
			\small
			\vspace{-0.6cm}			
			\begin{flushleft}
				\hspace{2.0cm} MP random seed | SSIM: 0.0008
			\end{flushleft}
		}		
		{	
			\small
			\vspace*{-0.6cm}	
			\begin{flushleft}
				\hspace{0.35cm}Input
				\hspace{0.45cm}$N_{SG}$=$50$
				\hspace{0.05cm}$N_{SG}$=$100$
				\hspace{0.05cm}$N_{SG}$=$200$
				\hspace{0.05cm}$N_{SG}$=$500$
				\hspace{0.05cm}$N_{SG}$=$800$
			\end{flushleft}		
		}	
		\vspace{-0.3cm}
		\includegraphics[width=\textwidth]{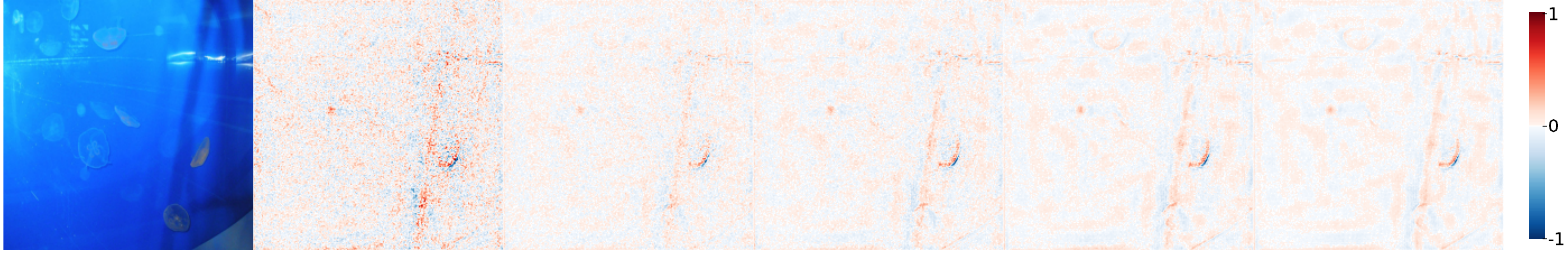}
		{	
			\small
			\vspace{-0.6cm}			
			\begin{flushleft}
				\hspace{1.8cm} SG sample size | SSIM: 0.2127
			\end{flushleft}
		}
		{	
			\small
			\vspace*{-0.6cm}
			\begin{flushleft}
				\hspace{0.3cm}Input
				\hspace{0.75cm}$\sigma$=$0.1$
				\hspace{0.55cm}$\sigma$=$0.2$
				\hspace{0.6cm}$\sigma$=$0.3$
			\end{flushleft}
		}
		\vspace{-0.3cm}			
		\includegraphics[width=0.68\textwidth]{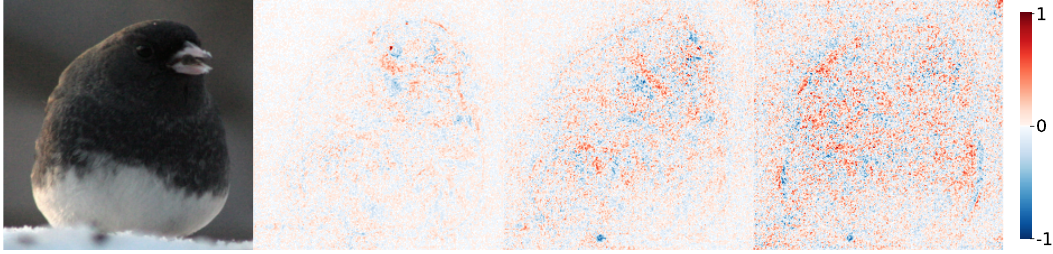}
		{	
			\small
			\vspace{-0.3cm}			
			\begin{flushleft}
				\hspace{1.0cm} SG Gaussian noise $\sigma$ | SSIM: 0.1182
			\end{flushleft}
			\vspace{-0.3cm}		
		}	
		{	
			\small
			\vspace*{-0.3cm}
			\begin{flushleft}
				\hspace{0.35cm}Input
				\hspace{0.7cm}$5\times5$
				\hspace{0.4cm}$17\times17$
				\hspace{0.3cm}$29\times29$
				\hspace{0.3cm}$41\times41$
				\hspace{0.2cm}$53\times53$										
			\end{flushleft}
		}
		\vspace{-0.3cm}
		\includegraphics[width=\textwidth]{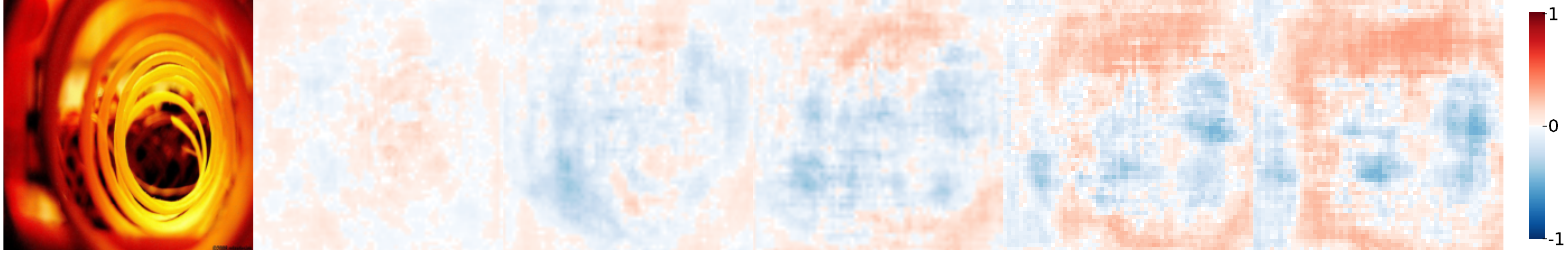}
		{	
			\small
			\vspace{-0.6cm}			
			\begin{flushleft}
				\hspace{1cm} SP-L: large patch size changes | SSIM: 0.1939
			\end{flushleft}
		}
		{	
			\small
			\vspace*{-0.6cm}
			\begin{flushleft}
				\hspace{0.35cm}Input
				\hspace{0.55cm}$52\times52$														
				\hspace{0.25cm}$53\times53$
				\hspace{0.3cm}$54\times54$														
			\end{flushleft}
		}
		\vspace{-0.3cm}		
		\includegraphics[width=0.68\textwidth]{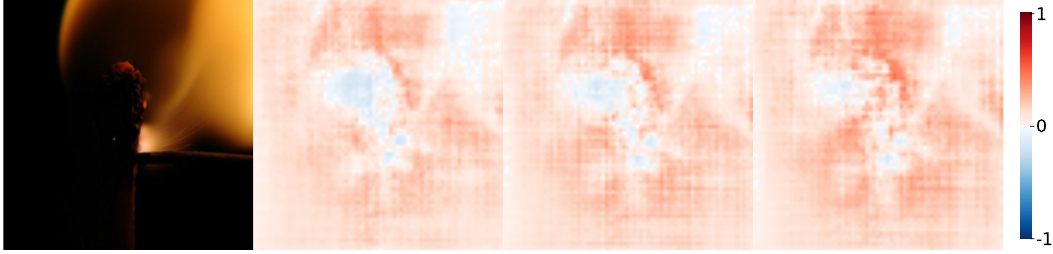}
		{	
			\small
			\vspace{-0.3cm}			
			\begin{flushleft}
				\hspace{0.5cm} SP-S: small patch size changes | SSIM: 0.8007
			\end{flushleft}
		}
		\caption{
			GoogleNet
		}
		\label{fig:googlenet}
	\end{subfigure}
	\begin{subfigure}[b]{0.49\textwidth}
		{	
			\small
			\begin{flushleft}
				\hspace{0.35cm}Input
				\hspace{0.8cm}500
				\hspace{0.8cm}1000
			\end{flushleft}
		}
		\vspace{-0.3cm}		
		\includegraphics[width=0.51\textwidth]{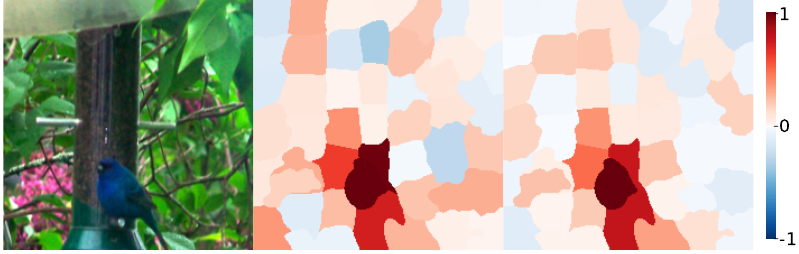}
		{	
			\small
			\vspace{-0.3cm}
			\begin{flushleft}
				\hspace{0.35cm} LIME sample size | SSIM: 0.0918
			\end{flushleft}
			\vspace{-0.3cm}		
		}		
		{	
			\small
			\vspace*{-0.3cm}
			\begin{flushleft}
				\hspace{0.35cm}Input
				\hspace{0.6cm}seed=0
				\hspace{0.45cm}seed=1
				\hspace{0.45cm}seed=2
				\hspace{0.45cm}seed=3
				\hspace{0.45cm}seed=4												
			\end{flushleft}
		}
		\vspace{-0.3cm}
		\includegraphics[width=\textwidth]{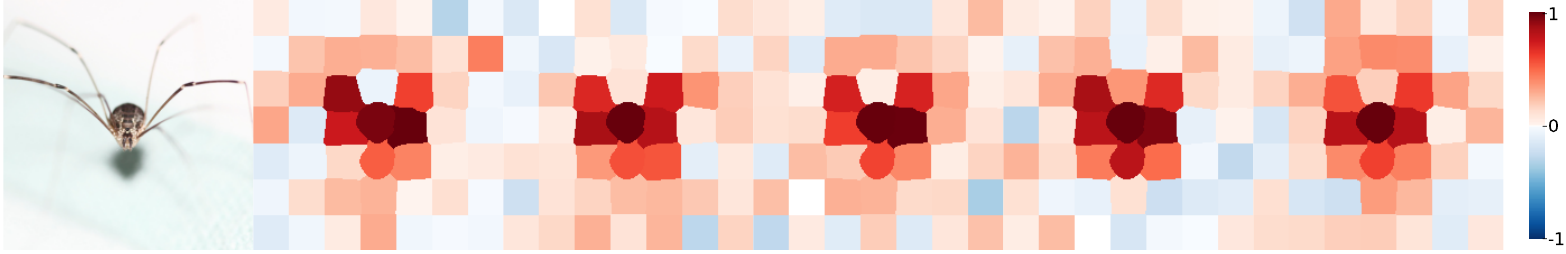}
		{	
			\small
			\vspace{-0.6cm}
			\begin{flushleft}
				\hspace{1.6cm} LIME random seed | SSIM: 0.1822
			\end{flushleft}
		}		
		{	
			\small
			\vspace*{-0.65cm}
			\begin{flushleft}
				\hspace{0.35cm}Input
				\hspace{0.3cm}$N_{iter}$=$10$
				\hspace{0.1cm}$N_{iter}$=$150$
				\hspace{0.01cm}$N_{iter}$=$300$
				\hspace{0.01cm}$N_{iter}$=$450$
			\end{flushleft}
		}
		\vspace{-0.3cm}
		\includegraphics[width=0.84\textwidth]{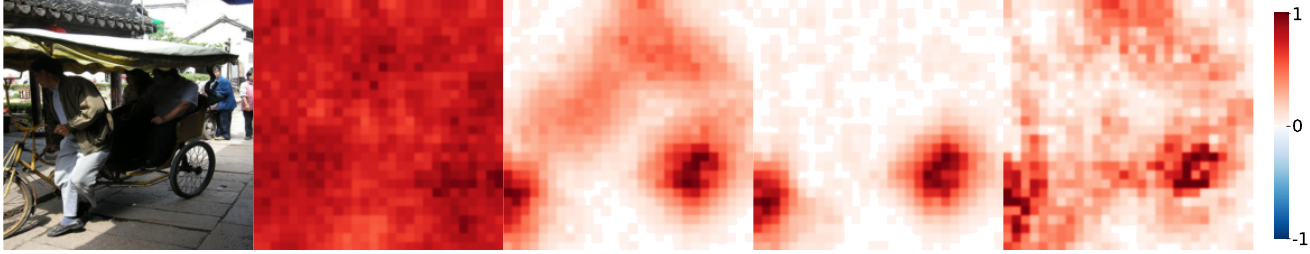}
		{	
			\small
			\vspace{-0.3cm}
			\begin{flushleft}
				\hspace{1.3cm} MP number of iterations | SSIM: 0.2676
			\end{flushleft}
			\vspace{-0.3cm}
		}		
		{	
			\small
			\vspace*{-0.3cm}
			\begin{flushleft}
				\hspace{0.3cm}Input
				\hspace{0.75cm}$b_R$=$5$
				\hspace{0.55cm}$b_R$=$10$
				\hspace{0.6cm}$b_R$=$30$
			\end{flushleft}
		}
		\vspace{-0.3cm}	
		\includegraphics[width=0.68\textwidth]{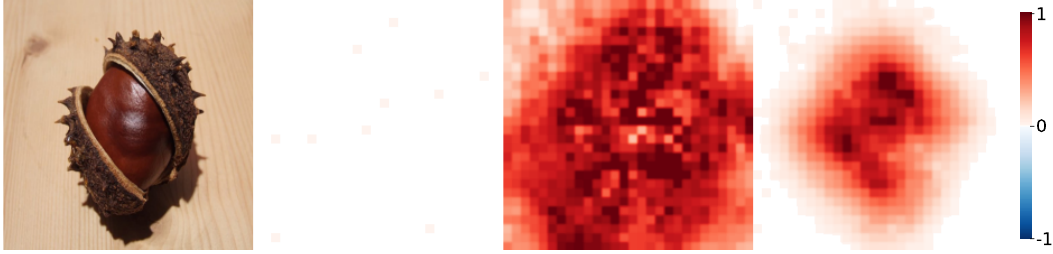}
		{	
			\small
			\vspace{-0.3cm}
			\begin{flushleft}
				\hspace{1.2cm} MP blur radius | SSIM: 0.0909
			\end{flushleft}
			\vspace{-0.3cm}
		}	
		{	
			\small
			\vspace*{-0.3cm}
			\begin{flushleft}
				\hspace{0.35cm}Input
				\hspace{0.5cm}seed=0
				\hspace{0.45cm}seed=1
				\hspace{0.45cm}seed=2
				\hspace{0.45cm}seed=3
				\hspace{0.45cm}seed=4												
			\end{flushleft}
		}
		\vspace{-0.3cm}
		\includegraphics[width=\textwidth]{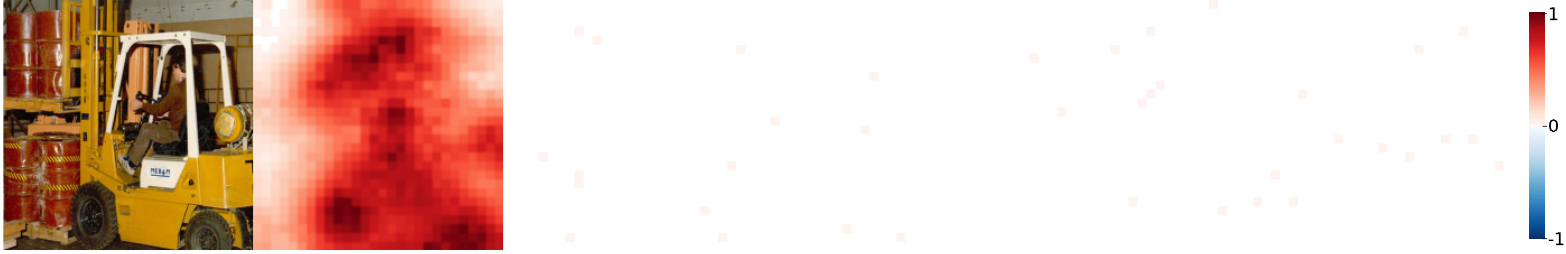}
		{	
			\small
			\vspace{-0.6cm}			
			\begin{flushleft}
				\hspace{2.0cm} MP random seed | SSIM: 0.0179
			\end{flushleft}
		}		
		{	
			\small
			\vspace*{-0.6cm}	
			\begin{flushleft}
				\hspace{0.35cm}Input
				\hspace{0.45cm}$N_{SG}$=$50$
				\hspace{0.05cm}$N_{SG}$=$100$
				\hspace{0.05cm}$N_{SG}$=$200$
				\hspace{0.05cm}$N_{SG}$=$500$
				\hspace{0.05cm}$N_{SG}$=$800$
			\end{flushleft}		
		}	
		\vspace{-0.3cm}
		\includegraphics[width=\textwidth]{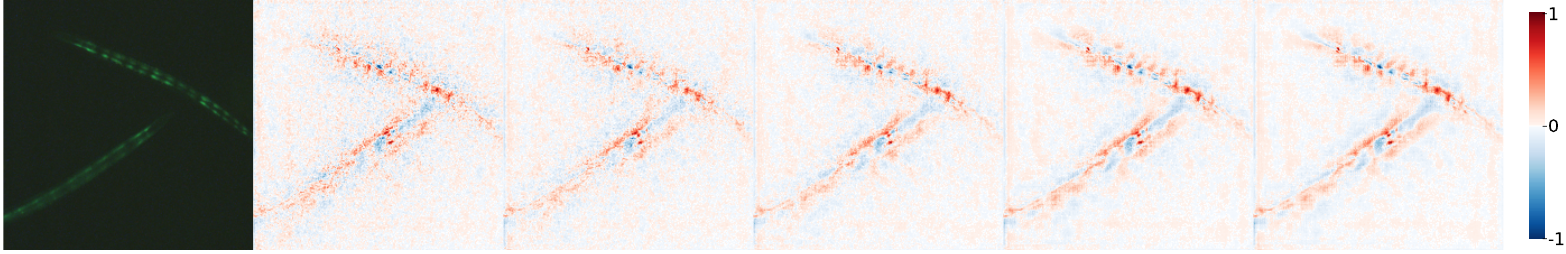}
		{	
			\small
			\vspace{-0.6cm}			
			\begin{flushleft}
				\hspace{1.8cm} SG sample size | SSIM: 0.3633
			\end{flushleft}
		}	
		{	
			\small
			\vspace*{-0.6cm}
			\begin{flushleft}
				\hspace{0.3cm}Input
				\hspace{0.75cm}$\sigma$=$0.1$
				\hspace{0.55cm}$\sigma$=$0.2$
				\hspace{0.6cm}$\sigma$=$0.3$
			\end{flushleft}
		}
		\vspace{-0.3cm}			
		\includegraphics[width=0.68\textwidth]{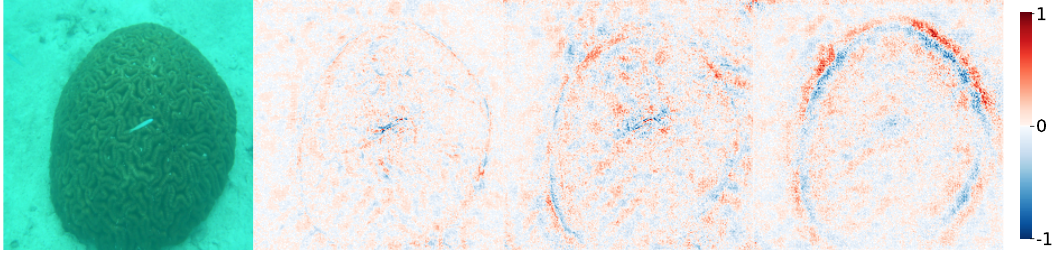}
		{	
			\small
			\vspace{-0.3cm}			
			\begin{flushleft}
				\hspace{1.0cm} SG Gaussian noise $\sigma$ | SSIM: 0.1399
			\end{flushleft}
			\vspace{-0.3cm}		
		}
		{	
			\small
			\vspace*{-0.3cm}
			\begin{flushleft}
				\hspace{0.35cm}Input
				\hspace{0.7cm}$5\times5$
				\hspace{0.4cm}$17\times17$
				\hspace{0.3cm}$29\times29$
				\hspace{0.3cm}$41\times41$
				\hspace{0.2cm}$53\times53$										
			\end{flushleft}
		}
		\vspace{-0.3cm}
		\includegraphics[width=\textwidth]{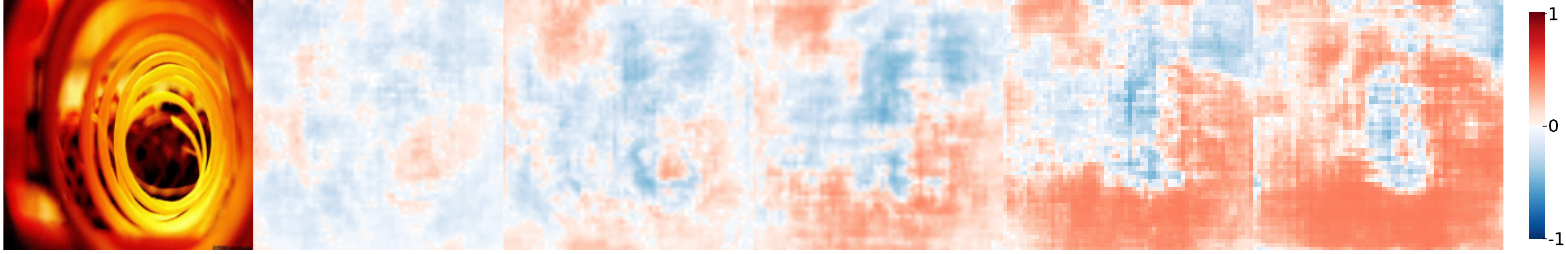}
		{	
			\small
			\vspace{-0.6cm}			
			\begin{flushleft}
				\hspace{1cm} SP-L: large patch size changes | SSIM: 0.2011
			\end{flushleft}
		}		
		{	
			\small
			\vspace*{-0.6cm}
			\begin{flushleft}
				\hspace{0.35cm}Input
				\hspace{0.55cm}$52\times52$														
				\hspace{0.25cm}$53\times53$
				\hspace{0.3cm}$54\times54$														
			\end{flushleft}
		}
		\vspace{-0.3cm}		
		\includegraphics[width=0.68\textwidth]{images/worst_case/worst_case_SPS_googlenet_ILSVRC2012_val_00042218.pdf}
		{	
			\small
			\vspace{-0.3cm}			
			\begin{flushleft}
				\hspace{0.5cm} SP-S: small patch size changes | SSIM: 0.7729
			\end{flushleft}
		}		
		\caption{
			ResNet
		}
		\label{fig:resnet}
	\end{subfigure}
	
	\caption{
		Examples where the explanations are the most inconsistent, under SSIM similarity, when a hyperparameter changes.
		Across the entire dataset, the reference images caused highest sensitivity (\ie lowest SSIM scores) for different attribution methods and their respective hyperparameter settings for both GoogLeNet (a) and ResNet (b).
	}
	\label{fig:worst_case}
\end{figure*}
\end{document}

%% file: math_commands.tex
\usepackage{amsmath,amsfonts,bm}

\def\eqref#1{equation~\ref{#1}}
\def\floor#1{\lfloor #1 \rfloor}
\def\1{\bm{1}}

\def\va{{\bm{a}}}

\def\vm{{\bm{m}}}

\def\vx{{\bm{x}}}

\def\vz{{\bm{z}}}

\DeclareMathAlphabet{\mathsfit}{\encodingdefault}{\sfdefault}{m}{sl}
\SetMathAlphabet{\mathsfit}{bold}{\encodingdefault}{\sfdefault}{bx}{n}

\def\gH{{\mathcal{H}}}

\def\gN{{\mathcal{N}}}

\def\sR{{\mathbb{R}}}

\DeclareMathOperator*{\argmin}{arg\,min}